\documentclass{article}
\usepackage[a4paper,left=1in, right=1in, top=1in, bottom=1in]{geometry}

\usepackage{natbib}
 \bibpunct[, ]{(}{)}{,}{a}{}{,}%

\usepackage{hyperref}

\newtheorem{lemma}{Lemma}
\newtheorem{observation}{Observation}

\usepackage{graphicx}
\usepackage{comment}
\usepackage{amsmath}
\usepackage{amssymb}
\usepackage[dvipsnames]{xcolor}
\usepackage{color,soul}
\usepackage{tikz,pgfplots}
\usepackage[acronym,nomain,nonumberlist, section=subsection]{glossaries}
\usepackage{enumitem}

\pgfplotsset{compat=1.18}
\usepackage{dirtytalk}
\usepackage{makecell}
\usepackage{booktabs}  
\usepackage{multirow}
\usepackage{rotating}
\usepackage{esvect}
\usepackage{siunitx}

\usetikzlibrary{patterns}
\usetikzlibrary{backgrounds, shapes}
\usetikzlibrary{fit,arrows.meta,calc,positioning,chains,shapes.geometric}
\usepgfplotslibrary{fillbetween}
 \usepackage{subcaption}
 \usepgfplotslibrary{statistics}
 \usepackage{algorithm}
\usepackage[noend]{algpseudocode}
\usepackage{adjustbox}
\definecolor{colorTSDark}{HTML}{071E47}
\definecolor{colorTSLight}{HTML}{aabad5}

\date{}
\title{Learning Dynamic Selection and Pricing of Out-of-Home Deliveries}
\author{Fabian Akkerman - University of Twente \cr Peter Dieter - Paderborn University \cr Martijn Mes - University of Twente}
\begin{document}
\maketitle








\begin{abstract}
Home delivery failures, traffic congestion, and relatively large handling times have a negative impact on the profitability of last-mile logistics. A potential solution is the delivery to parcel lockers or parcel shops, denoted by out-of-home (OOH) delivery. In the academic literature, models for OOH delivery were so far limited to static settings, contrasting with the sequential nature of the problem. We model the sequential decision-making problem of which OOH location to offer against what incentive for each incoming customer, taking into account future customer arrivals and choices. We propose Dynamic Selection and Pricing of OOH (DSPO), an algorithmic pipeline that uses a novel spatial-temporal state encoding as input to a convolutional neural network. We demonstrate the performance of our method by benchmarking it against two state-of-the-art approaches. Our extensive numerical study, guided by real-world data, reveals that DSPO can save $19,9\%$pt in costs compared to a situation without OOH locations, $7\%$pt compared to a static selection and pricing policy, and $3.8\%$pt compared to a state-of-the-art demand management benchmark. We provide comprehensive insights into the complex interplay between OOH delivery dynamics and customer behavior influenced by pricing strategies. The implications of our findings suggest practitioners to adopt dynamic selection and pricing policies.
\end{abstract}



\maketitle

\section{Introduction}
In 2021, the worldwide courier, express, and parcel (CEP) market was valued at $\$407.7$ billion, with an estimated annual growth rate of $6.3\%$ \citep{cepMarket}. Last-mile delivery, particularly in urban areas, accounts for $40-50\%$ of total CEP distribution costs. Factors such as traffic congestion, failed deliveries, and large handling times contribute significantly to these costs \citep{DALLACHIARA202026,Dalla2021,CHEN201753,RANJBARI2023103070}. For instance, up to $10\%$ of first delivery attempts fail, contributing to an average cost of $\$17$ per failure \citep{loq}. In the UK alone, the estimated costs resulting from failed home deliveries exceed one billion dollars per year \citep{Arnold2018}. Moreover, failed deliveries have environmental consequences, notably increasing CO\textsubscript{2} emissions \citep{Edwards2009}. These delivery failures not only impose financial burdens on service providers but also lead to customer dissatisfaction, negatively impacting the online shopping experience \citep{KEDIA2017587}. Traffic congestion, unsuccessful deliveries, and handling time collectively contribute to approximately $28\%$ of the costs and $25\%$ of the emissions {in the supply chain sector} \citep{CHEN201753}.

To mitigate long driving and handling times and failed deliveries, an alternative is the implementation of out-of-home (OOH) delivery systems. These systems involve delivering parcels to predetermined staffed locations or automated lockers from which customers can collect their items. The availability of OOH locations has increased by up to $36\%$ annually in Europe \citep{lme_ooh}, indicating a growing trend. {OOH} delivery can potentially save costs for logistics companies by reducing delivery failures and aggregating demand \citep{Song2009,Savelsbergh2016}. Previous studies have shown that OOH delivery can decrease driving distances by up to $60\%$ \citep{ENTHOVEN2020104919} and is generally favored by customers \citep{RANJBARI2023103070}. 

Integrating {OOH} delivery in last-mile logistics is an important and challenging research topic that provides many future research directions on the strategic (e.g., {OOH} facility location), tactical (e.g., {OOH} location capacity, customer discounts), and operational (e.g., dynamic selection and pricing of {OOH} deliveries) decision-making levels. In this paper, we tackle the emergent challenge of dynamically offering OOH delivery options and providing dynamic prices (discounts or charges) to steer customer behavior. Aside from providing problem insights, we propose a dynamic selection and pricing policy with a novel spatial-temporal state encoding utilizing a convolutional neural network (CNN). We provide empirical evidence that our method outperforms existing state-of-the-art benchmarks across both small synthetic instances and a large real-world case from Seattle, USA.

In the context of operational decision-making for {OOH} deliveries, previous research has focused on static problem settings, where all customer locations are known when decisions are made \citep[see, e.g.,][]{Grabenschweiger2021, MANCINI2021105361, ZHANG2023103128, galiullina2024demand}. However, in many real-world scenarios, customers arrive sequentially, and their delivery choices are subject to factors unknown to the route planner, e.g., determined by personal preferences or past experiences.

In this paper, we attempt to fill this research gap by considering the dynamic decision of allocating {OOH} locations to newly arriving online customers and the possibility of influencing customer behavior by providing monetary incentives. The incentives can be positive or negative, representing a fee or a discount for the customer. The selection and pricing of OOH locations are by no means trivial decisions as we need to anticipate future customer arrivals, customer behavior, and the impact of discounts and charges on customer choice. Our problem is dynamic as we make selection and pricing decisions at each time step without having foresight of future customer arrivals. Consequently, we model our problem as a Markov Decision Process (MDP). Exact methods are not suitable for solving this MDP, as the selection and pricing decision must be made swiftly for typical online retail platforms: a 100-millisecond delay in the load time of websites can decrease sales conversion by 7\% \citep{akamai}. Currently, many retailers give customers complete freedom in selecting an {OOH} location, and, in some countries, retailers offer a fixed customer discount for choosing an {OOH} delivery. In the future, if {OOH} delivery gets a larger share of total deliveries, smarter selection and pricing policies will be needed {to fully exploit the potential of {OOH} delivery.}

Our contribution to scientific literature is threefold. First, to our knowledge this paper is the first to formally define and study the OOH selection and pricing problem in a sequential decision-making context. Second, we compare several state-of-the-art solution methods from the literature and propose a novel machine learning-based approach. Two benchmark policies are derived from time slot demand management literature \citep{yang2016}, as this literature stream shares many similarities with our problem regarding dynamically arriving customer orders and intertwined pricing and routing decisions. However, our work focuses on the spatial dimension (where should customers be served?) rather than the temporal dimension (when should customers be served?) as studied before \citep[see, e.g.,][]{YANG2017935, Klein2018,YILDIZ2020230}. Furthermore, we benchmark against a proximal policy optimization (PPO) algorithm \citep{SchulmanEtAl2017}, which is a state-of-the-art actor-critic reinforcement learning approach. As a third contribution, our extensive numerical study offers novel insights for both practitioners and researchers into both the problem and various solution methods. The numerical study is performed on synthetic problem instances \citep{Gehring2002} as well as on instances derived from real-world data. We use publicly available Amazon order data of the greater Seattle area to construct these real-world instances \citep{Merchan2022}. Our code and used data can be found at: \url{https://github.com/frakkerman/ooh_code}. 

The remainder of this paper is structured as follows. In Section~\ref{sect:literature}, we present related work and outline the research gap. The problem and customer choice model are described in Section \ref{section:model}. The solution method is presented in Section \ref{section:sol}. Section \ref{section:numericalStudy} includes the numerical study and Section \ref{section:conclusion} concludes the paper.

\section{Related Work}\label{sect:literature}

First, in Section~\ref{section:relatedWork:pricing}, we introduce previous studies on last-mile logistics that consider demand management through offering or pricing {of delivery options}. Next, we discuss related work on {OOH} delivery in Section~\ref{section:relatedWork:deliveryPoints}.

\subsection{Demand Management in Last-mile Logistics}\label{section:relatedWork:pricing}

Demand management is a key focus in last-mile logistics research, particularly regarding the offering and pricing of time slots for attended home delivery (AHD). The AHD demand management field is typically divided into the following categories: static time slot offering, dynamic time slot offering, differentiated time slot pricing, and dynamic time slot pricing \citep{agatz2013,yang2016,klein2019}. While restricting the time slot choice is seen as potentially leading to lost sales and customer dissatisfaction \citep{ASDEMIR2009246}, pricing is advocated as a more effective approach to balancing profits against lost sales. 

In their study, \cite{Campbell2006} explore the challenges of home delivery scheduling where customers are incentivized to choose different times of the day and broader delivery windows. Their research is centered on developing algorithms to assess the viability and cost implications of various time slot options, taking into account existing customer requests. They operate under the assumption that the likelihood of customers selecting specific time slots is known in advance. \cite{ASDEMIR2009246} addresses this issue by considering a more advanced customer choice model, namely a multinomial logit model. Concerning routing costs, more recent research approximates these not solely based on requests already in the system, but also by anticipating future requests \citep{yang2016, Klein2018}. 

Currently, approximate dynamic programming approaches have been the method of choice in AHD pricing research, see, for instance, \citet{YANG2017935, KOCH2020633} and \citet{Vinsensius2020}. \cite{Ulmer2020Pricing} considers a same-day delivery problem, where booking and service horizons elapse simultaneously. By implementing an anticipatory pricing and routing policy, they incentivize customers to choose delivery deadlines that optimize fleet utilization, thereby increasing revenue and the number of same-day orders fulfilled. \citet{STRAUSS20211022} explore the concept of flexible time windows, where customers receive a financial incentive to be notified of their service window only shortly before dispatch. In contrast, \citet{YILDIZ2020230} shift the focus to incentives for selecting delivery days, rather than time windows within a day. {Recent work also studies demand management in logistics outside the AHD context, for instance, the matching of requests in peer-to-peer logistics, considering the finite resources of the offering party \citep{KARABULUT2022, ausseil2022}.} Collectively, these studies underscore a trend towards more dynamic and customer-focused solutions in demand management for last-mile logistics.

\subsection{Last-mile Logistics with {OOH} Delivery}\label{section:relatedWork:deliveryPoints}

OOH delivery is gaining attention in the scientific literature as it is already used in practice. {On a strategic level, OOH research typically focuses on} deciding on the location of OOH facilities and the capacity of such facilities. Facility location problems are often solved using exact or matheuristic approaches, as demonstrated in \citet{Deutsch2018, XU2021102280, KAHR2022102721, LUO2022105677, LIN2022102541, lyu2022, MANCINI2023}, and \citet{RAVIV2023103216}. In summary, the recent OOH facility location literature concludes that (i) adopting OOH enables logistics {service} providers to cope with large demand increases, (ii) OOH locations should be situated in densely populated areas that are frequently traversed by vehicle routes, (iii)  small and medium-sized OOH lockers are often preferred over larger, more costly lockers, and (iv) opening too many OOH locations in an area can {negatively} affect profitability. 

On a tactical and operational level, the literature considers heuristics for variants of the vehicle routing problem that incorporate OOH locations, such as the work of \citet{jiang2020,pan2021,DUMEZ2021103}, and \citet{PENG2023109044}. \citet{ZHOU2018765} discuss a multi-depot two-echelon VRP where customers have the option of home delivery or OOH delivery. \citet{MANCINI2021105361} and \citet{Grabenschweiger2021} explore the impacts of fixed discounts and heterogeneous locker boxes on customer acceptance in VRPs with OOH locations. Additionally, \citet{Ulmer20191} address same-day delivery scenarios where all customer requests require delivery to pick-up stations. Emerging topics in operational OOH research include the use of alternative delivery modalities and mobile parcel lockers to increase customer service, as discussed in studies by \citet{ENTHOVEN2020104919,GHADERI2022108549,VUKICEVIC2023106263,SCHWERDFEGER2022103780}, and \citet{LIU2023103234}. In most works, customer choice has been modeled as a constant function in relation to the distance traveled to OOH locations \citep{MANCINI2021105361,Grabenschweiger2021,SCHWERDFEGER2022103780,PENG2023109044, Janinhoff2023}, or using a ranking-based approach \citep{DUMEZ2021103}. Many studies do not model customer choice beyond a preference list, or assume full control by the retailer \citep{ZHOU2018765,Arnold2018,Sitek2019,ENTHOVEN2020104919}. 
{A few recent studies have incorporated customer choice models for OOH delivery. 
\cite{janinhoff2023stochastic} model customer choice based on a random utility derived from the distance to delivery points, while \cite{galiullina2024demand} model it using a price-based probability. \cite{ZHANG2023103128} propose a convex optimization model to find the customer distribution across delivery options. Only the latter two studies consider pricing.} For a comprehensive classification of recent OOH literature, we refer to \citet{Janinhoff2023}.

Concluding, only a limited number of papers consider the selection and pricing of OOH to influence customer behavior. Furthermore, all described studies assume static settings and do not model a sequential decision process as likely experienced by many service providers, leading to a call for research that incorporates these real-world complexities. To the best of our knowledge, this is the first paper that considers the offering and pricing of OOH delivery in a sequential decision-making context while modeling the stochastic nature of customer behavior and customer arrivals, as observed in practice.

\section{Model}\label{section:model}

In this section, we provide the model for the {OOH} selection and pricing problem. We provide the general problem description in Section~\ref{section:problem}, and detail the customer choice model in Section~\ref{section:custChoice}. We end the section with an overview of all notation.

\subsection{Problem Description}\label{section:problem}

Our model of the OOH selection and pricing problem is motivated by the situation at online retailers where customers choose between home delivery or delivery at a parcel locker or shop. We consider a retailer that serves customers using a fixed fleet of vehicles. Before the delivery day, customers order products on the retailer's website and, upon checkout, select a delivery option: home or OOH. We consider a fixed booking horizon $[0,T]$, where $T$ is the cutoff time, i.e., the time after which no new customer can request delivery for the next day. During the booking horizon, customers $c_t \in \mathcal{B}_T$ arrive at discrete time steps $t$, and the number of customers that arrive on a given day follows an unknown distribution ${D}$. We consider a situation where different delivery options $k \in \mathcal{K}$ represent different costs, and the customer can be nudged to specific options by (i) offering only a subset of all delivery options in $\mathcal{O} \subseteq \mathcal{K}$, and (ii) discounts and charges per delivery option $k$. {The set of delivery options \( \mathcal{K} \) includes a home delivery option \( h \) and a fixed set of OOH locations \( \mathcal{L} \). Thus, \( \mathcal{K} = \{ h \}  \cup \mathcal{L} \).} Without loss of generality and in line with retailer practice, we always offer home delivery as a delivery option. {For example, if \( \mathcal{L} = \{\text{OOH 1}, \text{OOH 2}, \text{OOH 3}\} \), then \( \mathcal{K} = \{h, \text{OOH 1}, \text{OOH 2}, \text{OOH 3}\} \). The subset of offered options \( \mathcal{O} \) might be \( \{h,\text{OOH 1}\} \), meaning that the customer is offered the choice between home delivery and OOH location 1.} {We conduct experiments with finite and infinite capacity lockers. The remaining capacity of an OOH location $l$ at time $t$ is denoted by $k_{t,l}$. } {Customers are divided into a finite set of segments $g\in\mathcal{G}$, with $\mu_g$ being the probability that a customer from segment $g$ arrives on an arbitrary time $t$.} Customer choice behavior is modeled using a multinomial logit (MNL) model, detailed in Section~\ref{section:custChoice}.

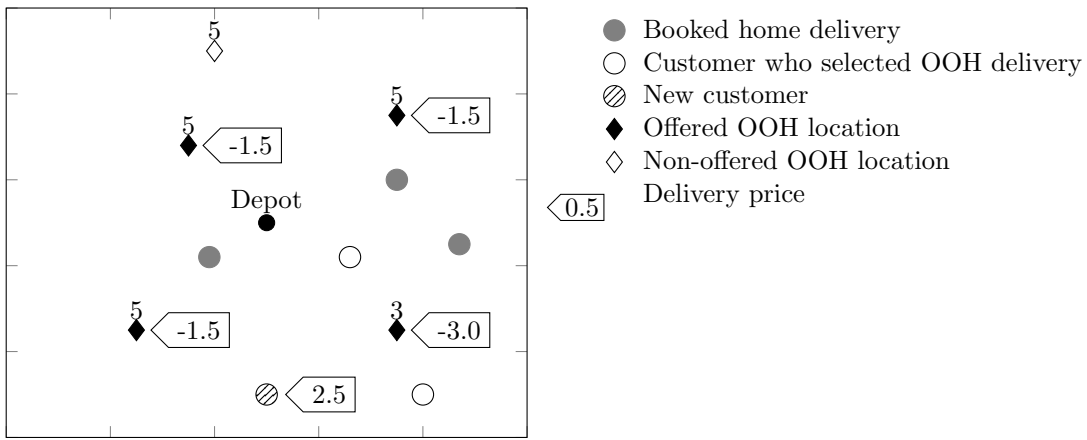
\begin{figure}[b]
    \centering

\begin{tikzpicture}
\centering

  \pgfplotsset{
      xmin=0, xmax=100,
      ymin=0,ymax=100
  }

  \begin{axis}[
  legend style={at={(1.6,1)},anchor=north, draw=none},
    legend columns=1,
   yticklabels={,,},
   xticklabels={,,},
   legend cell align={left}
  ]
    \addplot[only marks,mark=*,mark size=4pt,gray]
    coordinates{ 
      (39,42)
      (75,60)
      (87,45)
    };

    \addplot[only marks,mark=o,mark size=4pt]
    coordinates{ 
      (66,42)
      (80,10)
    };
    
        \addplot[only marks,mark=*,mark size=4pt,pattern=north east lines,pattern color=black]
    coordinates{ 
      (50,10)
    };

    \node[signal,draw=black,signal to=west] at (62,10) {2.5};
    \node[signal,draw=black,signal to=west] at (37,25) {-1.5};
    \node[signal,draw=black,signal to=west] at (87,25) {-3.0};
    \node[signal,draw=black,signal to=west] at (47,68) {-1.5};
    \node[signal,draw=black,signal to=west] at (87,75) {-1.5};

    \addplot[only marks,mark=diamond*,style={solid, fill=black}, mark size = 4pt]
    coordinates{ 
      (25,25)      
      (35,68) 
      (75,25)
      (75,75) 
    };

    \addplot[only marks,mark=diamond*,style={solid, fill=white}, mark size = 4pt]
    coordinates{ 
      (40,90)      
    };

            \addplot[mark=*,forget plot,only marks,mark size=3pt]
    coordinates{ 
      (50,50)      
    };
    \node [above] at (axis cs:  50,  50.1) {Depot};

    \node [above] at (axis cs:  75,  25.5) {3};
    \node [above] at (axis cs:  25,  25.5) {5};
    \node [above] at (axis cs:  75,  75.5) {5};
    \node [above] at (axis cs:  35,  68.5) {5};
    \node [above] at (axis cs:  40,  90.5) {5};

\addlegendimage{no markers, white}

     \addlegendentry{Booked home delivery}
     \addlegendentry{Customer who selected OOH delivery}
      \addlegendentry{New customer}
    \addlegendentry{Offered OOH location}
    \addlegendentry{Non-offered OOH location}
    \addlegendentry{Delivery price}

  \end{axis}

\node[signal,draw=black,signal to=west,  inner sep=0.06cm] at (7.58,3.05) {0.5};

\end{tikzpicture}
    \caption{Exemplary situation during the booking horizon.}
    \label{fig:examplaryModel}
\end{figure}

The main focus of this study is the pricing of delivery options, i.e., the delivery charge or discount given to home and OOH delivery. However, we also consider the selection of a subset of delivery options $\mathcal{O} \subseteq \mathcal{K}$ to be offered to the customer. We illustrate these decisions in Figure~\ref{fig:examplaryModel}. Here, at a given time step $t$, five customers already arrived and accepted a delivery option. Three customers requested home delivery, and two customers chose to have their parcels delivered at an OOH location. {Each OOH location has a finite capacity of $5$ parcels, the number shown above the OOH location indicates the remaining capacity.} Next, a new customer arrives, which requires the retailer to offer different delivery options. The offered delivery locations get a delivery charge or discount. Here, the home delivery will cost the customer a delivery charge of $2.5$, whereas choosing one of the four offered OOH locations will provide the customer a discount of $1.5$ or $3.0$, respectively. One OOH location is not offered to the customer.

An important characteristic of our problem is its dynamic nature. The retailer does not know how many customers will arrive before the cutoff time $T$, nor does it know the home locations of future customers. A customer that might seem remote and therefore expensive, might receive a high delivery charge for home delivery and a high incentive for choosing an OOH delivery. However, as the booking horizon unfolds, more customers might appear close to this customer's home location. This means that the customer, in hindsight, is not remote at all, and too high incentives were given. 

We consider the following sources of costs: (i) costs related to operational time expressed in driver salaries $C^w$, (ii) costs related to driving distance expressed in fuel costs $C^f$, (iii) costs related to providing discounts to customers $j_t^-$, and (iv) costs related to delivery failures $C^m$. We assume drivers to be service providers who are paid per hour. We do not include fleet costs since we consider a fixed fleet of $V$ vehicles. Sources of revenue are the sales revenue per customer $r_t$ and the collected delivery charges $j_t^+$. Note that the total costs and revenues are calculated after the booking horizon has ended at time $T$, when all customers $\mathcal{B}_T$ have been revealed. In line with practice, we model the possibility of delivery failure. With probability $\mathbb{P}^m$, home delivery will fail. Failed delivery will result in a fixed monetary penalty of $C^m$. We assume that delivery to an OOH location never fails. We do not consider delivery time windows. The goal of the retailer is to maximize the total profits.

The sequence of events at a given time step $t$ is as follows:
\begin{enumerate}[leftmargin=2cm]
    \item A customer $c_t$ arrives at time $t$ and fills the online basket.
    \item At the checkout, the customer gets to see different prices per delivery option $k \in \mathcal{O}$.
    \item The customer selects a delivery option $k$ $(c^{choice}_{k,t}=1)$ and leaves the online system.
    \item The {selected delivery location is added to} $\mathcal{B}_{t-1}$.
    \item {If applicable, the remaining capacity of the chosen OOH location is reduced.}
\end{enumerate}

During this event sequence, the decision-maker has to decide on the selection and pricing at step~2. Our assumed sequence of events omits the possibility that a customer first chooses a delivery option before completing the basket. After the booking period, at cutoff time $T$, the retailer plans routes serving all booked customers, after which the routes are executed by the fleet. The decision problem can be formulated as an MDP. We define a state space $\mathcal{S}$, a decision space $\mathcal{A}$, a cost function $c: \mathcal{S} \times \mathcal{A} \rightarrow \mathbb{R}$, and transition dynamics $\mathbb{P}: \mathcal{S} \times \mathcal{A} \times \mathcal{S} \rightarrow [0,1]$. Here, the function $\mathbb{P}$ maps the probability of reaching a state to each state-decision pair. A state $s_t \in \mathcal{S}$ at time $t$ is represented by a discrete tuple. A state consists of:
\begin{equation}
    s_t = [ c_t, \mathcal{B}_{t-1}, \vv{\kappa_t} ],
\end{equation}
where $c_t$ represents the newly arrived customer, $\mathcal{B}_{t-1}$ represents all booked delivery locations, and {$  \vv{\kappa_t}=\{k_{t,l}| l \in \mathcal{L}\}$ represents the vector of remaining capacities of all OOH locations}. We use $a^{selection}_k$ to denote the binary decision of selecting a delivery location $k$ to be offered to the customer. Next, $a^{pricing}_k$ denotes the pricing decision for this location.  Each delivery price is bound by a maximum discount and maximum charge, $a^{pricing}_{k} \in [a,b]$. Here, a negative price indicates a discount and a positive price indicates a delivery charge. We consider multi-dimensional decisions, represented as a vector {of tuples,} and for conciseness referred to as a decision:

\begin{equation}
   \vv{a_t} = [a^{selection}_k,a^{pricing}_k],\quad \forall k \in \mathcal{K}.
\end{equation}

The transition from state $s_t$ to state $s_{t+1}$ is done by using the information resulting from the decision and exogenous events. More precisely, we add the customer choice for a delivery location to the system, add the new stop, {reduce remaining OOH capacity if applicable,} and observe a new customer arrival. The cost or revenue from a discount or delivery charge at a given time step $t$ can be calculated using:
\begin{equation}
    j_t = \sum_{k\in\mathcal{K}}a^{pricing}_{k,t} c_{k,t}^{choice},
\end{equation}
where $j_t^- = -\min\{j_t,0\}$ denotes cost from providing discounts, and $j_t^+ = \max\{0,j_t\}$ denotes revenue from delivery charges. The binary variable $c_{k,t}^{choice}$ denotes the choice of the customer, where $k=h$ indicates home delivery.

The routing plan made after cutoff time $T$ is denoted by {${R}_T$}. The routing dimension of our problem consists of a capacitated vehicle routing problem (CVRP). The directed graph $\mathcal{G} = (\mathcal{V},\mathcal{E})$ models the system of vertices $\mathcal{B}_T \cup \mathcal{W}$ with delivery and depot locations. The travel distance and time on an edge $(i,j) \in \mathcal{E}$ is expressed by $d_{i,j}$ and $w_{i,j}$, respectively. Note that we model travel distance and time separately to account for variable travel speed and congestion. It is allowed to combine loads to the same location (one OOH location), but it is also allowed to split them over multiple vehicles. A vehicle always starts and ends at the same depot and has a fixed capacity of $K$ customers. A vehicle route must adhere to individual vehicle capacity constraints. A vehicle can only leave from a location $i$ after the service duration $l_i$, which accounts for the time needed for parking, walking, and delivering parcels. {The service duration at the depot, $l_0$, is zero.}

We calculate the total costs, together with routing costs, after the cutoff time $T$. Hence, the objective is to maximize total profits on any given day:
\small 
\begin{equation}\label{eq:profitfunc}
    \max  \left(\sum_{t=0}^T r_t+j_t^+ - j_t^- \right) -
         C^w\left(\sum_{i\in {R}_T}\sum_{\substack{j\in {R}_T \\ j \neq i}} w_{i,j} + l_i \right) - C^f \left({\sum_{i\in {R}_T}\sum_{\substack{j\in {R}_T \\ j \neq i}} d_{i,j}}\right) - C^m \left \lceil\mathbb{P}^m  \sum_{t=0}^T c_{k=h,t}^{choice} \right\rceil.
\end{equation}
\normalsize 

We consider a policy: $\pi: \mathcal{S} \rightarrow \mathcal{A}$ that defines the selection and pricing behavior. Our primary objective is to minimize costs as a key component of our analysis.

\subsection{Customer Choice Model}\label{section:custChoice}

Inspired by \citet{yang2016}, we model customer choice, $c^{choice}_{k,t}$, using the MNL model. For the MNL model, we assume customers are utility maximizers, i.e., a customer always selects a delivery option that has the highest utility. {Note that the pricing bounds $[a,b]$ ensure that prices remain realistic and reduce unfairness, given the objective to minimize costs.}  The utility of a delivery option $k$ to customer segment $g$ is given by:
\begin{equation}
    u_{k,g} =  -\beta_{g}^k  \mathrm{exp}\left[{d_{0,k}}\right] +  \beta_{g}^d a^{pricing}_{k}  + \epsilon.
\end{equation}\label{eq:betak}

Similar to the prevailing focus in related literature, a delivery location's utility is based on its distance from the customer's home address. {Hence, $\beta_{g}^k$ is the sensitivity of the utility to the distance between the home address and the offered delivery location.} $\beta_{g}^d$ is the sensitivity of the utility to the delivery price $a^{pricing}_{k}$, for brevity denoted without the subscript $t$.  We ensure that $d_{0,k}$ does not become too large, preventing the exponential term from dominating the utility. Similar to \citet{lyu2022}, who study the OOH facility location problem, we split the utility of home delivery ({$u_{k,g}$, with $k=h$}) from the utility given to OOH locations and consider it a tunable parameter. Apart from the deterministic components, we include $\epsilon$, which is an i.i.d. random variable that follows the standard Gumbel distribution with $\mu=0$ and $\beta=1$, as common for MNL models. Since we do not consider customer walk-aways, the probability that a customer chooses delivery option $k$ from all offered options $\mathcal{O}$ given the vector of delivery prices $\{a^{pricing}_{0},a^{pricing}_{1},\ldots\} = \vv{a}^{pricing}$ is:
\begin{equation}
    \mathbb{P}_{k,g}( \vv{a}^{pricing}) = \frac{\mathrm{exp}\left[  -\beta_{g}^k \mathrm{exp}\left[{d_{0,k}}\right]  +  \beta_{g}^d a^{pricing}_{k} \right]}{\sum_{k\in\mathcal{O}}\mathrm{exp}\left[ -\beta_{g}^k \mathrm{exp}\left[{d_{0,k}}\right]  +  \beta_{g}^d a^{pricing}_{k}\right]}.
\end{equation}

{To ease notation, we omit a base utility, as it is absorbed by the sensitivity parameters. A base utility can be added without loss of generality.} For more details on the MNL choice model, we refer to \citet{book_train_2009} and \citet{yang2016}. We end this section with a summary of all relevant notation, see Table \ref{tab:Notation}.

\begin{table}[h]
\caption{Summary of Notation.}
\label{tab:Notation}
 \begin{adjustbox}{max width=1\textwidth}
    \begin{tabular}{r l}
    \hline
        Variable & Description \\
    \hline
        \quad \textbf{General variables} &\\
          $T$ & the cutoff time after which no new customers can be booked\\
          {$l \in \mathcal{L}$} & {the set of all OOH locations}\\
          {$h$} & {the home delivery option offered to the customer}\\
          ${k} \in \mathcal{K}$ &  the set of all delivery locations, {\(\mathcal{K} = \mathcal{L} \cup \{h\}\)}\\
          ${D}$ & the CDF of the number of arriving customers \\
          {$g \in \mathcal{G}$} & {the customer segments}\\
          {$\mu_g$} & {the probability that a customer arrives from segment $g$}\\
          $V$ & the fleet size\\
          $K$ & the carrying capacity per vehicle per day\\
          $C^w$ & the salary costs per hour per driver\\
          $C^f$ & the fuel costs per distance unit\\
          $C^m$ & the fixed costs paid per delivery failure\\
          $j_{t}^-$ & the costs from a discount at time $t$\\
        $j_{t}^+$ & the revenue from a delivery charge at time $t$\\
          $r_{t}$ & the revenue from a customer sale at time $t$\\
          {${R}_T$} & the final routing plan serving all customers \\
           $d_{i,j}$ & the travel {distance} on an edge connecting vertices $i$ and $j$\\
          $w_{i,j}$ & the travel {time} on an edge connecting vertices $i$ and $j$\\
          $l_i$ & the service duration at location $i$\\
          $\mathbb{P}^m$ & the probability of delivery failure at any home address\\
          $u_{k,g} $ & the utility attributed to delivery option $k$ for segment $g$  \\
          $\beta_{g}^k$ & {the sensitivity of the utility for segment $g$ to the distance between the home and the OOH location}\\
          $\beta_{g}^d$ & the sensitivity of the utility of segment $g$ to the given price \\
         $\epsilon$ & the i.i.d. random MNL noise per delivery option $k$ \\
         $\mathbb{P}_{k,g}( \vv{a}^{pricing})$ & the probability a customer from segment $g$ selects option $k$, given the price vector $\vv{a}^{pricing}$\\
\cline{2-2}
        \textbf{State variables} &\\
        $c_t$ & the customer arriving at time $t$\\
            $\mathcal{B}_{t-1}$ &  {the booked delivery stops} at time $t-1$\\
            {$\vv{\kappa_t}$} & {the vector of remaining capacities of all OOH locations, $\vv{\kappa_t}=\{k_{t,l}| l \in \mathcal{L}\}$ at time $t$} \\

 \cline{2-2}
        \textbf{Policy variables} &\\
           $\mathcal{O} \gets a^{selection}(s_t)$ &  the subdecision of selecting a set of delivery locations to offer given state $s_t$\\
           $a^{selection}(s_t) \gets \pi^{selection}(s_t)$ &  the selection subpolicy given state $s_t$\\
            $ a^{pricing}_k(s_t,a^{selection}(s_t))$ &  the subdecision of pricing delivery locations given state $s_t$ and the selection decision\\
            $a^{pricing}(s_t) \gets \pi^{pricing}(s_t,a^{selection}(s_t))$ &  the pricing subpolicy given state $s_t$\\
            $[a,b]$ & the bounds of delivery prices\\
           $a(s_t) \gets \pi$ & the joint policy of selection and pricing given state $s_t$ \\
          \hline
    \end{tabular}
        \end{adjustbox}
\end{table}

\vspace{-1em}
\section{Solution Method}\label{section:sol}
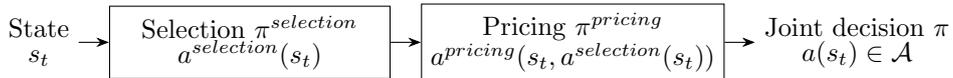
\begin{figure}[hbtp]
    \centering
        \begin{tikzpicture}[
      start chain = going right,
      node distance=15mm,
  alg/.style = {draw, align=center, font=\linespread{0.8}\selectfont}
                    ]
    \begin{scope}[every node/.append style={on chain, join=by -Stealth}]
        \node (n1) [draw=white,align=center, font=\linespread{0.8}\selectfont] {State \\ \(s_t\)};
        \node (n2) [alg,xshift=-1.1cm,minimum height=2.6em,minimum width=10.5em]  {Selection $\pi^{selection}$ \\ \(a^{selection}(s_t)\)};
        \node (n2) [alg,xshift=-1.1cm,minimum height=2.6em,minimum width=10.5em]  {Pricing $\pi^{pricing}$ \\ \(a^{pricing}(s_t,a^{selection}(s_t))\)};
        \node (n3) [xshift=-1.1cm,draw=white,align=center, font=\linespread{0.8}\selectfont]  {Joint decision $\pi$ \\ $a(s_t)\in\mathcal{A}$}; 
    \end{scope}
    
    \end{tikzpicture}
        \caption{Pipeline for dynamic selection and pricing of out-of-home delivery.}\label{fig:pipeline}
\end{figure}

Figure~\ref{fig:pipeline} depicts the used decision-making pipeline for Dynamic Selection and Pricing of OOH delivery (DSPO). The pipeline entails two sub decisions forming the overall policy $\pi$ together. First, the state, containing the booked and newly arrived customers, is used as input to obtain the heuristic selection decision $a^{selection}$. Next, the decision $a^{selection}$ and state $s_t$ are used to derive the pricing decision, which is obtained from a supervised machine learning model. Both decisions are combined in a joint selection and pricing policy, $\pi: \mathcal{S} \rightarrow \mathcal{A}$. The remainder of this section details each step in the pipeline. We discuss selection and subsequently pricing in Section~\ref{sect:selection}. Furthermore, we detail the training procedure of the supervised machine learning model in Section~\ref{sect:training} and discuss our algorithmic design choices in Section~\ref{sect:discussionChoice}.

\subsection{Selection and Pricing Decision}\label{sect:selection}
For the selection decision, we use a simple heuristic rule that selects a subset of OOH locations. To be precise, we select the $N$ OOH locations closest to the customer's home address {that still have remaining capacity, i.e., $k_{t,l}>0$}. We ensure that $N$ is sufficiently large to prevent a decline in the probability of customers choosing an OOH delivery option due to a limited offering, as the utility of an OOH location is mainly determined by the proximity of an OOH location to the home address, see Section~\ref{section:custChoice}. {We limit the offering to $N$ to reduce the number of computations, especially relevant when the total number of OOH locations is large. In Appendix~\ref{app:choice_validation}}, we conduct a sensitivity analysis for different levels of $N$.

For the pricing decision, we employ a cost approximation that estimates the costs of adding a delivery location, being a specific OOH location or the customer's home address, to the delivery route. Note that we aim to estimate the impact of a stop on the \emph{final} delivery schedule, since a simple cheapest insertion in the current route with the so-far known stops does not suffice. In the remainder of this subsection, we subsequently detail (i) how we encode the state, (ii) the machine learning model used for delivery cost approximation, (iii) how we obtain training data, and (iv) how we obtain prices from the cost approximation.

\paragraph{State Encoding}
Recall that the state consists of a {set} containing all currently known delivery locations, both OOH locations and home addresses. {As common in practice, we do not want to discriminate customers by providing different pricing schemes to customers with the same location.} Therefore, we aggregate customer locations by counting the number of delivery locations in a specific aggregation area. We denote the state encoding by $\phi(s_t)$. We divide the total service area in $M$ spatial areas.
\begin{figure}[hbtp]
    \centering
    \includegraphics[width=\textwidth]{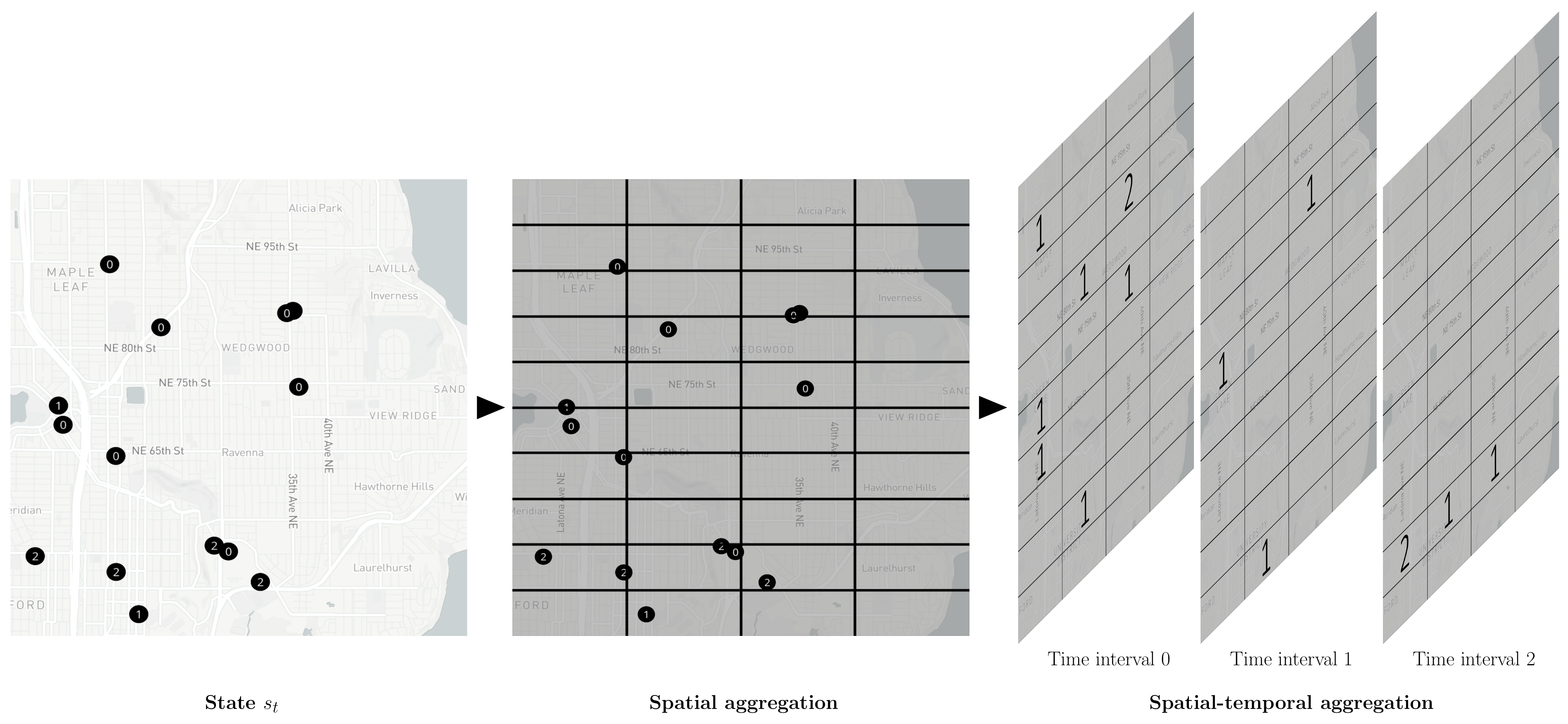}
    \caption{The state encoding for DSPO.}
    \label{fig:statencoding}
\end{figure}
Note that we do not know in advance how many customers will arrive during the booking horizon. However, the number of served customers on a day has a major influence on the per-customer delivery costs. Therefore, apart from the \emph{spatial aggregation} of the state, we also consider a \emph{temporal aggregation}. We aggregate customers based on $D^T$ pre-defined {and equal-length} time intervals of the booking horizon. An exemplary spatial-temporal state encoding $\phi(s_t)$ is shown in Figure~\ref{fig:statencoding}. Here, delivery locations are indicated by black dots, and the time intervals in which the customers arrived in the system ($D^T=\{0,1,2\}$) are indicated with the numbers in the dots. Although we do not know upfront how many customers will arrive on a specific day, this state encoding allows the CNN to estimate the arrival distribution. We consider both $M$ and $D^T$ to be tunable hyperparameters. The results of hyperparameter tuning can be found in Appendix~\ref{appendix:hyperparams}. {We provide the remaining capacity $k_{t,l}$ of an OOH location as a separate variable to our prediction model. In an infinite capacity setting, this variable is omitted.}

\paragraph{Cost Prediction Model}

\begin{figure}[hbtp]
    \centering
    \includegraphics[clip, trim=0cm 1cm 0.5cm 0.5cm,scale=0.5]{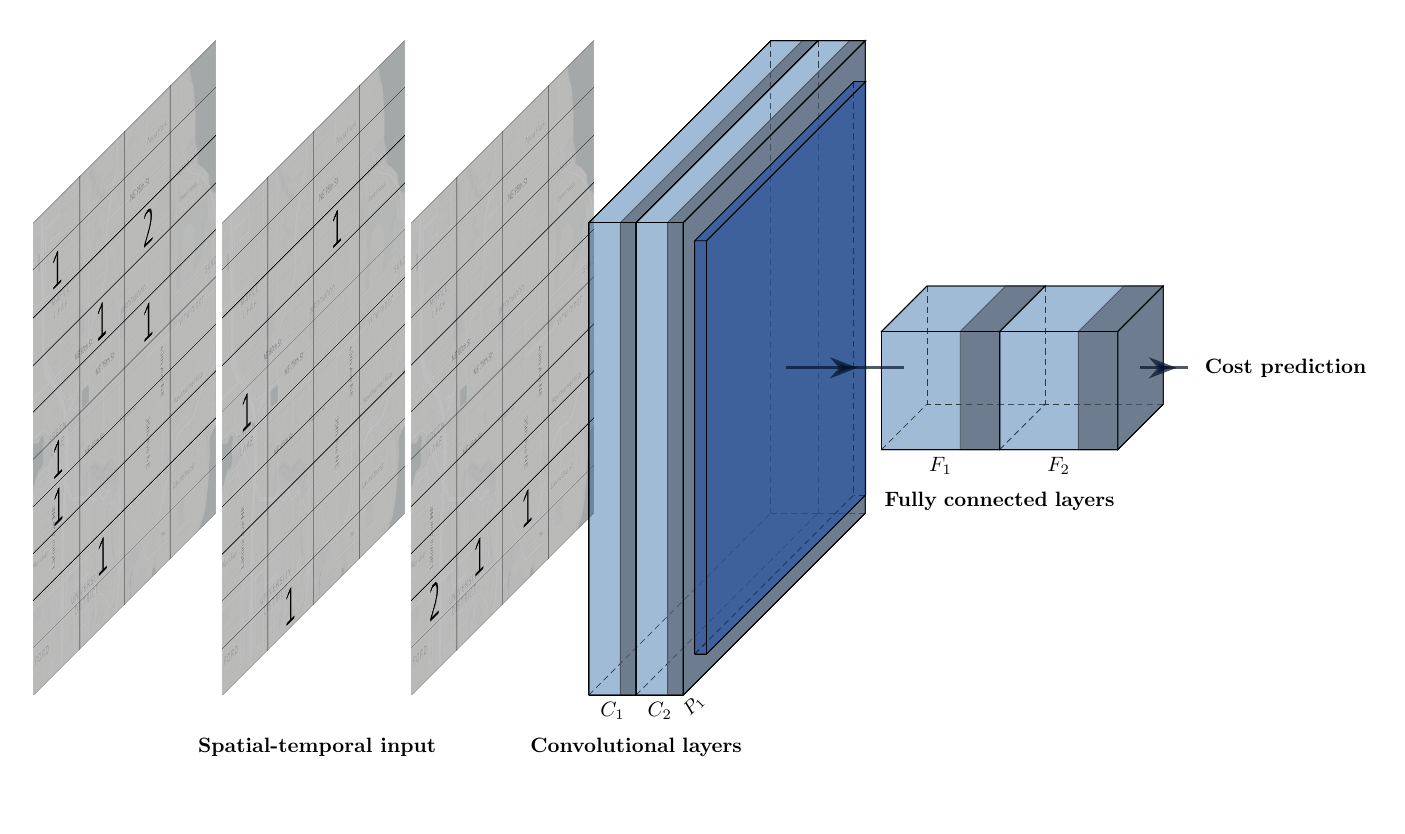}
    \caption{The CNN architecture for DSPO.}
    \label{fig:architecture}
\end{figure}

Our state encoding $\phi(s_t)$ is particularly suitable for CNNs. Unlike fully connected (FC) neural networks, CNNs take a multi-dimensional matrix as input and can abstract meaning from spatial relationships. In CNNs, convolution layers use kernel convolution, which is a process in which a small matrix, called kernel or filter, transforms the data by passing over multiple input dimensions. Kernel convolution transforms data into activation maps, which are feature abstractions from the raw data. In CNNs, convolution layers are commonly followed by \emph{pooling} layers, which allows the high-dimensional output data from the convolution operations to be reduced to a manageable dimension. Most CNNs use several convolution and pooling layers, see \citet{li2022} for an overview of popular CNN architectures. An efficient way to learn non-linear patterns from the CNNs output is to add FC layers after the final pooling layer. For a detailed explanation of CNNs, we refer to \citet{goodfellow}. Although many architectures for CNNs exist, we use what we consider a \say{vanilla} architecture, consisting of two convolution layers, an average pooling layer, and two FC layers. Our CNN architecture is depicted in Figure~\ref{fig:architecture}. {Remaining capacity is directly fed to the FC layers, after the convolutional layers.} We denote our neural network by $\mathcal{N}_\theta$, where $\theta$ are trainable weights. The neural network $\mathcal{N}_\theta$ takes as input the encoded state $\phi(s_t)$, and outputs a single value in $\mathbb{R}$. The goal of the neural network is to accurately predict the \emph{true} but unknown costs $C^{true}_{k,t}$ of \say{inserting} a delivery location $k$ for a customer that arrived at time $t$ in the final route ${R}_T$, given $\mathcal{B}_{t-1}$. Therefore, to obtain the expected costs of inserting a location $k$, we calculate:
\begin{equation}
    \hat{C}^{DSPO}_{k,t} = \mathcal{N}_\theta(\phi(s_{t},k)),
\end{equation}
where $(s_{t},k)$ is the current state with the potential new delivery location $k$ added. Note that we do inference for all offered delivery locations, i.e., the DSPO architecture yields a cost prediction vector for each offered OOH location and the home location. The hyperparameter tuning results of the CNN are summarized in Appendix~\ref{appendix:hyperparams}.

\begin{figure}[hbtp]
    \centering
    \begin{tikzpicture}
        \draw[ultra thick, -Stealth] (0,0) -- (12,0);
        
        \foreach \x in {3,6,9,12}
        \draw (\x cm,3pt) -- (\x cm,-3pt);
        
        \draw[ultra thick] (0,0) node[below=3pt,thick] {0} node[above=3pt] {};
        \draw[ultra thick] (3,0) node[below=3pt,thick] {t} node[above=3pt] {Store $\phi(s_t)$};
        \draw[ultra thick] (6,0) node[below=3pt,thick] {t+1} node[above=3pt] {Store $\phi(s_{t+1})$};
         \draw[ultra thick] (9,0) node[below=3pt,thick] {$t+2$} node[above=3pt] {Store $\phi(s_{t+2})$};
        \draw[ultra thick] (12,0) node[below=3pt,thick] {T} node[above=3pt] {};

        \draw[ultra thick] (14,0) node[thick, text width=3cm,align=center] {Solve CVRPs and obtain $C^{true}_{k,t} $} node[] {};
        
        \end{tikzpicture}
    \caption{An illustration of how training data is obtained during a booking horizon where three customers arrive.}
    \label{fig:datagen}
\end{figure}
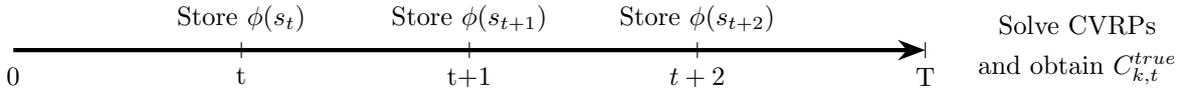

\paragraph{Obtaining Training Data}
For obtaining training data, we need to store features, i.e., the encoded state $\phi(s_t)$ as depicted in Figure~\ref{fig:statencoding}, and the actual costs related to this state, $C^{true}_{k,t}$. For a given customer, these costs are obtained by removing the customer's chosen delivery location from the final route ${R}_T$, obtaining a new route using a CVRP solver, and subsequently storing the difference in total routing time compared to the route that serves all customers. We calculate the costs related to $\phi(s_t)$ after the cutoff time $T$, when the accurate $C^{true}_{k,t}$ are known. As we know the service time $l_i$ related to a customer, we can obtain the total costs of salary and fuel related to a delivery location $i$. When $i$ is an OOH location, the costs related to $l_i$ are divided over the number of customers utilizing this delivery option. We disregard possible costs due to delivery failure in $C^{true}_{k,t}$, as in our model delivery failures cannot be predicted based on spatial-temporal data. Note that the features are stored \emph{during} the simulated booking horizon, but the actual costs related to the feature values are only obtained \emph{after} the cutoff time. The process of obtaining data during a simulation is illustrated in Figure~\ref{fig:datagen}. {For illustrative purpose, we assume that one customer arrives per time step $t$.} Here, we solve four CVRPs: one with three customers, and three CVRPs where the $1^{\textrm{st}}$, $2^{\textrm{nd}}$, and $3^{\textrm{rd}}$ customer is disregarded, respectively. This results in three \say{insertion} costs, which are the costs related to the stored features $\phi()$.

\paragraph{From Cost Prediction to Delivery Prices} We use the insertion cost for all offered delivery locations, ${C}^{true}_{k,t}$, to obtain optimal prices given the revenue, delivery costs, pricing revenue and costs, and the expected sensitivity of customers to delivery prices. For this, we use the proof in \citet{dong2009}. First, we simplify Equation~\ref{eq:profitfunc} and solve the online decision problem, for simplicity denoted without time $t$ {and customer segment $g$ indices}:
\begin{equation}\label{eq:onlinedecprob}
    \max \sum_{k\in\mathcal{O}} \left(r+a^{pricing}_k - C^{true}_k\right) \mathbb{P}_k(\vv{a}^{pricing}).
\end{equation}

Equation~\ref{eq:onlinedecprob} maximizes the profits and is concave in the customer selection probabilities as shown by Theorem~1 in \citet{dong2009}. We apply their result to obtain the following concave optimization problem:
\begin{equation}
    \max \sum_{k\in\mathcal{O}} \left(-\frac{1}{\beta^d}(-\beta^k\mathrm{exp}\left[{d_{0,k}}\right]-\textrm{ln}\mathbb{P}_k)+r-C^{true}_k\right)\mathbb{P}_k.
\end{equation}

Note that $\sum_{k\in\mathcal{O}}\mathbb{P}_k=1$ and we assume $\beta^d<0$. Inspired by the model in \citet{yang2016}, we can obtain the optimal prices for each delivery choice $k$:
\begin{lemma}
    \textit{$a^{*{pricing}}_k= C^{true}_k-r-\frac{m}{\beta^d}, \quad \forall k \in \mathcal{O}$},
\end{lemma}
where $a^{*{pricing}}_k$ denotes the optimal price for delivery option $k$ and $m$ is the unique solution to:
\begin{equation}
    (m-1)\mathrm{exp}(m) = {\sum_{k\in\mathcal{O}} \mathrm{exp}\left[-\beta^k \mathrm{exp}\left[{d_{0,k}}\right]+\beta^d\left(C_k^{true}-r\right)\right]},
\end{equation}
which we approximate using the Lambert $W_0$-function \citep{Corless1996}. For all proofs, we refer to \citet{dong2009}. Since in reality, we do not know the cost $C_k^{true}$ of inserting a location {at the moment we need to make a pricing decision}, we replace $C_k^{true}$ by our approximation $\hat{C}^{DSPO}_k$.

\subsection{Training the Pricing Policy}\label{sect:training}

Algorithm~\ref{algo1} provides a high-level overview of the DSPO training procedure. The main part of the algorithm involves a simulation over $I$ booking horizons. During this simulation, we employ the DSPO policy $\pi$, and store feature values and the accompanying observed costs. We start by obtaining a data set to train an initial neural network $\mathcal{N}_\theta$. This initial data set is obtained by running simulations without a pricing policy, i.e., no incentives are provided for delivery options. After the initial training phase (1), we start a simulation procedure in which we simulate a horizon until cutoff time $T$ (3). At each timestep, we employ the DSPO pipeline: first, we obtain $a^{selection}$ of the selection heuristic policy $\pi^{selection}$ (4). Next, we obtain the pricing decision $a^{pricing}$ from the neural network, $\pi^{pricing}$ (5). Finally, both sub decisions are combined in a single decision $a(s_t)$ (6). The decision is applied to the state, {the delivery choice of the customer is recorded}, and the next state, {with the arrival of a new customer}, is observed (7). Next, the encoded state $\phi(s_t)$ is stored in a memory buffer (8). At the end of a simulation horizon, we can calculate the true costs $C^{true}_{k,t}$ related to each insertion (9). {In experiments with finite capacity OOH locations, we add a small penalty to the cost function to steer the policy, see Appendix~\ref{appendix:dspo]}. } The neural network $\mathcal{N}_\theta$ is trained with the Adam algorithm \citep{kingma} (10). The process of data collection and updating continues until $I$ iterations are reached.

\begin{algorithm}[htbp]
\linespread{1}\selectfont
\caption{Training Algorithm for Dynamic Selection and Pricing of Out-of-home delivery}\label{algo1}
\begin{algorithmic}[1]

    \State Initial training phase of $\mathcal{N}_\theta$
    \For{$i= 1,2,\ldots,I$ episodes}
        \For{$t = 1,2,\ldots,T$}
            \State $a^{selection}(s_t) \gets \pi^{selection}(s_t)$
            \State $a^{pricing}(s_t) \gets \pi^{pricing}_{\theta}(a^{pricing}(s_t,a^{selection}(s_t))$
            \State $a(s_t) \gets (a^{selection}(s_t),a^{pricing}(s_t))$
            \State Apply $a(s_t)$ to state $s_t$ and transition to $s_{t+1}$
            \State Store encoded state $\phi(s_t)$
        \EndFor
        \State Obtain and store the true costs of insertion $C^{true}_{k,t}$ and relate them to $\phi(s_t)$
        \State Update $\mathcal{N}_\theta$ and using the batch of tuples ($\mathbf{S},\mathbf{C^{true}}$)
    \EndFor
\end{algorithmic}
\end{algorithm}

\subsection{Design Choices}\label{sect:discussionChoice}

A few technical comments about the design choices of our DSPO pipeline are in order. First, we base the selection policy $\pi^{selection}$ on a simple heuristic rule. We argue that it is better to keep customer service high and influence customer behavior by pricing, compared to limiting the offering to a smaller subset of delivery options \cite[cf.][]{ASDEMIR2009246}. Therefore, we select a sufficiently large subset of delivery locations of size $N$. Alternatively, our machine learning model could also be used to limit the offering to customers, e.g., by only offering the OOH locations that yield the lowest predicted costs. Nevertheless, we decided to use the heuristic rule as it {reduces the required computations} and enhances learning stability by inducing less noise in the collected data.

Our CNN architecture is relatively small compared to state-of-the-art CNNs. We found that this size of the CNN architecture allows for fast inference in a sequential training loop and requires less data to train. We noticed that larger-sized CNNs did not noticeably improve the predictive performance of DSPO.

Our proposed Algorithm~\ref{algo1} incorporates a repeating sequence of simulation steps, data collection, and neural network updates. Hence, our implementation technically resembles the structure of \emph{reinforcement learning} algorithms. The prediction model $\mathcal{N}_\theta$ is generic in the sense that it can be trained by any other algorithm, e.g., Deep $Q$-networks \citep{MnihEtAl2013}, Deterministic Policy Gradient \citep{silver2014}, and proximal policy optimization \citep{SchulmanEtAl2017}. {We benchmark our proposed method against the latter.}

\section{Results}\label{section:numericalStudy}

In this section, we discuss and analyze the performance of our policy and the benchmarks. This section is structured as follows. First, we discuss the experimental design in Section~\ref{sect:expdesign}. Next, we study a small-scale synthetic case in Section~\ref{subsection:StylCase}, and show the results for a real case based on Amazon delivery data from the city of Seattle, USA, in Section~\ref{subsection:amaCase}. {The synthetic case is designed to study fundamental problem characteristics, but it lacks the detailed structure found in real-world applications, such as those we examine in the Amazon case. }

\subsection{Experimental Design}\label{sect:expdesign}

We compare the DSPO policy with the following baselines and benchmark policies: 
\begin{itemize}
    \item NoOOH: the situation with only home deliveries.
    \item OnlyOOH: the situation in which customers are only offered OOH delivery.
    \item NoPricing: the situation where customers can select OOH delivery if they want, but do not obtain a discount or pay a delivery charge related to their delivery choice.
    \item StaticPricing: the baseline that provides customers a fixed discount for choosing OOH delivery and a fixed charge for choosing home delivery.
    \item Hindsight benchmark: the heuristic benchmark introduced by \citet{yang2016} for time slot pricing and adapted to the OOH context.
    \item Foresight benchmark: the improved  state-of-the-art variant of the Hindsight benchmark, also proposed by \citet{yang2016}, and adapted to the OOH context.
    \item Linear benchmark: instead of using the CNN, we use a linear regression model.
    \item PPO: the state-of-the-art reinforcement learning actor-critic method \citep{SchulmanEtAl2017} that directly outputs prices for all delivery options.
\end{itemize}
The NoOOH and {OnlyOOH} baselines can be considered an {approximate} upper and lower bound on costs, {without costs or revenue from pricing}. {For all methods, we select only those OOH locations that still have remaining capacity. We set the number of offered OOH locations to $N=20$.} Note that we use $\pi^{selection}$ for NoPricing to offer a limited set of OOH delivery locations. The StaticPricing benchmark can be considered the current situation for many retailers that offer OOH delivery and provide a static discount. The {Hindsight} benchmark estimates the costs ($\hat{C}^{Hindsight}_{k,t}$) of inserting a location $k$ with a cheapest insertion into {a} preliminary route plan ${R}_{t-1}$. {For this, we need to add a preliminary route plan $R_{t-1}$ to the state, see Appendix~\ref{appendix:hindforsight} for more details.} The Foresight benchmark calculates a weighted costs term ($\hat{C}^{Foresight}_{k,t}$) of (i) the insertion costs in ${R}_{t-1}$ and (ii) the average insertion costs in a pool of {$10$} historic final routes $R_T \in \mathcal{R}_T$:  $\hat{C}^{Foresight}_{k,t} = (1-\theta^H_t) \hat{C}^{Hindsight}_{k,t} + \theta^H_t\frac{1}{|\mathcal{P}|}\sum_{p\in\mathcal{P}}f(k,p)$, 
where $f(k,p)$ represents the insertion costs of feasibly inserting delivery location $k$ into historic route $p$, and $\theta^H_t$ is the weight. This weight decreases during the booking horizon, assuming that $\hat{C}^{Hindsight}_{k,t}$ will become more accurate over time. For PPO, we use a Gaussian policy that outputs prices directly. For all pricing policies, the pricing decision space is discretized by rounding prices to two decimals, i.e., the decision space size for a single delivery option $k$ is equal to $100(b-a)$. Hence, the complete decision over all options $\mathcal{K}$ is a vector of discrete valued numbers. {In practice, one could decide on different rounding schemes based on specific requirements and constraints.} We restrict prices to $[-10,2]$, i.e., a maximum discount of $10$ and maximum delivery charge of $2$ can be attributed to a delivery option. For the implementation details of the benchmarks, we refer to Appendix~\ref{appendix::benchmarks}.

The problem and solution methods are implemented in Python 3.10, using PyTorch 2.0.1 \citep{pytorch}. All vehicle routes are obtained using HGS-CVRP \citep{hgs1,hgs2}, with the Hygese 0.0.0.8 Python wrapper. Computations were conducted on one thin CPU node of a high-performance cluster. The node is equipped with a $2.6$ GHz AMD Rome 7H12 processor, has $128$ CPU cores, and $160$ GB of memory. All results are reported over a separate test set. For the synthetic case, the reported results correspond to averages over $30$ seeds. For the Amazon case, we report results over $20$ seeds.

\subsection{Synthetic Case}\label{subsection:StylCase}

We start with a study of a relatively small synthetic case to gain more structural insights into the solution and analyze and validate the performance of the proposed policies and benchmarks. First, we introduce the relevant settings in Section~\ref{synth:probset}. Next, we provide insights into the cost factors in Section~\ref{sect:resultchoicemodel}. {We conduct a sensitivity analysis for multiple customer segments in Section~\ref{sect:new_cust_sens},} and compare DSPO with the baselines and benchmarks in Section~\ref{compareDSPOsynth}, {provide insights into the influence of capacities on performance and decision-making in Section~\ref{sect:capacitysynth}}, study the decision-making of DSPO when the cost distribution changes in Section~\ref{sensDSPOsynth}, and end with an ablation study in Section~\ref{ablation}.

\subsubsection{Problem Setting}\label{synth:probset}

We utilize the Gehring and Homberger benchmark problems for data generation \citep{Gehring2002}. Specifically, we employ the random (R), clustered (C), and random-clustered (RC) instances and partition the $200$ location instances in a test and training set. Next, we randomly select $|\mathcal{L}|=10$ locations, which we label as OOH delivery locations {with infinite capacity}. The instances are illustrated in Appendix~\ref{appendix:problsettings}. During data collection and evaluation, we randomly sample customer's home locations from the respective data set. Driving times are based on Euclidean distances and we assume a fixed vehicle speed of $30$ distance units per hour. The salary costs $C^w$ are $30$ per hour and the fuel costs $C^f$ are $0.3$ per distance unit. In our experiments, we model non-uniform service durations based on spatial information. This closely resembles reality, as service durations in, for instance, apartment buildings, are expected to be higher compared to rural areas or suburbs. Service times at location $i$ are obtained by projecting the service area onto the domain $(x \in [-3,3], y \in [-2,2])$ of:
\begin{equation}\label{eq:camel}
    l_i(x,y) = (4-2.1x^2+ \frac{y}{3}) x^2+ xy + (-4+4x^2) y^2,
\end{equation}
which is an optimization function proposed in \citet{sixhump}, see Appendix~\ref{appendix:problsettings} for a visualization of this function. We bound the service duration on $l_i\in [1,10]$ minutes. The probability of delivery failure at a home address is given by $\mathbb{P}^m=0.1$, with fixed costs of $C^m=10$. We draw the number of arriving customers on a day from the negative binomial distribution ${D}$, where the probability mass function is given by:
\begin{equation}
    \mathbb{P}(n;r,p) = \left(\frac{n+r-1}{n}  \right)(1-p)^np^r,
\end{equation}
where $r=90$, $p=0.5$ and $\mathbb{E}[{D}] = \frac{r(1-p)}{p} = 90$. {The interarrival time of customers is uniformly distributed.} The maximum number of customers served on a day is limited by the fleet capacity: we consider a limited fleet of $9$ vehicles with a maximum capacity of $10$ parcels per vehicle. {We assume a single parcel per customer. Since we do not have customer choice data, we use a MNL tuning procedure to obtain realistic choice behavior. The tuning procedure is further detailed in Appendix~\ref{appendix:mnl}. Note that we consider a single customer segment $g$ in our experiments.} All remaining problem parameter values are summarized in Appendix~\ref{appendix:problsettings}.

\subsubsection{Analysis of Cost Factors}\label{sect:resultchoicemodel}

\begin{figure}[hbtp]
\centering
\begin{subfigure}[t]{0.475\textwidth}
    \centering
   \begin{tikzpicture}[scale = 0.84]
            \begin{axis}[
                legend style={at={(0.5,-0.1)},anchor=north, draw=none},
                legend columns=2,
                ymin = 50 ,ymax=100.1,
                xmin=0, xmax=1,
                 xlabel = Percentage home delivery,
                xlabel near ticks,
                ylabel= Costs (scaled),
                ylabel near ticks,
                  xticklabel={\pgfmathparse{\tick*100}\pgfmathprintnumber{\pgfmathresult}\%},
                  xticklabel shift={0.1cm},
                  legend cell align={left}
                        ]

            \addplot[name path = A,draw=none,forget plot] %
                coordinates {
                  (0.001,0.01)
                (0.999,0.01)

                };

                 \addplot[name path = B,draw=none,area legend,forget plot] %
                coordinates {
                  (0,70.1989158690006)
                (0.106367041198501,73.0435183974276)
                (0.208988764044943,74.625795771875)
                (0.291385767790262,74.7883944412934)
                (0.410486891385767,75.4755049475452)
                (0.511235955056179,76.3636783245834)
                (0.611235955056179,76.8794482974695)
                (0.714232209737827,76.8619645695752)
                (0.814981273408239,77.0840079138346)
                (0.893258426966292,76.8724548063119)
                (0.999625468164794,76.8962119055795)

                };

             \addplot[name path = C,draw=none,area legend,forget plot] %
                coordinates {
                         (0,70.690645716032)
            (0.106367041198501,75.662913026846)
            (0.208988764044943,79.2979374096527)
            (0.291385767790262,81.1087271076808)
            (0.410486891385767,84.1782228280139)
            (0.511235955056179,87.0816843263975)
            (0.611235955056179,89.5977588658236)
            (0.714232209737827,91.6405139236913)
            (0.814981273408239,93.8778453892962)
            (0.893258426966292,95.2320737589526)
            (0.999625468164794,97.0834956406073)

                };

                             \addplot[name path = D,draw=none,area legend,forget plot] %
                coordinates {
                (0,70.690645716032)
                (0.106367041198501,75.9732491969725)
                (0.208988764044943,79.9076824199716)
                (0.291385767790262,81.9588733765483)
                (0.410486891385767,85.3758581887836)
                (0.511235955056179,88.5732648623927)
                (0.611235955056179,91.3810991110573)
                (0.714232209737827,93.7243557421109)
                (0.814981273408239,96.2556323829412)
                (0.893258426966292,97.8382419482189)
                (0.999625468164794,100)

                };

        \addplot [black!65] fill between [of = A and B];
        \addplot [black!20] fill between [of = C and D];
        \addplot [black!45] fill between [of = B and C];
                
              \addlegendentry{Travel costs}
              \addlegendentry{Delivery fail. costs}
             \addlegendentry{Service costs}

            \end{axis}
                        \node at (3.5,7.0) {Cost distribution and OOH share};
    \end{tikzpicture}

\end{subfigure}
\begin{subfigure}[t]{0.475\textwidth}
    \centering
   \begin{tikzpicture}[scale = 0.84]
            \begin{axis}[
                legend style={at={(0.5,-0.1)},anchor=north, draw=none},
                legend columns=2,
                ymin = 50 ,ymax=100.1,
                xmin=0, xmax=10,
                 xlabel = Fixed OOH discount as $\%$ of revenue,
                xlabel near ticks,
                ylabel= Costs (scaled),
                ylabel near ticks,
                  xticklabel shift={0.1cm},
                  legend cell align={left},
              xticklabel={\pgfmathparse{\tick*2}\pgfmathprintnumber{\pgfmathresult}\%},
                  xticklabel shift={0.1cm},
                  legend cell align={left},
                        ]

            \addplot[name path = A,draw=none,forget plot] %
                coordinates {
                  (0.001,0.01)
                (10,0.01)

                };

                 \addplot[name path = B,draw=none,area legend,forget plot] %
                coordinates {
                (0,96.8570363389598)
                (1,94.4431072265369)
                (2,92.4328153739355)
                (3,91.7247274791541)
                (4,90.5968477116407)
                (5,89.3235183707887)
                (6,87.3415563810481)
                (7,85.2017880937023)
                (8,83.3667947665263)
                (9,82.1682311535212)
                (10,80.4355627612331)

                };

             \addplot[name path = C,draw=none,area legend,forget plot] %
                coordinates {
                       (0,96.8570363389598)
                    (1,95.0803675708954)
                    (2,93.6317388505804)
                    (3,94.388148918396)
                    (4,94.7014579296626)
                    (5,95.2386015671419)
                    (6,95.5180967994325)
                    (7,96.2648858665634)
                    (8,96.9965407983443)
                    (9,98.6977840856508)
                    (10,100)

                };

        \addplot [black!55] fill between [of = A and B];
        \addplot [black!25] fill between [of = B and C];
        \addplot[  only marks,black,mark=star,color=black,mark size=2.5pt]%
        coordinates{(2,93.6317388505804)};

              \addlegendentry{Operational costs}
              \addlegendentry{Discount costs}
                          \addlegendentry{Min. costs}

            \end{axis}
                        \node at (3.5,7.0) {Cost distribution and discounts};
    \end{tikzpicture}

\end{subfigure}

 \caption{Sensitivity analysis of costs and discounts, results based on the RC instances.}\label{stylized:costdist}
\end{figure}
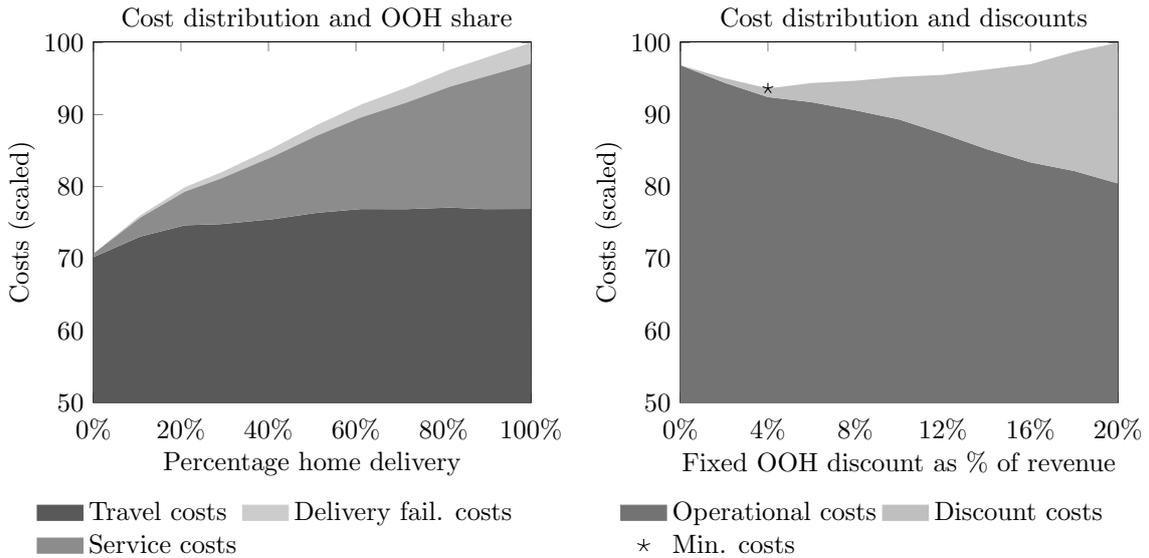

Figure~\ref{stylized:costdist} shows a sensitivity analysis of the cost distribution for the RC instances. The left figure depicts the total costs ({min-max} scaled {for visualization purpose}) divided into the categories: (i) travel costs, (ii) service costs, and (iii) delivery failure costs compared to the share of OOH deliveries. Travel costs are the costs related to fuel and salary ($C^f$ and $C^w$) during travel, service costs are the salary costs during service time $l_i$, and failure costs are the costs paid for delivery failures ($C^m$). The graph is obtained by using the NoPricing benchmark and tuning the MNL parameters to obtain different home delivery rates. We observe that travel costs make up the largest portion of the costs, but are less sensitive to the share of OOH deliveries. The travel costs only slightly increase from $0\%$ home delivery to $20\%$ home delivery, but thereafter stay relatively stable. All areas are visited by vehicles, so the travel distances do not significantly increase as the number of stops in the area increases with more home deliveries. The service and delivery failure costs are elastic, i.e., as more customers select home delivery, these costs increase significantly. 

\begin{observation}
    \textit{Travel, service, and delivery failure costs rise with home deliveries, OOH delivery mainly impacts costs through reduced service durations and delivery failures.}
\end{observation}

The right graph in Figure~\ref{stylized:costdist} illustrates the total operational costs, including travel, service, and delivery failure expenses, along with the costs incurred from offering fixed discounts, in situations where no delivery charge is applied. The graph shows that (i) providing discounts can significantly reduce operational costs, and (ii) a balance needs to be struck when offering discounts, making sure they do not harm the overall profitability. The concept of finding a balance aligns with the findings from \citet{LIN2022102541}, who examine the optimal number of OOH locations to open. Overall, the total costs seem approximately convex in the discount percentage and the the minimum of costs lies at $4\%$ fixed discounts as a percentage of revenue per customer $r$. {Considering the approximate convexity,} we note that the determined discount value is optimal within the fixed discount policy, and does not imply a globally optimal policy.  

\begin{observation}
    \textit{Offering discounts can substantially reduce operational costs, but they might compromise overall profitability.}
\end{observation}

\begin{figure}[!hbtp]
    \centering
   \begin{tikzpicture}
  \begin{axis} [
         boxplot/draw direction=y,
    /pgfplots/boxplot/box extend=0.4,
     xtick={1,4,7},
    xticklabels={R, C, RC},
     ylabel = Travel costs (scaled),
                ylabel near ticks,
    cycle list={{black!25},{black!55},{black!85}},
    every axis plot/.append style={fill,fill opacity=0.75},
     x tick label style={
        text width=2.5cm,
        align=center
    },
    ymax=80,ymin=55,
      x=1.5cm,
       legend style={at={(0.5,-0.1)},draw=none,anchor=north},
                legend columns=3,
    ]
       \addplot+ [
        area legend, boxplot prepared={
            upper quartile=85.9725685785536 / 1.297,
            lower quartile=81.0317955112219 / 1.297,
            upper whisker=92.3940149625935 / 1.297,
            median=83.3852867830424 / 1.297,
            lower whisker=74.8129675810474 / 1.297,
            draw position=0.5
        },
    ] coordinates {};

    \addplot+ [
        area legend, boxplot prepared={
            upper quartile=94.4513715710723 / 1.297,
            lower quartile=90.928927680798 / 1.297,
            upper whisker=100 / 1.297,
            median=91.8640897755611 / 1.297,
            lower whisker=86.284289276808 / 1.297,
            draw position=1.0
        },
    ] coordinates {};

    \addplot+ [
        area legend, boxplot prepared={
            upper quartile=93.6720698254364 / 1.297,
            lower quartile=89.2768079800499 / 1.297,
            upper whisker=97.1945137157107 / 1.297,
            median=92.1134663341646 / 1.297,
            lower whisker=85.6608478802993 / 1.297,
            draw position=1.5
        },
    ] coordinates {};

    \addplot+ [
        boxplot prepared={
            upper quartile=86.6427680798005 / 1.297,
            lower quartile=80.7356608478803 / 1.297,
            upper whisker=91.7082294264339 / 1.297,
            median=83.2917705735661 / 1.297,
            lower whisker=72.9471674092934 / 1.297,
            draw position=3.5
        },
    ] coordinates {};

    \addplot+ [
        boxplot prepared={
            upper quartile=92.4563591022444 / 1.297,
            lower quartile=87.5779301745636 / 1.297,
            upper whisker=97.9426433915212 / 1.297,
            median=90.71072319202 / 1.297,
            lower whisker=84.7231063017187 / 1.297,
            draw position=4.0
        },
    ] coordinates {};

    \addplot+ [
        boxplot prepared={
            upper quartile=92.7369077306733 / 1.297,
            lower quartile=89.4482543640898 / 1.297,
            upper whisker=95.4488778054863 / 1.297,
            median=91.1159600997506 / 1.297,
            lower whisker=84.28927680798 / 1.297,
            draw position=4.5
        },
    ] coordinates {};

    \addplot+ [
        boxplot prepared={
            upper quartile=89.9725685785536 / 1.297,
            lower quartile=83.0317955112219 / 1.297,
            upper whisker=95.3940149625935 / 1.297,
            median=88.3852867830424 / 1.297,
            lower whisker=73.8129675810474 / 1.297,
            draw position=6.5
        },
    ] coordinates {};

    \addplot+ [
        boxplot prepared={
            upper quartile=94.4513715710723 / 1.297,
            lower quartile=90.928927680798 / 1.297,
            upper whisker=100 / 1.297,
            median=91.8640897755611 / 1.297,
            lower whisker=86.284289276808 / 1.297,
            draw position=7
        },
    ] coordinates {};

    \addplot+ [
        boxplot prepared={
            upper quartile=93.6720698254364 / 1.297,
            lower quartile=89.2768079800499 / 1.297,
            upper whisker=97.1945137157107 / 1.297,
            median=92.1134663341646 / 1.297,
            lower whisker=85.6608478802993 / 1.297,
            draw position=7.5
        },
    ] coordinates {};

\addlegendentry{$0\%$ home delivery}
\addlegendentry{$50\%$ home delivery}
\addlegendentry{$100\%$ home delivery}

  \end{axis}
\end{tikzpicture}
    \caption{Travel costs for different spatial distributions and OOH shares, given $\mathbf{C^f=2.0}$, over 30 replications.}
    \label{fig:boxplot_travelcosts}
\end{figure}
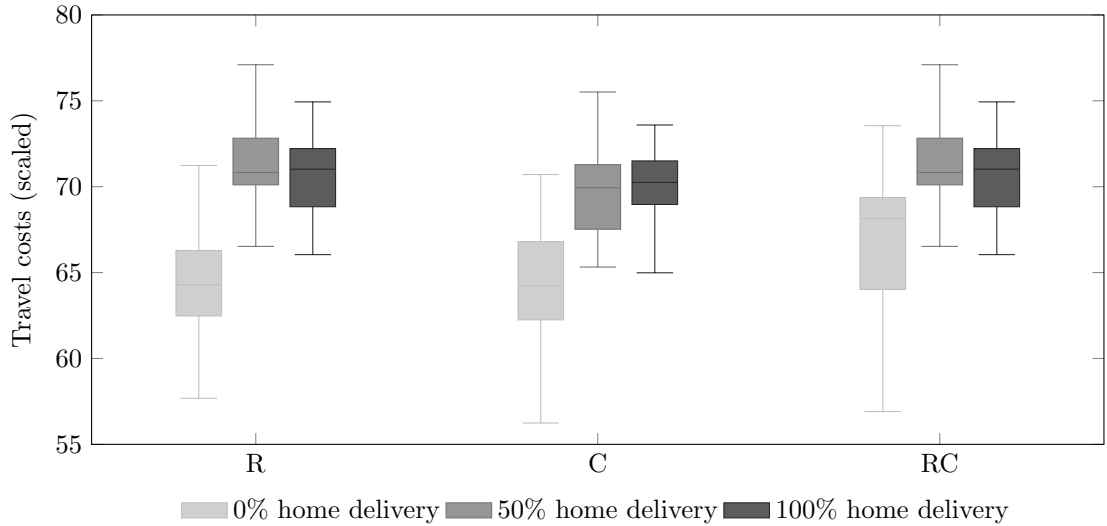

Next, we study the effect on travel costs when using the random (R), clustered (C), and RC instances of \citet{Gehring2002}. Figure~\ref{fig:boxplot_travelcosts} shows boxplots for the R, C, and RC instances (left to right). For each instance type, we show the effect of $0\%$, $50\%$, and $100\%$ of the customers choosing for home instead of OOH delivery. The results confirm the observations from Figure~\ref{stylized:costdist}: the travel costs increase from $0\%$ to $50\%$, but thereafter the costs hardly increase. We do not find significant differences in travel costs between the R, C, and RC instances. Therefore, for the remaining experiments, we will focus solely on the RC instances.

\begin{observation}
    \textit{Only offering OOH delivery (no home deliveries) leads to significant travel cost savings across spatial dispersion types (Random, Clustered, and Random Clustered).}
\end{observation}

 \vspace{-1em}
\subsubsection{{Analysis of Customer Segments}}\label{sect:new_cust_sens}

{In our experiments, we so far considered a single customer segment $g$. This means that all customers are willing to accept OOH delivery, which might not be realistic. When more customer choice data becomes available for research, one may define distinctive customer segments, like we already modeled in Section~\ref{section:custChoice}. In this section, we study the effects of modeling three customers segments: customers who (i) only consider home delivery and are non-sensitive to incentives, (ii) consider both OOH and home delivery and are somewhat sensitive to both the incentives ($\beta^d$) and the required distance to travel to an OOH location ($\beta^k$), and (iii) consider both OOH and home delivery and are highly sensitive to incentives ($\beta^d$) and less sensitive to distance ($\beta^k$). We define the customer choice model parameters per segment in Appendix~\ref{appendix:mnl}. Figure~\ref{fig:sensitivityMNL_3segments} shows a sensitivity analysis of the shares of the different customer segments. We vary the parameter $\mu_1$, $\mu_2$, and $\mu_3$, which define the probability that an arrived customer is from segment 1, 2, or 3, respectively. The different lines in the graph represent different levels of $\mu_1$, and the x-axis of the left and right graph define the value for $\mu_2$. The value of $\mu_3$ is implicit, i.e., $\mu_3 = 1.0-\mu_1-\mu_2$. The results are shown for the RC instance, using the StaticPricing benchmark, and infinite OOH capacity. Note that we defined different static pricing levels for each setting, aimed at finding the lowest costs during simulation. On the left, we show the percentage of customers choosing home delivery. This follows a logical pattern, as a larger share of segment 1 customers indicates more home deliveries, and, vice versa for segment~1 and~2. The right graph shows the total costs (operational + pricing costs) for the different values of $\mu_1$, $\mu_2$, and $\mu_3$. We observe that a higher share of customers that select OOH delivery results in lower costs. Moreover, the cost reduction seems to show only a slightly increasing effect in the OOH deliveries, i.e.,  as a higher share of customers choosing home delivery ($\mu_1$ increase) also results in higher costs. We conjecture that this effect is caused by the economies of scale by bundling more deliveries into OOH locations. Considering the share of segment 2 and 3 customers, we see that it is easier to nudge segment 3 customers to OOH delivery, resulting in more OOH deliveries and lower costs from pricing. 
\begin{observation}
\textit{The delivery costs decreases as the percentage of customers choosing OOH delivery rises.}
\end{observation}
For the remainder of experiments, we use a single customer segment again and rely on our tuning procedure detailed in Appendix~\ref{appendix:mnl}.}

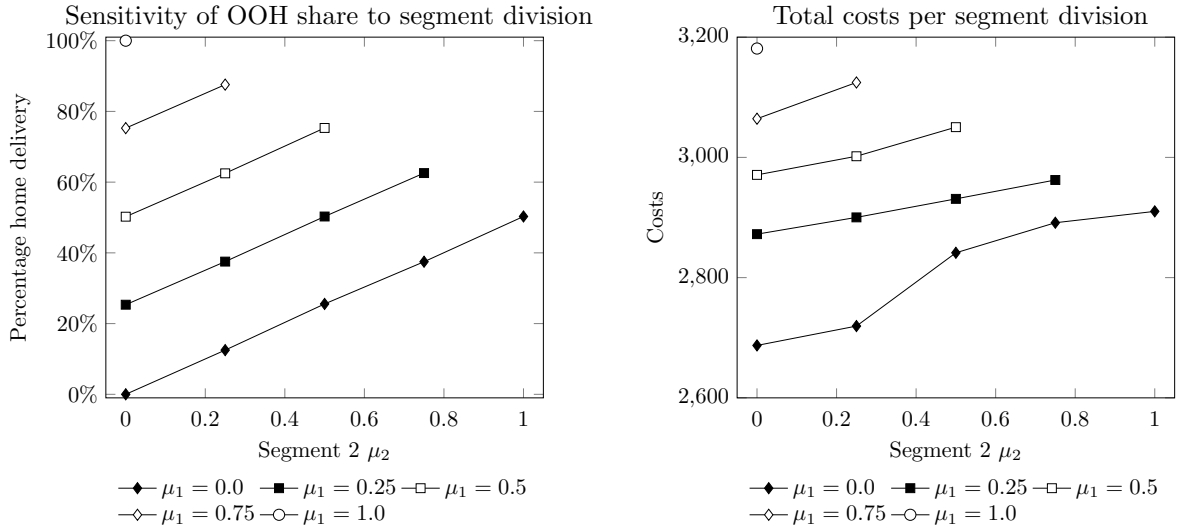
\begin{figure}[hbtp]
\centering
\begin{subfigure}{.5\textwidth}
     \centering
    \centering
    \begin{tikzpicture}[scale = 0.84]
            \begin{axis}[
                legend style={at={(0.5,-0.2)},anchor=north, draw=none},
                legend columns=3,
                ymin = -1 ,ymax=101,
                xmin=-0.05, xmax=1.05,
                 xlabel = Segment 2 $\mu_2$,
                xlabel near ticks,
                ylabel= Percentage home delivery,
                ylabel near ticks,
                legend cell align={left},
                  yticklabel={$\pgfmathprintnumber{\tick}\%$},
                  xticklabel shift={0.1cm},
                  legend cell align={left},
                        ]

        \addplot[ mark options={fill=black, mark size=2.5pt},mark=diamond*] %
                coordinates {
                    (0,0)
                    (0.25,12.50)
                    (0.5,25.557)
                    (0.75,37.49)
                    (1,50.29)

                };

 \addplot[mark options={fill=black}, mark=square*] %
                coordinates {
                (0,25.34)
                (0.25,37.53)
                (0.5,50.28)
                (0.75,62.56)

                };

                 \addplot[mark options={fill=white}, mark=square*] %
                coordinates {
                    (0,50.23)
                    (0.25,62.49)
                    (0.5,75.32)

                };

                 \addplot[mark options={mark size=2.5pt,fill=white}, mark=diamond*] %
                coordinates {
                    (0,75.27)
                    (0.25,87.57)

                };

                \addplot[mark options={mark size=2.5pt,fill=white}, mark=*] %
                coordinates {
                    (0,100)

                };

                \addlegendentry{$\mu_1=0.0$}
                \addlegendentry{$\mu_1=0.25$}
                \addlegendentry{$\mu_1=0.5$}
                \addlegendentry{$\mu_1=0.75$}
                \addlegendentry{$\mu_1=1.0$}

            \end{axis}
            \node at (3.5,6.0) {Sensitivity of OOH share to segment division};
    \end{tikzpicture}

\end{subfigure}%
\begin{subfigure}{.5\textwidth}
    \centering
    \begin{tikzpicture}[scale = 0.84]
            \begin{axis}[
                legend style={at={(0.5,-0.2)},anchor=north, draw=none},
                legend columns=3,
                ymin = 2600 ,ymax=3200,
                xmin=-0.05, xmax=1.05,
                 xlabel = Segment 2 $\mu_2$,
                xlabel near ticks,
                ylabel= Costs,
                ylabel near ticks,
                legend cell align={left},
                  xticklabel shift={0.1cm},
                  legend cell align={left},
                        ]

      \addplot[ mark options={fill=black, mark size=2.5pt},mark=diamond*] %
                coordinates {
                    (0,2687.2)
                    (0.25,2719.4)
                    (0.5,2841.3)
                    (0.75,2891.3)
                    (1,2910.2)
                };

 \addplot[mark options={fill=black}, mark=square*] %
                coordinates {
                (0,2872.3)
                (0.25,2900.2)
                (0.5,2931.0)
                (0.75,2962.4)

                };

                 \addplot[mark options={fill=white}, mark=square*] %
                coordinates {
                    (0,2970.9)
                    (0.25,3001.9)
                    (0.5,3050.3)

                };

                 \addplot[mark options={mark size=2.5pt,fill=white}, mark=diamond*] %
                coordinates {
                    (0,3064.3)
                    (0.25,3124.6)

                };

                \addplot[mark options={mark size=2.5pt,fill=white}, mark=*] %
                coordinates {
                    (0,3181.09)

                };

                \addlegendentry{$\mu_1=0.0$}
                \addlegendentry{$\mu_1=0.25$}
                \addlegendentry{$\mu_1=0.5$}
                \addlegendentry{$\mu_1=0.75$}
                \addlegendentry{$\mu_1=1.0$}

            \end{axis}
            \node at (3.5,6.0) {Total costs per segment division};
    \end{tikzpicture}

\end{subfigure}
  \caption{Analysis of the customer segment division, results based on the RC instances using StaticPricing.}\label{fig:sensitivityMNL_3segments}
\end{figure}

\subsubsection{Comparison of DSPO with Benchmarks}\label{compareDSPOsynth}

Table~\ref{tab:resultsSynth} shows the results on the synthetic case for all benchmarks and DSPO. {We show the relative savings compared to the NoOOH baseline, and the $95\%$ confidence interval of these savings.} The difference in costs ($27.2\%$pt) between NoOOH and OnlyOOH indicates an {approximate} upper bound on the savings a policy can yield. Note, however, that OnlyOOH does not incur costs for providing discounts, as all other policies do. The StaticPricing benchmark provides a static discount ($5$) for choosing OOH delivery and a static delivery charge ($2$) for home delivery for each customer. These values were selected after an experimental evaluation aimed at finding the lowest costs during simulation. Even using StaticPricing, $7.9\%$ in total costs can be saved compared to not offering OOH and $3.9\%$pt in costs can be saved compared to the NoPricing benchmark. This shows that even StaticPricing can already save significantly in costs. The Hindsight benchmark does not convincingly beat the StaticPricing benchmark. The Hindsight benchmark overestimates the costs of delivery and provides too high discounts. The Hindsight benchmark is solely basing its pricing on the preliminary route plan ${R}_{t-1}$, which is especially inaccurate at the beginning of the booking horizon. The Foresight benchmark, as proposed by \citet{yang2016}, tries to overcome this by using a pool of historic routes $\mathcal{R}_T$ to determine costs. However, the accuracy of the prediction based on historical route data is constrained by variability in daily customer arrivals, which follow a negative binomial distribution ${D}$. As with the Hindsight benchmark, this also results in an overestimation of the costs, although the Foresight benchmark does outperform the StaticPricing benchmark, which is in line with the results obtained by \citet{yang2016}.

\begin{table}[t]
	\caption{Results on the synthetic case, reported on the RC instances.}
	\label{tab:resultsSynth}
	\begin{center}
 \resizebox{\textwidth}{!}{%
		\begin{tabular}{l c c c c c c c }
			\hline
			  & {\makecell{$\%$ Home \\ delivery}} &  {\makecell{Travel\\ costs}} & {\makecell{Service \\ costs}} & {\makecell{Delivery fail. \\ costs}}  & {\makecell{Discount costs \\ (Charge revenue)}}  &  {\makecell{Avg. discount \\ (Avg. charge)}} & $\%$ Savings (95$\%$ CI) \\
			\hline

			\makecell[l]{NoOOH} & $100\%$ & 2353.7 & 523.3 & 89.0 & - & - & - \\
			
			\makecell[l]{OnlyOOH} & $0\%$ & 2145.1 & 15 & 0.0 & - & - & $27.2\%$ ($\pm0.2\%$) \\

            \makecell[l]{NoPricing} & $81.7\%$ & 2346.1 & 427.5 & 72.7 & - & - & $4.0\%$ ($\pm0.1\%$)  \\

             \makecell[l]{StaticPricing} & $59.1\%$ & 2331.3 & 296.9 & 50.5 & 159.3 (107.0) & $5 \pm 0$ ( $2 \pm 0$) & $7.9\%$ ($\pm0.2\%$) \\

             \makecell[l]{Hindsight benchmark} &  $40.9\%$ & 2257.5 & 205.2 & 34.9 & 386.0 (87.3) &  $9.7 \pm 1.5$ ( $2.0 \pm 0.1$) & $5.7\%$ ($\pm0.3\%$)\\

             \makecell[l]{Foresight benchmark} & $45.6\%$ & 2238.5 & 229.1 & 39.0 & 320.3 (104.5)  & $9.4 \pm 2.0$ ( $2.0 \pm 0.2$) & $8.2\%$ ($\pm0.2\%$) \\

             \makecell[l]{Linear benchmark} & $70.4\%$ & 2265.7 & 354.8 & 60.3 & 190.4 (95.8) & $1.8 \pm 0.4$  ($5.6 \pm 2.8$) & $6.4\%$ ($\pm0.3\%$)\\
             
            \makecell[l]{PPO benchmark} & $61.7\%$ & 2329.7 & 308.5 & 52.5 & 172.5 (52.7) & $3.6 \pm 2.5$ ( $1.4 \pm 0.7$) & $5.2\%$ ($\pm0.2\%$) \\

             \makecell[l]{DSPO} & $70.4\%$ & 2196.2 & 341.0 & 58.0 & 192.4 (84.3) & $5.2 \pm 3.7$ ($1.8 \pm 0.5$) & $8.8\%$ ($\pm0.2\%$) \\

			\hline
		\end{tabular}}
	\end{center}
\end{table}

\begin{observation}
    \textit{Cheapest insertion as a cost predictor (Hindsight benchmark) overestimates home delivery costs, leading to excessive discounts. This overestimation is only partially compensated by incorporating historic routes (Foresight benchmark).}
\end{observation}

Next, we compare three learning approaches: the linear benchmark, PPO, and DSPO. We observe that the linear benchmark is unable to differentiate between customers in terms of costs, which is reflected by the low deviation in discounts and charges. Interestingly, both PPO and DSPO find a policy that can differentiate between customers, as reflected by the deviation of discounts. However, DSPO outperforms PPO, mainly because of lower travel costs and slightly higher revenue from delivery charges. Even though the use of PPO results in fewer delivery stops, DSPO can recognize and nudge remote customers better to OOH options, which yields lower travel costs. We conjecture that PPO learns a policy that differentiates between customers but does not always provide discounts to the right customers. Note that PPO requires $1$ million episodes to learn a performant policy due to the sparse cost structure of our problem, see Appendix~\ref{appendix:benchmarkppo} for the convergence curve. The convergence curve of DSPO, which only requires \num[group-separator={,}]{100000} episodes, is depicted in Appendix~\ref{appendix:dspo]}.

\begin{observation}
    \textit{DSPO can differentiate between customers, identify the customers that are most expensive for home delivery, and adjust pricing accordingly.}
\end{observation}

\vspace{1em}

\subsubsection{{Analysis of Finite Capacity OOH}}\label{sect:capacitysynth}

{So far, we assumed infinite capacity of all OOH locations. However, in reality many OOH locations will have a limited capacity, see \citet{sethuraman}. In this section, we study the effect of finite capacity OOH locations. Furthermore, we study the effect of different capacities per OOH location.}

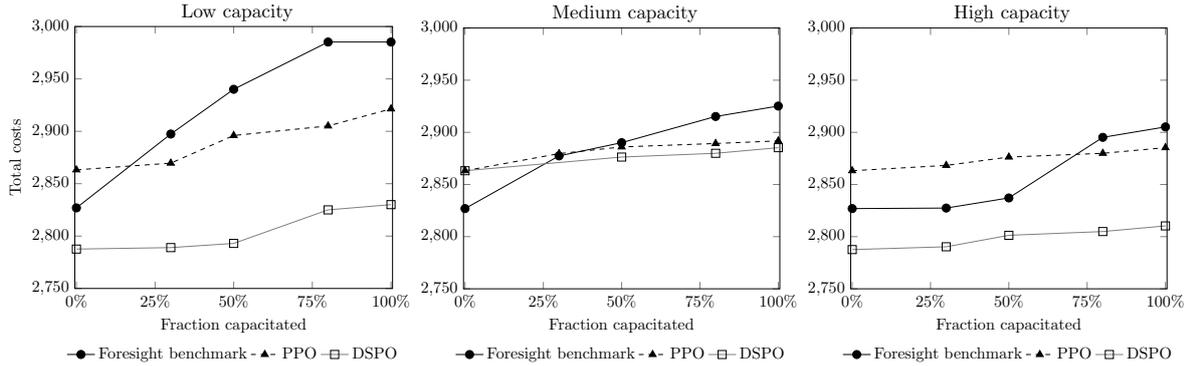
\begin{figure}[hbtp]
\begin{subfigure}[t]{.34\textwidth}
    \centering
    \begin{adjustbox}{width=\textwidth}
    \begin{tikzpicture}[scale = 0.84]
            \begin{axis}[
                legend style={at={(0.5,-0.2)},anchor=north, draw=none, legend columns=3},
                 xtick distance=25,
               ymin = 2750 ,ymax=3000,
                xmin=-0.5, xmax=100.5,
                 xlabel = Fraction capacitated,
                xlabel near ticks,
                ylabel style={align=center},
                ylabel= {Total costs},
                xticklabel={$\pgfmathprintnumber{\tick}\%$}
                        ]

        \addplot[ mark options={fill=black, mark size=2.5pt},mark=*] %
                coordinates {
                    (0,2826.9)
                     (30,2897.4)
                    (50,2940.1)
                    (80,2985.2)
                    (100,2985.2)
                };
                \addlegendentry{Foresight benchmark}

       \addplot[dashed, mark=triangle*, mark options={fill=black, mark size=2.5pt, solid}]%
                coordinates {
                   (0,2863.2)
                     (30,2869.6)
                    (50,2896)
                    (80,2905.1)
                    (100,2921.4)
                };
                \addlegendentry{PPO}

          \addplot[gray, mark options={black,solid,fill=black, mark size=2.5pt},mark=square] %
                coordinates {
                   (0,2787.6)
                     (30,2789)
                    (50,2793)
                    (80,2825)
                    (100,2830)
                };
                \addlegendentry{DSPO}
            \end{axis}
            \node at (3.5,6.0) {Low capacity };
    \end{tikzpicture}
\end{adjustbox}
\end{subfigure}%
\begin{subfigure}[t]{.32\textwidth}
     \centering
    \centering
    \begin{adjustbox}{width=\textwidth}
   \begin{tikzpicture}[scale = 0.84]
            \begin{axis}[
                 legend style={at={(0.5,-0.2)},anchor=north, draw=none, legend columns=3},
                 xtick distance=25,
                ymin = 2750 ,ymax=3000,
                xmin=-0.5, xmax=100.5,
                 xlabel = Fraction capacitated,
                xlabel near ticks,
                ylabel near ticks,
                 xticklabel={$\pgfmathprintnumber{\tick}\%$},
                        ]

        \addplot[ mark options={fill=black, mark size=2.5pt},mark=*] %
                coordinates {
                   (0,2826.9)
                     (30,2877.4)
                    (50,2890.1)
                    (80,2915.2)
                    (100,2925.2)
                };
                \addlegendentry{Foresight benchmark}

       \addplot[dashed, mark=triangle*, mark options={fill=black, mark size=2.5pt, solid}] %
                coordinates {
                   (0,2863.2)
                     (30,2879.6)
                    (50,2886)
                    (80,2889.3)
                    (100,2891.8)
                };
                \addlegendentry{PPO}

            \addplot[gray, mark options={black,solid,fill=black, mark size=2.5pt},mark=square] %
                coordinates {
                    (0,2863.2)
                     (30)
                    (50,2876.3)
                    (80,2879.9)
                    (100,2885.3)
                };
                \addlegendentry{DSPO}

            \end{axis}
             \node at (3.5,6.0) {Medium capacity};
    \end{tikzpicture}
    \end{adjustbox}
\end{subfigure}%
\begin{subfigure}[t]{.32\textwidth}
     \centering
    \centering
    \begin{adjustbox}{width=\textwidth}
   \begin{tikzpicture}[scale = 0.84]
            \begin{axis}[
                  legend style={at={(0.5,-0.2)},anchor=north, draw=none, legend columns=3},
                 xtick distance=25,
                ymin = 2750 ,ymax=3000,
                xmin=-0.5, xmax=100.5,
                xlabel = Fraction capacitated,
                xlabel near ticks,
                ylabel near ticks,
                 xticklabel={$\pgfmathprintnumber{\tick}\%$},
                        ]
\addplot[ mark options={fill=black, mark size=2.5pt},mark=*] %
                coordinates {
                   (0,2826.9)
                     (30,2827.4)
                    (50,2837.1)
                    (80,2895.2)
                    (100,2905.2)
                };
                \addlegendentry{Foresight benchmark}

       \addplot[dashed, mark=triangle*, mark options={fill=black, mark size=2.5pt, solid}] %
                coordinates {
                    (0,2863.2)
                     (30, 2868.3)
                    (50,2876.3)
                    (80,2879.9)
                    (100,2885.3)
                };
                \addlegendentry{PPO}

        \addplot[gray, mark options={black,solid,fill=black, mark size=2.5pt},mark=square] %
                coordinates {
                    (0,2787.6)
                     (30,2790.3)
                    (50,2801.3)
                    (80,2804.9)
                    (100,2810.3)
                };
                \addlegendentry{DSPO}

            \end{axis}

             \node at (3.5,6.0) {High capacity};
    \end{tikzpicture}

    \end{adjustbox}
\end{subfigure}
 \caption{The total costs incurred by Foresight benchmark, PPO, and DSPO for different fractions of capacitated OOH locations (x-axis) with low capacity (Left), medium capacity (Middle column) and high capacity (Right).}\label{fig:capacity_analysis}
\end{figure}

{First, we show the general effect of a capacitated system on the performance of the Foresight benchmark, PPO, and DSPO, see Figure~\ref{fig:capacity_analysis}. We randomly assign OOH locations to have limited capacity, varying the capacity of capacitated OOH locations between $\{2, 3, 5\}$, and the fraction of capacitated OOH locations varying between $\{0\%,30\%,50\%, 80\%, 100\%\}$. Note that DSPO and PPO are retrained for every setting. The Foresight benchmark is unable to cope with the finite capacity system, as it does not differentiate its pricing between OOH locations depending on remaining capacity. Hence, the costs for the situation with $100\%$ finite capacity OOH locations and low capacity (left) are higher than the NoOOH baseline, caused by the added costs of providing discounts. PPO, being a learning method that can utilize information considering the remaining OOH capacities, is better able to cope with the capacitated systems. In all settings, it incurs a slight increase in costs when the share of capacitated OOH locations is increased. DSPO has a similar pattern in cost increase across the experimental settings. 
Nonetheless, the overall advantage of DSPO compared to PPO remains consistent with the earlier results from Table~\ref{tab:resultsSynth}. We conjecture that DSPOs superior performance in a capacitated context is due to its underlying CNN architecture being better able to deal with this added complexity. DSPO effectively deters customers with lower delivery costs from choosing OOH delivery, encouraging them to opt for home delivery instead. This strategy preserves future capacity for more expensive customer deliveries, resulting in savings on routing costs. We establish that the finite capacity problem is inherently more challenging, as the pricing policy needs to adapt, e.g., by nudging a customer to a less preferred option as the most-preferred option has no remaining capacity.
\begin{observation}
\textit{Finite capacity at OOH locations increases costs, and DSPO consistently outperforms both the Foresight benchmark and PPO by better managing capacity.}
\end{observation}}

\begin{figure}[!ht]
\centering
\begin{subfigure}[t]{.35\textwidth}
    \centering
    \begin{adjustbox}{width=\textwidth}
    \begin{tikzpicture}[scale = 0.84]
            \begin{axis}[
               ymin = 0 ,ymax=100.5,
                xmin=-0.5, xmax=90,
                 xlabel = Booking horizon,
                xlabel near ticks,
                ylabel= Total remaining capacity,
                ylabel near ticks,
                yticklabel={$\pgfmathprintnumber{\tick}\%$}
                        ]

\addplot[ name path = A] %
                coordinates {
                   (0,100)
(1,100.094581628303)
(2,96.0973472892839)
(3,97.0770057714769)
(4,93.230451545609)
(5,94.0860913821033)
(6,93.9089719714684)
(7,90.6027149279)
(8,87.7853726717135)
(9,89.2759740125274)
(10,85.3182502081902)
(11,85.0444949889555)
(12,84.8461339773437)
(13,82.215875653071)
(14,81.4637350641583)
(15,79.5027380248646)
(16,80.0625220936506)
(17,76.7173735433648)
(18,76.9458912574507)
(19,73.6897913336828)
(20,72.6936564800459)
(21,70.5472491719241)
(22,71.3975599126256)
(23,68.6199695720202)
(24,68.5505077806201)
(25,67.424666306975)
(26,67.6194438542206)
(27,65.4235166323885)
(28,62.5826430032461)
(29,62.1020343127683)
(30,60.8774135606002)
(31,60.3233670760151)
(32,56.7335197636335)
(33,54.6298186924138)
(34,55.7835232067492)
(35,54.1088912024002)
(36,51.6013141484265)
(37,51.0564667712577)
(38,49.5598594090527)
(39,49.2032190487655)
(40,45.4292921392259)
(41,44.9357328292765)
(42,46.0580308090132)
(43,42.2490373655985)
(44,44.5662157306455)
(45,38.9782240546039)
(46,38.957713259347)
(47,38.7491530500602)
(48,37.186718372802)
(49,33.8426526710497)
(50,33.2397816401761)
(51,32.5941664594436)
(52,30.8235455436262)
(53,28.4644524522193)
(54,29.5997955395787)
(55,27.0565658781285)
(56,26.0919245237812)
(57,24.0734057935001)
(58,23.8289469182184)
(59,20.8110139459414)
(60,20.2975344938855)
(61,18.9855118347632)
(62,17.51092620941)
(63,16.2280116286374)
(64,14.6220534054784)
(65,13.8176327450332)
(66,13.2977739966643)
(67,9.27795324846028)
(68,7.67585508000608)
(69,10.2892423975626)
(70,7.29037756986124)
(71,5.44809944957142)
(72,4.65313851171506)
(73,1.50027896974953)
(74,0.901525441441402)
(75,2.31266768156428)
(76,0)
(77,0)
(78,0)
(79,0)
(80,0)
(81,0)
(82,0)
(83,0)
(84,0)
(85,0)
(86,0)
(87,0)
(88,0)
(89,0)
(90,0)

};

\addplot[name path = B,draw=none,forget plot] %
                coordinates {
(0,85)
(1,83.2542806705165)
(2,82.297726970973)
(3,83.4693224634016)
(4,79.3085001658783)
(5,76.9250675078146)
(6,77.875544209248)
(7,75.2128948297125)
(8,72.9824757918238)
(9,72.2224987844972)
(10,72.639411195826)
(11,70.5909474819879)
(12,67.9204900825704)
(13,65.439978339995)
(14,65.4262958093113)
(15,65.1537484775383)
(16,61.7074584536228)
(17,62.1349311400102)
(18,59.9878507663539)
(19,59.5206158072533)
(20,58.1274563219783)
(21,56.4691334399164)
(22,55.7207592057596)
(23,55.2592906505509)
(24,52.0638607035002)
(25,52.0468364097478)
(26,49.9511458798209)
(27,50.2569091748287)
(28,50.0702867276814)
(29,47.0858358346499)
(30,46.8998116462416)
(31,43.2636491355178)
(32,42.2951795408044)
(33,40.8849184729113)
(34,39.1289794250645)
(35,38.1639404337465)
(36,37.0089263096877)
(37,35.484331458953)
(38,32.196828660989)
(39,34.2628363305523)
(40,31.2954500726555)
(41,28.9304019076852)
(42,30.0891017468675)
(43,28.0076792705359)
(44,27.0123215079484)
(45,24.9660623000643)
(46,23.4668507820825)
(47,22.7774520622072)
(48,20.3850145247915)
(49,19.5529249433422)
(50,19.3858510265683)
(51,19.6022207636497)
(52,17.0772655877386)
(53,14.2864166535604)
(54,15.0500842842052)
(55,11.5738846157051)
(56,11.0953816772655)
(57,7.90060790550465)
(58,6.34342289620796)
(59,5.79183289327641)
(60,5.75869325238601)
(61,3.60092501927163)
(62,3.73403910164453)
(63,0.832262459305674)
(64,-0.591317937770931)
(65,-1.44152698517953)
(66,-4.19653883693712)
(67,-4.04384726083589)
(68,-6.02650032866053)
(69,-6.11114111194815)
(70,-6.97081136760624)
(71,-9.65859863119084)
(72,-12.5444978042453)
(73,-11.3990028490448)
(74,-13.1263055978969)
(75,-15.6416761641068)
(76,-15)
(77,-15)
(78,-15)
(79,-15)
(80,-15)
(81,-15)
(82,-15)
(83,-15)
(84,-15)
(85,-15)
(86,-15)
(87,-15)
(88,-15)
(89,-15)
(90,-15)

};

 \addplot[name path = C,draw=none,forget plot] %
                coordinates {
             (0,115)
(1,110.9839682766)
(2,111.523699214803)
(3,111.644464992949)
(4,108.602188752413)
(5,108.368235440299)
(6,107.018281072533)
(7,104.887583714873)
(8,104.251290559621)
(9,101.795382385418)
(10,100.873313148573)
(11,100.610929206409)
(12,99.3188251207773)
(13,98.5278439826255)
(14,96.9926745961096)
(15,94.0004582074579)
(16,92.3862988739046)
(17,92.5663350339635)
(18,92.4281438668118)
(19,87.7729498119434)
(20,90.6114652734656)
(21,87.1754090508221)
(22,86.7378726474587)
(23,84.8452228093469)
(24,82.9200855647376)
(25,81.2045525976302)
(26,81.6960458927911)
(27,78.975808241326)
(28,76.6545410718616)
(29,75.3304250502304)
(30,77.2028462279869)
(31,72.483507810891)
(32,72.5511832502362)
(33,69.1606512011972)
(34,70.225213732189)
(35,69.6353531083106)
(36,66.3736205466539)
(37,66.3421740435597)
(38,65.214959855447)
(39,61.3445969565563)
(40,59.8456624650767)
(41,59.2537439111737)
(42,59.4461638085907)
(43,58.9448278730224)
(44,56.4746766076278)
(45,54.4059523841547)
(46,53.1740947906734)
(47,54.0496875035457)
(48,51.8333443187036)
(49,50.5676803746574)
(50,49.8295902860586)
(51,46.7918704680917)
(52,46.1989302666907)
(53,45.2485364463863)
(54,44.1138813296836)
(55,41.025870745768)
(56,41.6937930351178)
(57,38.1968690360245)
(58,38.7096555283592)
(59,37.2319293319003)
(60,35.1409248220747)
(61,34.6466028869021)
(62,33.594321666268)
(63,30.2747192543106)
(64,28.8440820647845)
(65,28.2236487121784)
(66,25.5976939414164)
(67,25.146370925393)
(68,24.6732102691096)
(69,23.314275334313)
(70,21.1237842757842)
(71,19.6790201715405)
(72,19.6101713810496)
(73,20.7603716874208)
(74,16.883342291973)
(75,15.2433102265702)
(76,15)
(77,15)
(78,15)
(79,15)
(80,15)
(81,15)
(82,15)
(83,15)
(84,15)
(85,15)
(86,15)
(87,15)
(88,15)
(89,15)
(90,15)

};
                     
        \addplot [black!35] fill between [of = B and C];

            \end{axis}
            \node at (3.5,6.1) {PPO};
    \end{tikzpicture}
\end{adjustbox}
\end{subfigure}%
\begin{subfigure}[t]{.325\textwidth}
     \centering
    \centering
    \begin{adjustbox}{width=\textwidth}
   \begin{tikzpicture}[scale = 0.84]
            \begin{axis}[
                ymin = 0 ,ymax=100.5,
                xmin=-0.5, xmax=90.5,
                 xlabel = Booking horizon,
                xlabel near ticks,
                ylabel near ticks,
                yticklabel={$\pgfmathprintnumber{\tick}\%$},
                        ]

        \addplot[ name path = A] %
                coordinates {
(0,100)
(1,100.246552062568)
(2,96.431093910338)
(3,94.5967720056366)
(4,95.9618413479283)
(5,92.0874630225632)
(6,92.900746295847)
(7,90.4784827496403)
(8,88.7333907033944)
(9,89.8755619354328)
(10,89.2290940351161)
(11,90.0299232218686)
(12,88.2040212820679)
(13,90.9544429947846)
(14,89.3630784825761)
(15,90.4691728568989)
(16,88.6121659424703)
(17,87.9612207531391)
(18,87.5705510261927)
(19,87.7600849378132)
(20,87.0605179963005)
(21,88.2589064886916)
(22,88.1921423099885)
(23,86.0346086412606)
(24,82.8948573928603)
(25,83.1046606958703)
(26,81.1246655591977)
(27,79.4430079498052)
(28,78.38407835293)
(29,79.9909690834065)
(30,79.9468813444966)
(31,76.7411233253312)
(32,76.9708076923655)
(33,74.7869528647922)
(34,75.5364691233057)
(35,76.3128522574052)
(36,74.0268324702208)
(37,73.1604912222623)
(38,73.6415255927124)
(39,72.2692894719112)
(40,72.8119933436623)
(41,72.0043526448642)
(42,71.8078073929908)
(43,73.406191548066)
(44,70.3530995059781)
(45,72.8409824029064)
(46,72.4376932370228)
(47,71.2856262644956)
(48,69.2149159585499)
(49,67.3840908237743)
(50,66.2713260187171)
(51,67.1833770240137)
(52,66.9229691704116)
(53,67.7354944987665)
(54,68.1342145671023)
(55,66.4908850342491)
(56,63.6914237499008)
(57,63.2954209011012)
(58,60.3155758227252)
(59,56.8210555762825)
(60,57.7420528499834)
(61,55.1189857169601)
(62,53.2832835900791)
(63,50.9655218858001)
(64,46.0827117961762)
(65,43.1075536043063)
(66,41.3711733095742)
(67,41.1224438913507)
(68,39.3786385226873)
(69,40.0033230557192)
(70,41.7554120731759)
(71,35.6642583625963)
(72,35.3437796486024)
(73,31.8516001712715)
(74,28.143244337104)
(75,25.2406924983756)
(76,22)
(77,21)
(78,18)
(79,14)
(80,11)
(81,9)
(82,8)
(83,6)
(84,5)
(85,2)
(86,0)
(87,0)
(88,0)
(89,0)
(90,0)

};

\addplot[name path = B,draw=none,forget plot] %
                coordinates {
(0,80)
(1,79.6247127982235)
(2,76.0912495087688)
(3,75.5177442718149)
(4,76.037778533177)
(5,72.390577882037)
(6,70.5128453233002)
(7,71.3698333563384)
(8,69.8060776488225)
(9,68.5297531707465)
(10,71.4450702645455)
(11,70.4664984653461)
(12,70.963611071924)
(13,70.5632265903283)
(14,70.516066009079)
(15,67.6846333971001)
(16,69.0971468905304)
(17,69.1571663335397)
(18,68.6575537397643)
(19,67.2414231084181)
(20,66.7022390092365)
(21,67.723716210542)
(22,66.3567962373097)
(23,64.7901880482558)
(24,62.7514809085206)
(25,61.0203753768639)
(26,59.6235631081558)
(27,60.5979711461582)
(28,59.5451932588218)
(29,59.7424917667118)
(30,58.9145188945479)
(31,58.2282708334988)
(32,56.7307212503322)
(33,55.2768323114241)
(34,56.9703347762459)
(35,54.9876643261666)
(36,53.5096606558489)
(37,54.1177967804404)
(38,54.7995236568861)
(39,54.0612645222882)
(40,51.6127001390296)
(41,51.8262764947735)
(42,50.8932535064098)
(43,51.7242736244142)
(44,51.1151264318253)
(45,50.6407723471217)
(46,51.3122989295216)
(47,51.3278509478303)
(48,48.7876229104317)
(49,49.4678245698033)
(50,46.2839164121424)
(51,46.7342101926322)
(52,44.6746902118499)
(53,46.6644817866232)
(54,46.8507933620789)
(55,45.1888583588838)
(56,44.573124399997)
(57,41.761587552003)
(58,38.7437775069201)
(59,40.505182605767)
(60,34.8820972216951)
(61,36.1656568050085)
(62,33.5318287975619)
(63,28.1727284897427)
(64,28.0485052489047)
(65,23.9589747994544)
(66,21.775479723808)
(67,20.5420786191456)
(68,21.8234225716816)
(69,21.2553597505511)
(70,21.6562363077549)
(71,16.9698722192383)
(72,14.5885322989827)
(73,12.4724693366987)
(74,7.84197755920197)
(75,7.60103381940323)
(76,2)
(77,1)
(78,-2)
(79,-6)
(80,-9)
(81,-11)
(82,-12)
(83,-14)
(84,-15)
(85,-18)
(86,-20)
(87,-20)
(88,-20)
(89,-20)
(90,-20)

};

 \addplot[name path = C,draw=none,forget plot] %
                coordinates {
            (0,120)
(1,119.034741687274)
(2,117.127244987803)
(3,116.451410542879)
(4,114.221080458237)
(5,112.086521423498)
(6,112.857574787745)
(7,112.159073427597)
(8,110.282716354466)
(9,109.706824510449)
(10,110.753066522902)
(11,109.120302948818)
(12,110.232228411393)
(13,108.789225032081)
(14,110.465896912169)
(15,108.622017970285)
(16,109.166668177992)
(17,110.092926839597)
(18,107.559596528952)
(19,108.91423999721)
(20,107.640705101638)
(21,108.296558776572)
(22,106.292987035187)
(23,104.834824669133)
(24,103.128496786344)
(25,105.395376636997)
(26,102.032934315052)
(27,100.448105577452)
(28,98.3091134039802)
(29,99.6984942818066)
(30,99.0102707882365)
(31,96.3698627407107)
(32,97.8476574037172)
(33,96.525248778065)
(34,95.8247953278217)
(35,95.9907254726663)
(36,93.3867145781383)
(37,93.0444604757058)
(38,92.3824939377304)
(39,92.4136263290775)
(40,90.544962977933)
(41,93.7396375864086)
(42,93.5090591204744)
(43,92.2407127679125)
(44,92.9622357324504)
(45,91.0739206108184)
(46,91.012282107728)
(47,89.8524459746848)
(48,89.6974850627901)
(49,88.3560647301314)
(50,86.8842454777222)
(51,86.4322675179046)
(52,85.8506656719784)
(53,87.3784707436467)
(54,87.0034686084463)
(55,86.3720899193162)
(56,83.6065108427215)
(57,82.2484020386879)
(58,80.0138547363915)
(59,78.3411094217629)
(60,75.2943236296521)
(61,76.8424052161364)
(62,73.5456526408683)
(63,70.0670929295856)
(64,68.3627782735666)
(65,62.9542826369154)
(66,62.7499477779175)
(67,61.5174517089027)
(68,60.6925428885168)
(69,59.3596793325509)
(70,61.612552765802)
(71,55.9799837708677)
(72,54.4589251728472)
(73,52.0302334082485)
(74,46.703776211041)
(75,45.955199577171)
(76,42)
(77,41)
(78,38)
(79,34)
(80,31)
(81,29)
(82,28)
(83,26)
(84,25)
(85,22)
(86,20)
(87,20)
(88,20)
(89,20)
(90,20)

};
                     
        \addplot [black!35] fill between [of = B and C];

            \end{axis}
             \node at (3.5,6.1) {DSPO};
    \end{tikzpicture}
    \end{adjustbox}
\end{subfigure}%
\\
\begin{subfigure}[t]{.35\textwidth}
    \centering
    \begin{adjustbox}{width=\textwidth}
    \begin{tikzpicture}[scale = 0.84]
            \begin{axis}[
               ymin = 0 ,ymax=10,
                xmin=-0.5, xmax=90,
                xlabel = Booking horizon,
                xlabel near ticks,
                ylabel= Avg. accepted OOH discount,
                ylabel near ticks,
                        ]

  \addplot[
        ybar,
    ]
    coordinates {
        (5, 3.65)  
        (10,3.82)
        (15, 3.64)   
        (20,3.92)
        (25, 4.01)   
        (30,3.92)
        (35, 3.92)   
        (40,3.72)
        (45, 3.81) 
        (50,3.42)
        (55, 3.56)
        (60,3.98)
        (65, 3.74)  
        (70,3.73)
        (75, 2.65)   
        (80,2.31)
        (85, 2.03)   
    };
    \addplot[
        only marks,
        mark=-,
        error bars/.cd,
        y dir=both,
        y explicit,
    ]
    coordinates {
        (5, 3.65)  +- (0, 2.5)  
        (10,3.82)+- (0, 2.5) 
        (15, 3.64)   +- (0, 2.4) 
        (20,3.92)+- (0, 2.5) 
        (25, 4.01)   +- (0, 2.2)  
        (30,3.92)+- (0, 2.5) 
        (35, 3.92)   +- (0, 2.7) 
        (40,3.72)+- (0, 2.4) 
        (45, 3.81) +- (0, 2.6) 
        (50,3.42)+- (0, 2.5) 
        (55, 3.56)+- (0, 2.4) 
        (60,3.98)+- (0, 2.6) 
        (65, 3.74)  +- (0, 2.4)  
        (70,3.73)+- (0, 1.7)
        (75, 2.65)   +- (0, 1.5)  
        (80,2.31)+- (0, 0.9) 
        (85, 2.03)   +- (0, 0.8)   

    };
       
            \end{axis}
    \end{tikzpicture}
\end{adjustbox}
\end{subfigure}%
\begin{subfigure}[t]{.325\textwidth}
     \centering
    \centering
    \begin{adjustbox}{width=\textwidth}
   \begin{tikzpicture}[scale = 0.84]
            \begin{axis}[
                ymin = 0 ,ymax=10,
                xmin=-0.5, xmax=90,
                xlabel = Booking horizon,
                xlabel near ticks,
                ylabel near ticks,
                        ]

  \addplot[
        ybar,
    ]
    coordinates {
        (5, 1.65)  
        (10,1.92)
        (15, 2.24)   
        (20,2.02)%
        (25, 1.91)   
        (30,2.02)
        (35, 2.52)   
        (40,2.72)
        (45, 3.22) 
        (50,3.57)
        (55, 5.86)%
        (60,5.96)
        (65, 6.54)  
        (70,6.73)
        (75, 6.65)   
        (80,6.81)
        (85, 7.03)   
    };
    \addplot[
        only marks,
        mark=-,
        error bars/.cd,
        y dir=both,
        y explicit,
    ]
    coordinates {
        (5, 1.65)  +- (0, 3.5)  
        (10,1.92)+- (0, 3.5) 
        (15, 2.24)    +- (0, 3.4)
        (20,2.02)+- (0, 3.5) 
        (25, 1.91)    +- (0, 3.2) 
        (30,2.02)+- (0, 3.5) 
        (35, 2.52)   +- (0, 3.7)  
        (40,2.72)+- (0, 4.4) 
        (45, 3.22)  +- (0, 4.6) 
        (50,3.57)+- (0, 4.5) 
        (55, 5.86)+- (0, 4.4)
        (60,5.96)+- (0, 4.6) 
        (65, 6.54)   +- (0, 4.4) 
        (70,6.73)+- (0, 4.7)
        (75, 6.65)  +- (0, 1.5)   
        (80,6.81)+- (0, 1.9) 
        (85, 7.03)  +- (0, 1.8)     

    };
            \end{axis}
    \end{tikzpicture}
    \end{adjustbox}
\end{subfigure}%

 \caption{The remaining total capacity (Top) and accepted aggregated OOH discounts (Bottom) during the booking horizon for a system with only capacitated OOH locations with capacity of 3; data collected over \num[group-separator={,}]{1000} episodes with a shaded area of 2 standard deviations.}\label{fig:hor_cap}
\end{figure}
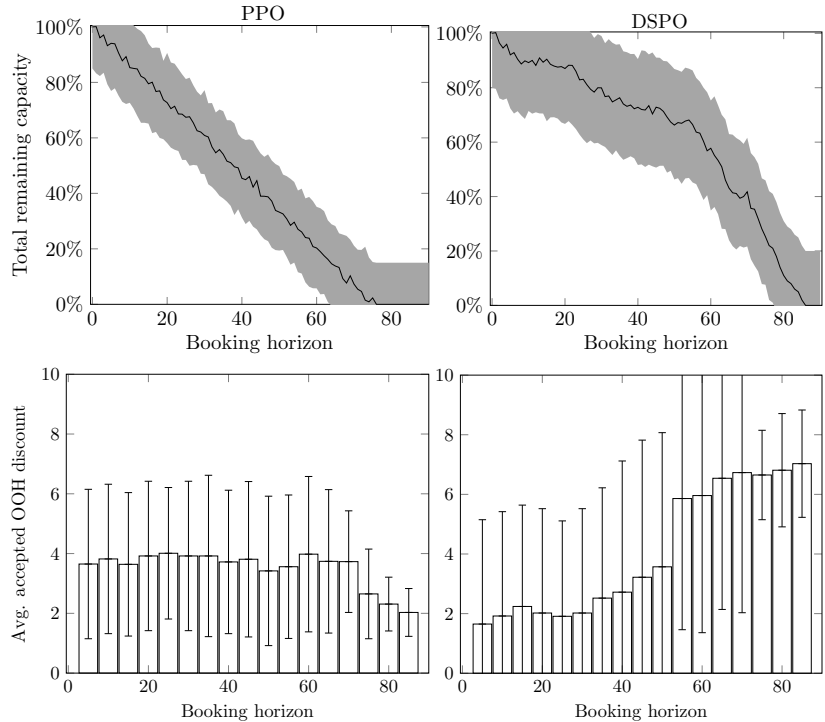

{Next, we visualize the decision making of PPO and DSPO during a booking horizon in a capacitated system with each location having a capacity of $3$, see Figure~\ref{fig:hor_cap}. We intentionally study a system with limited capacity to amplify their effects. In the top row, we show the remaining total OOH capacity during a booking horizon (collected over \num[group-separator={,}]{1000} episodes). The figure confirms that PPO is less able to preserve capacity for later arriving customers compared to DSPO. This way, DSPO can provide OOH delivery longer during the booking horizon, potentially saving capacity for customers that are relatively more expensive to serve at home. However, both policies make sure that OOH locations are fully exploited. Note that over the \num[group-separator={,}]{1000} replications, the capacity of PPO reduces approximately linearly during the booking horizon, but in single episodes still approximately $60\%$ of the customers select home delivery. In the bottom row, we display the average accepted discount for OOH delivery over the same booking horizon. The bars are aggregated over five time steps. The figure illustrates that PPO significantly adjusts its incentives only when there is no remaining capacity. In contrast, DSPO offers lower incentives at the beginning of the booking horizon to deter customers from choosing OOH delivery, thereby preserving some capacity. Later, DSPO provides higher incentives as it gains more certainty about future customer arrivals, effectively nudging more customers towards OOH delivery.
\begin{observation}
\textit{DSPO better preserves OOH capacity than PPO, offering lower initial incentives to deter OOH delivery early and higher incentives later to manage capacity effectively.}
\end{observation}}

\subsubsection{Sensitivity of DSPO to Fuel and Salary Costs and Service Time}\label{sensDSPOsynth}

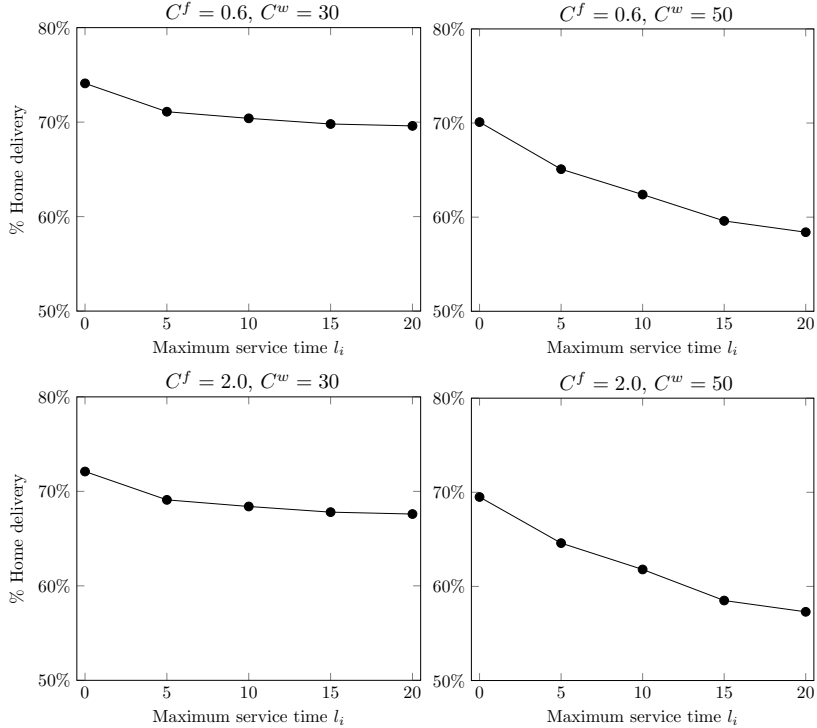
\begin{figure}[hbtp]
\centering
\begin{subfigure}[t]{.35\textwidth}
    \centering
    \begin{adjustbox}{width=\textwidth}
    \begin{tikzpicture}[scale = 0.84]
            \begin{axis}[
                xtick distance=5,
               ymin = 50 ,ymax=80,
                xmin=-0.5, xmax=20.5,
                 xlabel = Maximum service time $l_i$,
                xlabel near ticks,
                ylabel= $\%$ Home delivery,
                ylabel near ticks,
                yticklabel={$\pgfmathprintnumber{\tick}\%$}
                        ]

        \addplot[ mark options={fill=black, mark size=2.5pt},mark=*] %
                coordinates {
                    (0,74.1)
                     (5,71.1)
                    (10,70.4)
                    (15,69.8)
                    (20,69.6)
                };
                
            \end{axis}
            \node at (3.5,6.0) {$C^f=0.6$, $C^w=30$};
    \end{tikzpicture}
\end{adjustbox}
\end{subfigure}%
\begin{subfigure}[t]{.325\textwidth}
     \centering
    \centering
    \begin{adjustbox}{width=\textwidth}
   \begin{tikzpicture}[scale = 0.84]
            \begin{axis}[
                xtick distance=5,
                ymin = 50 ,ymax=80,
                xmin=-0.5, xmax=20.5,
                 xlabel = Maximum service time $l_i$,
                xlabel near ticks,
                ylabel near ticks,
                yticklabel={$\pgfmathprintnumber{\tick}\%$},
                        ]

        \addplot[ mark options={fill=black, mark size=2.5pt},mark=*] %
                coordinates {
                     (0,70.1)
                     (5,65.1)
                    (10,62.4)
                    (15,59.6)
                    (20,58.4)
                };

            \end{axis}
             \node at (3.5,6.0) {$C^f=0.6$, $C^w=50$};
    \end{tikzpicture}
    \end{adjustbox}

\end{subfigure}%
\\
\begin{subfigure}[t]{.35\textwidth}
    \centering
    \begin{adjustbox}{width=\textwidth}
    \begin{tikzpicture}[scale = 0.84]
            \begin{axis}[
                xtick distance=5,
               ymin = 50 ,ymax=80,
                xmin=-0.5, xmax=20.5,
                xlabel = Maximum service time $l_i$,
                xlabel near ticks,
                ylabel= $\%$ Home delivery,
                ylabel near ticks,
                yticklabel={$\pgfmathprintnumber{\tick}\%$}
                        ]

       \addplot[ mark options={fill=black, mark size=2.5pt},mark=*] %
                coordinates {
                    (0,72.1)
                     (5,69.1)
                    (10,68.4)
                    (15,67.8)
                    (20,67.6)
                };
            \end{axis}
             \node at (3.5,6.0) {$C^f=2.0$, $C^w=30$};
    \end{tikzpicture}
\end{adjustbox}
\end{subfigure}%
\begin{subfigure}[t]{.325\textwidth}
     \centering
    \centering
    \begin{adjustbox}{width=\textwidth}
   \begin{tikzpicture}[scale = 0.84]
            \begin{axis}[
                xtick distance=5,
                ymin = 50 ,ymax=80,
                xmin=-0.5, xmax=20.5,
                xlabel = Maximum service time $l_i$,
                xlabel near ticks,
                ylabel near ticks,
                yticklabel={$\pgfmathprintnumber{\tick}\%$}
                        ]

      \addplot[ mark options={fill=black, mark size=2.5pt},mark=*] %
                coordinates {
                 (0,69.5)
                     (5,64.6)
                    (10,61.8)
                    (15,58.5)
                    (20,57.3)
                };

            \end{axis}
             \node at (3.5,6.0) {$C^f=2.0$, $C^w=50$};
    \end{tikzpicture}
    \end{adjustbox}

\end{subfigure}%

 \caption{The percentage of home deliveries of DSPO for normal fuel costs (Top) and high fuel costs (Bottom), given normal salary costs (Left) and high salary costs (Right), for different levels of the maximum service time.}\label{fig:sensDSP}
\end{figure}

Next, we show how DSPO can adjust its policy to different levels of fuel and salary costs ($C^f$ and $C^w$). Figure~\ref{fig:sensDSP} depicts the percentage of home delivery for different settings. From left to right, the salary costs increase from $30$ to $50$ per hour, and from top to bottom the costs of fuel increase from $0.6$ to $2.0$ per distance unit. On the x-axis of each subfigure, we increase the upper bound on service time $l_i$. Note that DSPO is retrained for every setting. The left column shows that higher fuel costs have a modest effect on the percentage of customers who are nudged toward OOH delivery. In contrast, higher salary costs have a bigger impact, as DSPO decides to nudge more customers to OOH delivery options as the service times $l_i$ increase.

\begin{observation}
    \textit{When salary costs increase and the service times are high, DSPO will adapt by giving higher discounts and nudging more customers to OOH delivery.}
\end{observation}

\newpage
\subsubsection{Ablation Study for DSPO}\label{ablation}
\begin{figure}[hbtp]
    \centering
            \begin{tikzpicture}
            \begin{axis}[           tickwidth         = 0pt,
                                ytick             = data,
                                 xlabel = Total costs,
                                  xlabel near ticks,
                                 xmin = 0,
                                symbolic y coords={
                                    {DSPO w/o CNN and retraining},
                                    {DSPO w/o retraining},
                                    {DSPO w/o CNN},
                                    {DSPO}
                                     },
                                 enlarge x limits={upper,value=0.3},
                                  enlarge y limits=0.25,
                                 y=22pt,
                                  xtick={0,1000,...,3000},
                                 point meta=explicit symbolic,
                                nodes near coords,
                                nodes near coords align={horizontal}, 
                            ]

                 \addplot [xbar=0.5cm, bar shift=0pt,nodes near coords, bar width=10pt,color=black,fill=black,style={pattern color=black,mark=none},pattern=north west lines]  coordinates {

                        (3225.3,{DSPO w/o CNN and retraining})%
                        (2904.6,{DSPO w/o retraining})%
                        (3200.2,{DSPO w/o CNN})%
                        (-1,{DSPO})%
                       
                        };

          \addplot [xbar=0.5cm, bar shift=0pt, bar width=10pt,nodes near coords,color=black,fill=black!25, opacity=0.6] coordinates {  
                
                        (2787.6,{DSPO})%
                        };

             \addplot [xbar=0.5cm, bar shift=0pt, bar width=10pt,nodes near coords,color=black,fill=black!25] coordinates {  
                
                         (2787.6,{DSPO w/o CNN and retraining})%
                        (2787.6,{DSPO w/o retraining})%
                        (2787.6,{DSPO w/o CNN})%
                        };

            \node at (axis cs:3152.7,{DSPO}) {2787.6};
            \node at (axis cs:3620.7,{DSPO w/o CNN}) {+14.8$\%$};
            \node at (axis cs:3269.7,{DSPO w/o retraining}) {+4.2$\%$};
            \node at (axis cs:3645.7,{DSPO w/o CNN and retraining}) {+15.7$\%$};

            \end{axis}
            \end{tikzpicture}
              \caption{Barchart showing the total costs for different ablations of DSPO.}    \label{fig:ablation1}
\end{figure}
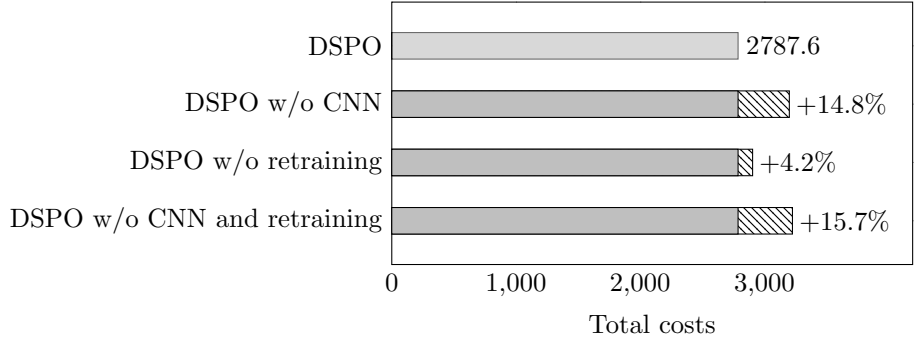

To obtain a better understanding of the working of DSPO, we conduct three ablations, whose results are summarized in Figure~\ref{fig:ablation1}. The first ablation is the removal of the CNN network structure, i.e., the matrix state representation is flattened and directly fed to the FC layers, similar to the state as fed to the Linear benchmark, see Appendix~\ref{appendix::benchmarks}. This way, we show the added value of the feature extraction layers of the CNN. Without the convolutional layers, the total costs increase by $14.8\%$ compared to the non-ablated DSPO. This shows that the CNN layers help in extracting information from the encoded state. Without this feature extraction, the FC-layers are less able to find a performant policy. The second ablation is done on the training algorithm. Instead of updating the weights of the neural network after every iteration (Step 10 in Algorithm~\ref{algo1}), we only give it a one-shot opportunity by providing it with a dataset to train on once, without the opportunity to retrain on newer observations, i.e., we only conduct the initial training phase (Step 1 in Algorithm~\ref{algo1}).  We find that not retraining DSPO increases costs by $4.2\%$. This shows that retraining on new observations is important and motivates the retraining in a \emph{reinforcement learning} fashion. Finally, we show the effect of doing both ablations simultaneously. Not surprisingly, this yields worse results, incurring $15.7\%$ more costs compared to non-ablated DSPO.

\begin{observation}
    \textit{The utilization of convolutional neural networks for state representation, coupled with iterative training loops, is a powerful combination enhancing predictive performance.}
\end{observation}

\subsection{Amazon Case}\label{subsection:amaCase}

In this section, we show the results for a real-world inspired case. First, we explain the problem setting in Section~\ref{sect:amasetting}. Next, we compare DSPO with the benchmarks in Section~\ref{sect:amatable} and analyse the pricing decisions of DSPO and compare it against the best-performing benchmark in Section~\ref{sect:amaanalysiprice}.

\subsubsection{Problem Setting}\label{sect:amasetting}

\begin{figure}[hbtp]
    \centering
     \begin{subfigure}{1.0\textwidth}
     \centering
    \includegraphics[scale=0.33]{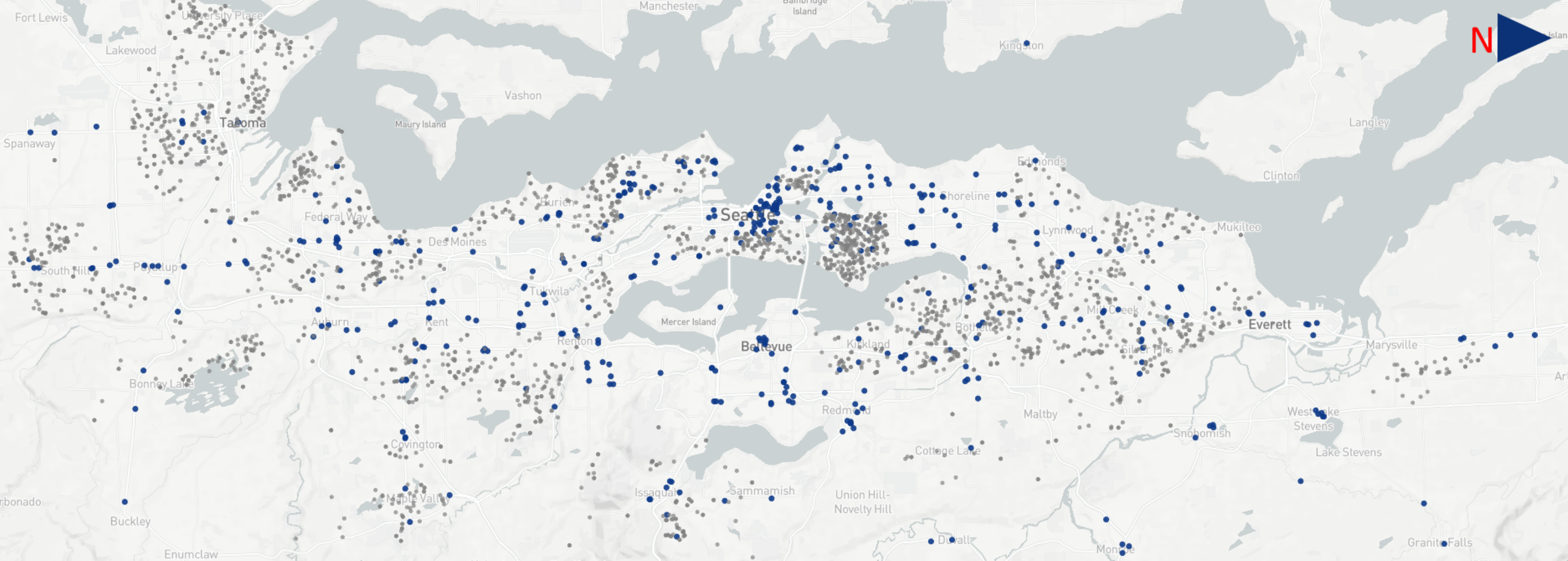}
    \end{subfigure}\\
    \begin{subfigure}{0.4\textwidth}
    \centering
    \includegraphics[width=\textwidth]{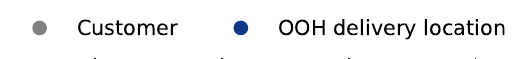}
    \end{subfigure}
   
    \caption{Seattle instance map with customer and OOH locations (North oriented to the right).}
    \label{fig:seattle_instance1}
\end{figure}

We study a real-world case using the order data of the large retailer Amazon in the city of Seattle, USA. We use publicly available data of Amazon from the greater Seattle area \citep{Merchan2022}. Amazon offers OOH delivery in Seattle, but these locations are not contained in the publicly available data set. Hence, we built a tool that scrapes the Amazon site for OOH locations, see \citet{AmazonDeliveryPoints}. The data set with the customer and OOH locations is illustrated in Figure~\ref{fig:seattle_instance1}. 

For each arriving customer, we draw a customer location from the data set. We calculate the driving distances and times using the real road network, including congestion on an average day, obtained from ArcGIS 10.8 \citep{arcgis}. Again, we draw the service times $l_i$ from Equation~\ref{eq:camel} with as bound $l_i\in[1,10]$.  The number of arriving customers on a day is drawn from the negative binomial distribution ${D}$ with $r=700$ and $p=0.5$, which yields $\mathbb{E}[{D}] = 700$. {We consider uniformly distributed interarrival times for each customer.} Each vehicle in the fleet of $25$ vehicles has a capacity of $100$ packages. The fleet operates from a single central depot. Note that the fleet capacity is much higher than the expected number of customers on a day ($700$). This allows us to cope with possible busy days, i.e., when we draw $\gg 700$ from ${D}$. The data set contains $|\mathcal{L}|=299$ OOH locations. {In the data, $38\%$ of the OOH locations are capacitated, with each having a capacity of 42 parcels \cite[cf.][]{AmazonHub}.} The costs of salary, fuel, and delivery failure are $C^f=0.3, C^w=30$, and $C^m=10$, respectively. The probability of delivery failure is $\mathbb{P}^m=0.1$.
\vspace{-2em}
\subsubsection{Comparison of DSPO with Benchmarks}\label{sect:amatable}

Table~\ref{tab:resultsAmazon} shows the results of DSPO and the benchmarks on the Seattle case. We observe that, compared to the synthetic case, overall differences between policies are larger. Since we utilize actual road network driving times, the logistics cost benefits of customers opting for OOH delivery are amplified, particularly for those in remote or traffic-dense areas, which aligns with the results from \citet{Janinhoff2023}. By providing customers the option of OOH delivery, without offering discounts (NoPricing), $9.8\%$ in costs can be saved compared to only offering home delivery (NoOOH). The StaticPricing benchmark offers discounts of $5$ for choosing OOH delivery and a delivery charge of $2$ for home delivery and can save even more ($13.8\%$) by nudging a large number of customers to OOH delivery. The Hindsight and Foresight benchmarks outperform the StaticPricing benchmark: both find a policy that nudges many customers to OOH delivery. Compared to the StaticPricing benchmark, both policies save mainly on travel and service costs. The performance of the learning policies differs a lot: the Linear benchmark and PPO are unable to find a performant policy, whereas DSPO finds the best overall policy, saving $20.8\%$ in total costs compared to NoOOH. Clearly, the linear benchmark is unable to abstract the complex relationships from the encoded state, and hence, it does not yield an accurate estimation of the costs. Even after significant tuning effort, PPO does not converge within \num[group-separator={,}]{600000} episodes. We conjecture that PPO is unable to deal with the sparse cost and long episode length as encountered with this problem: only at the cutoff time $T$, the costs are revealed, which troubles learning convergence. All convergence curves can be found in Appendix~\ref{appendix:dspo]} and~\ref{appendix::benchmarks}. DSPO, however, can find a performant policy that nudges almost half of the customers to OOH delivery. It seems that, compared to the synthetic case, it is more profitable to nudge many customers to OOH delivery, whereas for the synthetic case, it was sometimes better to provide lower discounts since home delivery was not always as expensive. Compared to the state-of-the-art benchmarks, DSPO saves $19.9\%$pt (compared to PPO) and $3.8\%$pt (compared to the Foresight benchmark) in total costs. {Due to the limited number of finite capacity OOH locations ($38\%$) and their fairly high capacity, no OOH location reaches its capacity on a single delivery day.}

\begin{observation}
    \textit{DSPO finds the best policy and saves $19.9\%$pt in total cost compared to not offering OOH delivery, $7\%$pt compared to a static pricing policy, and $3.8\%$pt compared to the best-performing state-of-the-art benchmark (Foresight benchmark) for the Amazon case.}
\end{observation}

\begin{table}[htbp]
	\caption{Results on the Amazon Seattle case.}
	\label{tab:resultsAmazon}
	\begin{center}
 \resizebox{\textwidth}{!}{%
		\begin{tabular}{l c c c c c c c }
			\hline
			  & {\makecell{$\%$ Home \\ delivery}} &  {\makecell{Travel\\ costs}} & {\makecell{Service \\ costs}} & {\makecell{Delivery fail. \\ costs}}  & {\makecell{Discount costs \\ (Charge revenue)}}  &  {\makecell{Avg. discount \\ (Avg. charge)}} & $\%$ Savings (95$\%$ CI)  \\
			\hline

			\makecell[l]{NoOOH} & $100\%$ & 2091.4 & 2948.1 & 839.5 & - & - & - \\
			
			\makecell[l]{OnlyOOH} & $0\%$ & 1404.7 & 378.0 & 0.0 & - & - & $69.7\%$ ($\pm0.7\%$) \\

            \makecell[l]{NoPricing} & $81.4\%$ & 2221.9 & 2400.5 & 683.5 & - & - & $9.8\%$ ($\pm0.9\%$)  \\

             \makecell[l]{StaticPricing} & $56.3\%$ & 2178.5 & 1658.9 & 472.4 & 1738.5 (983.5) & $5 \pm 0$ ( $2 \pm 0$) & $13.8\%$ ($\pm1.0\%$)\\

             \makecell[l]{Hindsight benchmark} &  $53.3\%$ & 2099.0 & 1461.0 & 416.0 & 1830.6 860.0) & $5.2 \pm 0.6$ ( $2 \pm 0$) & $15.9\%$ ($\pm1.2\%$) \\

             \makecell[l]{Foresight benchmark} & $54.2\%$ & 2075.8 & 1485.6 & 423.0 & 1767.5 (876.7) & $5.2 \pm 1.2$ ( $2 \pm 0.1$) & $17.0\%$ ($\pm1.1\%$) \\

             \makecell[l]{Linear benchmark} & $62.8\%$ & 2176.6 & 1914.0 & 545.0 & 1610.9 (806.5) & $4.0 \pm 2.1$ ( $1.7 \pm 0.5$) & $7.5\%$ ($\pm1.2\%$) \\
             
            \makecell[l]{PPO benchmark} & $60.4\%$ & 2193.5 & 1784.5 & 508.1 & 1820.3 (478.9) & $3.6 \pm 2.5$ ( $1.4 \pm 0.7$) & $0.9\%$  ($\pm1.4\%$)\\

             \makecell[l]{DSPO} & $52.8\%$ & 2068.3 & 1311.8 & 409.2 & 1748.8 (881.4) & $5.1 \pm 2.3$ ( $2 \pm 0.1$) & $20.8\%$ ($\pm1.3\%$) \\

			\hline
		\end{tabular}}
	\end{center}
\end{table}
\vspace{-2em}

\subsubsection{Analysis of Pricing Decisions}\label{sect:amaanalysiprice}

\begin{figure}[hbtp]
\centering
\begin{subfigure}{.45\textwidth}
     \centering
    \centering
    \begin{tikzpicture}[scale = 0.84]
            \begin{axis}[
            ytick={-5.4,-5,-4.6,-4.2,-3.8},
                ymin=-5.4, ymax=-3.8,
                xmin=-10,xmax=1020,
                 label = Time on booking horizon,
                xlabel near ticks,
                ylabel= Accepted discount,
                ylabel near ticks,
                  xticklabel shift={0.1cm},
                  legend cell align={left},
                  xlabel = Time on booking horizon,
                 colormap = {whiteblack}{color(0cm)  = (black!20);color(1cm) = (black)},
                        ]
              \addplot[
            scatter,
            only marks,
            scatter src=explicit,
            mark=*,
            scatter/use mapped color={
                draw=mapped color,
                fill=mapped color,
                y filter/.code={\pgfmathparse{#1*-1}\pgfmathresult}
            },
            visualization depends on={\thisrow{C} \as \perpointmarksize},
            scatter/@pre marker code/.append style={
                /tikz/mark size={1pt+abs(\perpointmarksize/50)}
            },
        ] table [meta=C] {
A	B	C
0	-4.70606	14.15981199
3.61296	-4.75636	13.63102233
7.225884	-4.65576	38.89541716
10.838844	-4.68091	24.50058754
14.451804	-4.68091	25.49941246
18.064764	-4.68091	10.10575793
21.677688	-4.68091	45.29964747
25.290648	-4.70606	49.94124559
28.903608	-4.75636	40.36427732
32.516532	-4.73121	37.36780259
36.1296	-4.70606	19.38895417
39.74256	-4.70606	39.77673325
43.35552	-4.75636	24.26556992
46.96848	-4.70606	25.79318449
50.58144	-4.70606	48.29612221
54.1944	-4.73121	36.42773208
57.80736	-4.78152	45.65217391
61.42032	-4.80667	83.72502938
65.03292	-4.83182	42.77320799
68.64588	-4.85697	31.37485311
72.25884	-4.80667	29.96474736
75.8718	-4.80667	62.22091657
79.48476	-4.78152	36.6039953
83.09772	-4.73121	34.13631022
86.71068	-4.78152	21.91539365
90.32364	-4.78152	52.29142186
93.9366	-4.83182	36.36897767
97.54956	-4.83182	39.95299647
101.16252	-4.88212	26.85076381
104.77548	-4.88212	50.05875441
108.38844	-4.85697	43.30199765
112.0014	-4.88212	68.91891892
115.61436	-4.90727	83.60752056
119.22732	-4.80667	36.89776733
122.84028	-4.80667	99.47121034
126.45324	-4.78152	49.70622797
130.0662	-4.73121	24.61809636
133.67916	-4.80667	29.02467685
137.29212	-4.80667	35.01762632
140.90508	-4.85697	37.5440658
144.51804	-4.83182	40.65804935
148.131	-4.83182	30.25851939
151.74396	-4.83182	42.53819036
155.35692	-4.75636	99.29494712
158.96988	-4.70606	29.67097532
162.58284	-4.60545	15.98119859
166.1958	-4.53	32.25616921
169.80876	-4.55515	82.49118684
173.42172	-4.65576	54.52408931
177.03468	-4.65576	48.76615746
180.64764	-4.70606	36.83901293
184.2606	-4.75636	43.83078731
187.87356	-4.83182	30.61104583
191.48616	-4.80667	73.50176263
195.09912	-4.80667	34.01880141
198.71208	-4.78152	37.30904818
202.32504	-4.80667	34.66509988
205.938	-4.80667	22.44418331
209.55096	-4.78152	31.96239718
213.16392	-4.80667	32.25616921
216.77688	-4.83182	20.03525264
220.38984	-4.78152	22.85546416
224.0028	-4.75636	63.57226792
227.61576	-4.78152	30.61104583
231.22872	-4.80667	33.31374853
234.84168	-4.78152	54.70035253
238.45464	-4.80667	42.53819036
242.0676	-4.83182	41.36310223
245.68056	-4.83182	47.76733255
249.29352	-4.75636	15.45240893
252.90648	-4.78152	22.20916569
256.51944	-4.78152	32.02115159
260.1324	-4.78152	24.20681551
263.74536	-4.80667	13.16098707
267.35832	-4.78152	47.59106933
270.97128	-4.70606	33.31374853
274.58424	-4.68091	9.518213866
278.1972	-4.63061	46.65099882
281.81016	-4.60545	20.21151586
285.42312	-4.63061	35.25264395
289.03608	-4.60545	28.14336075
292.64904	-4.60545	18.91891892
296.262	-4.5803	46.23971798
299.87496	-4.63061	28.84841363
303.48792	-4.63061	42.8907168
307.10088	-4.68091	29.55346651
310.71384	-4.73121	19.21269095
314.32644	-4.68091	21.386604
317.9394	-4.73121	36.36897767
321.55236	-4.70606	42.8319624
325.16532	-4.70606	34.54759107
328.77828	-4.68091	35.54641598
332.39124	-4.68091	39.306698
336.0042	-4.70606	32.49118684
339.61716	-4.80667	14.86486486
343.23012	-4.88212	15.27614571
346.84308	-4.93242	49.70622797
350.45604	-4.88212	51.99764982
354.069	-4.80667	56.81551116
357.68196	-4.85697	48.64864865
361.296	-4.83182	15.62867215
364.9068	-4.80667	34.13631022
368.5212	-4.80667	18.33137485
372.1356	-4.78152	100
375.7464	-4.78152	40.1880141
379.3608	-4.83182	50.23501763
382.9716	-4.80667	35.48766157
386.586	-4.80667	15.86368978
390.1968	-4.85697	25.61692127
393.8112	-4.83182	34.19506463
397.4256	-4.80667	40.89306698
401.0364	-4.88212	38.30787309
404.6508	-4.85697	43.83078731
408.2616	-4.88212	52.35017626
411.876	-4.90727	23.73678026
415.4904	-4.93242	75.79318449
419.1012	-4.93242	75.14688602
422.7156	-4.93242	7.990599295
426.3264	-4.90727	13.92479436
429.9408	-4.93242	33.66627497
433.5552	-4.95758	34.95887192
437.166	-5.00788	35.01762632
440.7804	-5.05818	89.01292597
444.3912	-5.03303	33.31374853
448.0056	-5.05818	23.73678026
451.62	-5.08333	20.79905993
455.2308	-5.08333	10.28202115
458.8452	-5.08333	31.37485311
462.456	-5.08333	41.71562867
466.0704	-5.03303	29.61222092
469.6848	-5.00788	89.48296122
473.2956	-4.98273	24.38307873
476.91	-5.00788	17.8613396
480.5208	-5.03303	14.15981199
484.1352	-5.00788	30.19976498
487.7496	-5.00788	47.00352526
491.3604	-5.00788	35.19388954
494.9748	-5.00788	78.84841363
498.5856	-4.98273	20.91656874
502.2	-4.95758	18.33137485
505.8144	-4.98273	25.49941246
509.4252	-5.00788	32.84371328
513.0396	-5.05818	36.89776733
516.6504	-5.05818	39.42420682
520.2648	-5.00788	9.811985899
523.8792	-4.98273	78.90716804
527.49	-4.98273	12.98472385
531.1044	-5.03303	38.48413631
534.7152	-5.03303	32.72620447
538.3296	-5.03303	25.44065805
541.944	-5.05818	14.21856639
545.5548	-5.08333	33.60752056
549.1692	-5.08333	45.41715629
552.78	-5.10848	38.30787309
556.3944	-5.10848	45.65217391
560.0088	-5.05818	30.55229142
563.6196	-5.08333	38.07285546
567.234	-5.08333	40.1880141
570.8448	-5.08333	73.03172738
574.4592	-5.08333	34.13631022
578.0736	-5.05818	45.59341951
581.6844	-5.03303	31.02232667
585.2988	-5.05818	50.58754407
588.9096	-5.03303	63.86603995
592.524	-5.03303	42.8319624
596.1384	-5.03303	19.6827262
599.7492	-5.00788	51.64512338
603.3636	-4.98273	42.00940071
606.9744	-5.00788	29.20094007
610.5888	-5.05818	77.96709753
614.2032	-5.05818	25.0293772
617.814	-5.05818	8.871915394
621.4284	-5.05818	35.66392479
625.0392	-5.08333	36.36897767
628.6536	-5.05818	38.77790834
632.2644	-5.10848	32.66745006
635.8788	-5.13364	65.74618096
639.4932	-5.15879	39.18918919
643.104	-5.18394	81.43360752
646.7184	-5.23424	28.78965922
650.3292	-5.20909	19.6827262
653.9436	-5.20909	85.7226792
657.558	-5.23424	43.12573443
661.1688	-5.25939	30.78730905
664.7832	-5.28455	37.01527615
668.394	-5.25939	52.7027027
672.0084	-5.28455	22.67920094
675.6228	-5.3097	62.63219741
679.2336	-5.33485	94.30082256
682.848	-5.33485	77.37955347
686.4588	-5.3097	79.72972973
690.0732	-5.28455	34.60634548
693.6876	-5.28455	39.71797885
697.2984	-5.25939	47.06227967
700.9128	-5.28455	12.10340776
704.5236	-5.33485	40.30552291
708.138	-5.36	20.97532315
711.7524	-5.33485	24.91186839
715.3632	-5.36	58.46063455
718.9776	-5.3097	34.31257344
722.5884	-5.3097	37.83783784
726.2028	-5.28455	52.46768508
729.8172	-5.33485	25.67567568
733.428	-5.3097	94.53584019
737.0424	-5.33485	36.31022327
740.6532	-5.36	34.43008226
744.2676	-5.33485	45.94594595
747.882	-5.3097	11.75088132
751.4928	-5.28455	45.29964747
755.1072	-5.25939	35.60517039
758.718	-5.33485	8.75440658
762.3324	-5.33485	87.30904818
765.9468	-5.3097	46.4159812
769.5576	-5.28455	40.24676851
773.172	-5.28455	37.60282021
776.7828	-5.23424	19.09518214
780.3972	-5.18394	20.38777908
784.0116	-5.18394	30.84606345
787.6224	-5.20909	24.0893067
791.2368	-5.15879	8.989424207
794.8476	-5.13364	38.95417156
798.462	-5.10848	31.96239718
802.0764	-5.13364	30.55229142
805.6872	-5.13364	44.18331375
809.3016	-5.10848	40.54054054
812.9124	-5.10848	40.83431257
816.5268	-5.08333	91.9506463
820.1412	-5.08333	31.49236193
823.752	-5.10848	18.15511163
827.3664	-5.10848	31.60987074
830.9772	-5.13364	53.76028202
834.5916	-5.10848	51.29259694
838.206	-5.10848	30.66980024
841.8168	-5.10848	37.36780259
845.4312	-5.10848	26.73325499
849.042	-5.10848	22.56169213
852.6564	-5.05818	26.55699177
856.2708	-5.03303	75.55816686
859.8816	-5.03303	82.60869565
863.496	-5.03303	37.5440658
867.1068	-5.03303	45.71092832
870.7212	-5.00788	24.44183314
874.3356	-4.98273	75.73443008
877.9464	-4.98273	30.08225617
881.5608	-5.00788	21.21034078
885.1716	-5.03303	31.96239718
888.786	-5.03303	17.21504113
892.3968	-5.05818	67.09753231
896.0112	-5.05818	29.14218566
899.6256	-5.10848	14.15981199
903.2364	-5.05818	25.44065805
906.8508	-5.00788	24.20681551
910.4616	-4.95758	36.66274971
914.076	-4.95758	93.47826087
917.6904	-4.93242	52.82021152
921.3012	-4.95758	16.86251469
924.9156	-4.93242	39.65922444
928.5264	-4.93242	51.05757932
932.1408	-4.93242	24.44183314
935.7552	-4.95758	8.049353702
939.366	-4.95758	35.31139835
942.9804	-4.98273	23.61927145
946.5912	-4.95758	79.96474736
950.2056	-5.00788	18.7426557
953.82	-5.03303	23.56051704
957.4308	-4.98273	37.77908343
961.0452	-5.03303	19.6239718
964.656	-5.05818	29.96474736
968.2704	-5.00788	23.61927145
971.8848	-5.03303	37.36780259
975.4956	-5.03303	22.3266745
979.11	-5.00788	23.26674501
982.7208	-4.98273	15.1586369
986.3352	-4.93242	29.4359577
989.9496	-5.00788	22.09165687
993.5604	-4.93242	19.80023502
997.1748	-4.85697	39.306698
1000.7856	-4.85697	40.65804935
1004.4	-4.85697	49.82373678

        };

            \end{axis}
                        \node at (3.5,6.0) {Foresight benchmark discount decisions};
    \end{tikzpicture}

\end{subfigure}%
\begin{subfigure}{.55\textwidth}
    \centering
    \begin{tikzpicture}[scale = 0.84]
                       \begin{axis}[
                ytick={-5.4,-5,-4.6,-4.2,-3.8},
                ymin=-5.4, ymax=-3.8,
                 xmin=-10,xmax=1020,
                 xlabel = Time on booking horizon,
                xlabel near ticks,
                ylabel= Accepted discount,
                ylabel near ticks,
                  xticklabel shift={0.1cm},
                  legend cell align={left},
                   colorbar,
                   colormap = {whiteblack}{color(0cm)  = (black!20);color(1cm) = (black)},
                  colorbar style={
                     ylabel= Dist. from home to OOH (scaled)},
                        ]
              \addplot[
            scatter,
            only marks,
            scatter src=explicit,
            mark=*,
            scatter/use mapped color={
                draw=mapped color,
                fill=mapped color,
            },
            visualization depends on={\thisrow{C} \as \perpointmarksize},
            scatter/@pre marker code/.append style={
                /tikz/mark size={1pt+abs(\perpointmarksize/50)}
            },
        ] table [meta=C] {
A	B	C
0	-3.98833	15.285702041
3.61296	-4.025	15.285702041
7.225884	-4.135	19.28570204
10.838844	-4.19	15.285702041
14.451804	-4.26333	44.28570204
18.064764	-4.3	15.285702041
21.677688	-4.28167	44.28570204
25.290648	-4.245	44.28570204
28.903608	-4.39167	15.285702041
32.516532	-4.31833	15.285702041
36.1296	-4.3	15.285702041
39.74256	-4.33667	15.285702041
43.35552	-4.355	15.285702041
46.96848	-4.20833	44.28570204
50.58144	-4.22667	15.285702041
54.1944	-4.22667	15.285702041
57.80736	-4.28167	15.285702041
61.42032	-4.26333	15.285702041
65.03292	-4.26333	44.28570204
68.64588	-4.355	15.285702041
72.25884	-4.3	15.285702041
75.8718	-4.33667	15.285702041
79.48476	-4.355	15.285702041
83.09772	-4.33667	15.285702041
86.71068	-4.50167	15.285702041
90.32364	-4.575	44.28570204
93.9366	-4.48333	15.285702041
97.54956	-4.59333	15.285702041
101.16252	-4.59333	15.285702041
104.77548	-4.575	15.285702041
108.38844	-4.66667	44.28570204
112.0014	-4.70333	44.28570204
115.61436	-4.72167	44.28570204
119.22732	-4.77667	15.285702041
122.84028	-4.75833	15.285702041
126.45324	-4.70333	15.285702041
130.0662	-4.77667	15.285702041
133.67916	-4.795	15.285702041
137.29212	-4.86833	74.28570204
140.90508	-4.86833	15.285702041
144.51804	-4.88667	15.285702041
148.131	-4.94167	74.28570204
151.74396	-4.99667	74.28570204
155.35692	-5.015	74.28570204
158.96988	-4.99667	74.28570204
162.58284	-4.96	74.28570204
166.1958	-5.05167	74.28570204
169.80876	-5.07	100
173.42172	-5.10667	15
177.03468	-5.125	16.78570204
180.64764	-5.19833	100
184.2606	-5.21667	9.841262041
187.87356	-5.19833	11.23014204
191.48616	-5.18	12.61903204
195.09912	-5.21667	100
198.71208	-5.29	30
202.32504	-5.21667	9.841262041
205.938	-5.235	8.452372041
209.55096	-5.235	100
213.16392	-5.18	100
216.77688	-5.14333	100
220.38984	-5.07	100
224.0028	-5.03333	74.28570204
227.61576	-5.10667	18.17460204
231.22872	-5.05167	15.285702041
234.84168	-4.86833	74.28570204
238.45464	-4.86833	15.285702041
242.0676	-4.88667	15.285702041
245.68056	-4.86833	15.285702041
249.29352	-4.83167	15.285702041
252.90648	-4.85	74.28570204
256.51944	-4.905	74.28570204
260.1324	-4.92333	74.28570204
263.74536	-4.97833	15.285702041
267.35832	-4.97833	15.285702041
270.97128	-4.96	74.28570204
274.58424	-5.015	15.285702041
278.1972	-5.015	15.285702041
281.81016	-5.015	15.285702041
285.42312	-4.97833	74.28570204
289.03608	-4.96	15.285702041
292.64904	-4.96	74.28570204
296.262	-4.96	74.28570204
299.87496	-4.94167	74.28570204
303.48792	-4.92333	74.28570204
307.10088	-4.97833	15.285702041
310.71384	-4.905	74.28570204
314.32644	-4.905	15.285702041
317.9394	-4.83167	74.28570204
321.55236	-4.85	15.285702041
325.16532	-4.85	74.28570204
328.77828	-4.795	15.285702041
332.39124	-4.72167	15.285702041
336.0042	-4.72167	15.285702041
339.61716	-4.81333	15.285702041
343.23012	-4.75833	15.285702041
346.84308	-4.77667	15.285702041
350.45604	-4.685	44.28570204
354.069	-4.685	44.28570204
357.68196	-4.75833	15.285702041
361.296	-4.66667	15.285702041
364.9068	-4.685	15.285702041
368.5212	-4.61167	15.285702041
372.1356	-4.61167	44.28570204
375.7464	-4.55667	15.285702041
379.3608	-4.41	44.28570204
382.9716	-4.50167	15.285702041
386.586	-4.50167	44.28570204
390.1968	-4.50167	15.285702041
393.8112	-4.50167	44.28570204
397.4256	-4.52	44.28570204
401.0364	-4.63	15.285702041
404.6508	-4.61167	15.285702041
408.2616	-4.61167	15.285702041
411.876	-4.66667	15.285702041
415.4904	-4.77667	15.285702041
419.1012	-4.77667	74.28570204
422.7156	-4.795	15.285702041
426.3264	-4.81333	74.28570204
429.9408	-4.86833	15.285702041
433.5552	-4.86833	74.28570204
437.166	-4.795	15.285702041
440.7804	-4.795	15.285702041
444.3912	-4.77667	74.28570204
448.0056	-4.63	44.28570204
451.62	-4.70333	15.285702041
455.2308	-4.72167	44.28570204
458.8452	-4.72167	44.28570204
462.456	-4.575	15.285702041
466.0704	-4.685	15.285702041
469.6848	-4.66667	15.285702041
473.2956	-4.61167	44.28570204
476.91	-4.55667	15.285702041
480.5208	-4.575	44.28570204
484.1352	-4.63	44.28570204
487.7496	-4.64833	15.285702041
491.3604	-4.75833	44.28570204
494.9748	-4.75833	44.28570204
498.5856	-4.72167	44.28570204
502.2	-4.77667	15.285702041
505.8144	-4.77667	15.285702041
509.4252	-4.83167	74.28570204
513.0396	-4.905	74.28570204
516.6504	-4.81333	15.285702041
520.2648	-4.81333	15.285702041
523.8792	-4.85	15.285702041
527.49	-4.77667	15.285702041
531.1044	-4.85	15.285702041
534.7152	-4.795	15.285702041
538.3296	-4.66667	44.28570204
541.944	-4.74	44.28570204
545.5548	-4.66667	15.285702041
549.1692	-4.685	15.285702041
552.78	-4.685	44.28570204
556.3944	-4.72167	15.285702041
560.0088	-4.85	15.285702041
563.6196	-4.77667	74.28570204
567.234	-4.77667	15.285702041
570.8448	-4.77667	15.285702041
574.4592	-4.83167	74.28570204
578.0736	-4.81333	15.285702041
581.6844	-4.81333	15.285702041
585.2988	-4.92333	15.285702041
588.9096	-4.85	74.28570204
592.524	-4.75833	44.28570204
596.1384	-4.85	15.285702041
599.7492	-4.75833	15.285702041
603.3636	-4.74	44.28570204
606.9744	-4.75833	44.28570204
610.5888	-4.59333	44.28570204
614.2032	-4.59333	44.28570204
617.814	-4.575	15.285702041
621.4284	-4.55667	44.28570204
625.0392	-4.52	15.285702041
628.6536	-4.61167	15.285702041
632.2644	-4.64833	44.28570204
635.8788	-4.55667	15.285702041
639.4932	-4.575	44.28570204
643.104	-4.63	15.285702041
646.7184	-4.59333	15.285702041
650.3292	-4.59333	44.28570204
653.9436	-4.48333	15.285702041
657.558	-4.42833	44.28570204
661.1688	-4.42833	15.285702041
664.7832	-4.37333	44.28570204
668.394	-4.31833	15.285702041
672.0084	-4.37333	15.285702041
675.6228	-4.37333	15.285702041
679.2336	-4.42833	44.28570204
682.848	-4.42833	44.28570204
686.4588	-4.42833	44.28570204
690.0732	-4.355	15.285702041
693.6876	-4.37333	15.285702041
697.2984	-4.44667	15.285702041
700.9128	-4.52	15.285702041
704.5236	-4.48333	15.285702041
708.138	-4.465	44.28570204
711.7524	-4.39167	44.28570204
715.3632	-4.355	15.285702041
718.9776	-4.42833	44.28570204
722.5884	-4.33667	15.285702041
726.2028	-4.37333	44.28570204
729.8172	-4.355	15.285702041
733.428	-4.355	44.28570204
737.0424	-4.50167	15.285702041
740.6532	-4.44667	44.28570204
744.2676	-4.53833	44.28570204
747.882	-4.48333	15.285702041
751.4928	-4.465	44.28570204
755.1072	-4.41	15.285702041
758.718	-4.39167	15.285702041
762.3324	-4.39167	44.28570204
765.9468	-4.39167	15.285702041
769.5576	-4.355	15.285702041
773.172	-4.37333	44.28570204
776.7828	-4.22667	15.285702041
780.3972	-4.11667	19.28570204
784.0116	-4.17167	19.28570204
787.6224	-4.15333	19.28570204
791.2368	-4.19	44.28570204
794.8476	-4.17167	19.28570204
798.462	-4.26333	44.28570204
802.0764	-4.31833	95.285702041
805.6872	-4.19	44.28570204
809.3016	-4.08	15.285702041
812.9124	-3.97	15.285702041
816.5268	-4.04333	19.28570204
820.1412	-3.97	19.28570204
823.752	-4.06167	15.285702041
827.3664	-4.135	15.285702041
830.9772	-4.17167	19.28570204
834.5916	-4.245	15.285702041
838.206	-4.33667	15.285702041
841.8168	-4.26333	15.285702041
845.4312	-4.3	15.285702041
849.042	-4.355	15.285702041
852.6564	-4.28167	15.285702041
856.2708	-4.22667	15.285702041
859.8816	-4.26333	15.285702041
863.496	-4.26333	15.285702041
867.1068	-4.15333	19.28570204
870.7212	-4.31833	15.285702041
874.3356	-4.19	44.28570204
877.9464	-4.19	44.28570204
881.5608	-4.025	19.28570204
885.1716	-4.06167	15.285702041
888.786	-4.11667	15.285702041
892.3968	-4.26333	44.28570204
896.0112	-4.355	15.285702041
899.6256	-4.33667	15.285702041
903.2364	-4.41	44.28570204
906.8508	-4.42833	44.28570204
910.4616	-4.50167	44.28570204
914.076	-4.50167	15.285702041
917.6904	-4.37333	15.285702041
921.3012	-4.39167	15.285702041
924.9156	-4.245	15.285702041
928.5264	-4.19	15.285702041
932.1408	-4.20833	15.285702041
935.7552	-4.31833	44.28570204
939.366	-4.37333	44.28570204
942.9804	-4.28167	15.285702041
946.5912	-4.19	15.285702041
950.2056	-4.3	15.285702041
953.82	-4.41	44.28570204
957.4308	-4.48333	15.285702041
961.0452	-4.61167	15.285702041
964.656	-4.53833	44.28570204
968.2704	-4.42833	15.285702041
971.8848	-4.52	44.28570204
975.4956	-4.465	15.285702041
979.11	-4.52	44.28570204
982.7208	-4.48333	15.285702041
986.3352	-4.70333	15.285702041
989.9496	-4.81333	15.285702041
993.5604	-4.92333	15.285702041
997.1748	-4.94167	15.285702041
1000.7856	-5.015	74.28570204
1004.4	-4.92333	15.285702041

        };

            \end{axis}
            \node at (3.5,6.0) {DSPO discount decisions};
    \end{tikzpicture}
\end{subfigure}
  \caption{Analysis of the accepted OOH discounts over the booking horizon given the distance from the customer's home to the OOH location, moving average over 20 time steps. }\label{fig:pricing}
\end{figure}
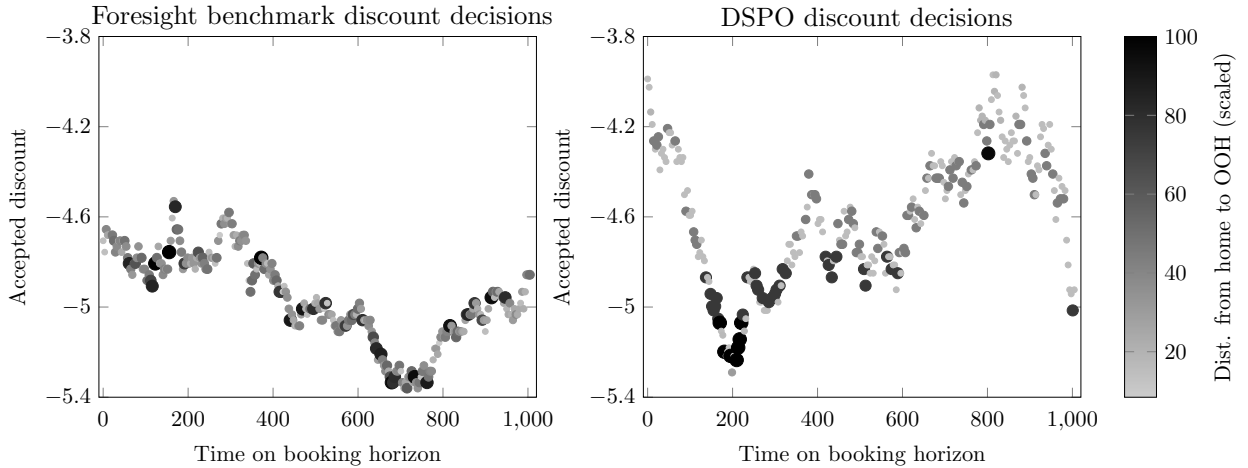

Figure~\ref{fig:pricing} depicts the accepted discounts for OOH delivery during a booking horizon for the Foresight benchmark (left) and DSPO (right). The color and size of the scatter points indicate the distance from the customer's home to the OOH location. Results are plotted over a single, exemplary booking horizon. At the beginning of the booking horizon, the Foresight benchmark seems to provide similar discounts to all customers. As time progresses and the booking horizon advances, the estimated cost for a new customer increases, leading to higher discounts (in the time interval from $600$ to $800$). Only towards the end of the horizon, the cost estimation returns to the level of the beginning of the horizon, providing similar discounts. Comparing this pricing behavior with DSPO, we observe three major differences: (i) DSPO has a slightly higher deviation in accepted discounts, (ii) DSPO provides higher discounts to customers that have to travel further for an OOH location, and (iii) DSPO seems to provide highest discounts in the time interval from $200$ to $600$ of the booking horizon. 

\begin{observation}
    \textit{DSPO can adapt to customer behavior by providing higher discounts to customers who are less likely to select OOH delivery.}
\end{observation}

It seems that the $200-600$ interval of the booking horizon is most crucial in nudging customers. DSPO seems to anticipate customer arrivals and customer behavior by giving higher discounts to remote customers early in the booking horizon. After this interval, nudging customers is of less importance to DSPO, as it provides lower discounts. Only at the end of the booking horizon, DSPO provides higher discounts again. 

\begin{observation}
    \textit{DSPO can adapt its pricing policy to the time of arrival in the booking horizon.  Specifically, the interval between $20\%$ and $60\%$ of the elapsed booking horizon is critical for effectively nudging customers.}
\end{observation}

\section{Conclusion}\label{section:conclusion}

In this paper, we studied the dynamic selection and pricing of out-of-home (OOH) deliveries. The studied problem is novel since it considers (i) stochastic customer arrivals, (ii) stochastic customer choice, and (iii) dynamic decision-making, as we make sequential decisions {on the customer incentives provided for OOH delivery}, without knowing the customers that will arrive in the {remainder of the day}. We defined an MDP for the studied problem, {presented a novel solution,} and studied a small synthetic case, before moving to a real-world inspired case of deliveries in Seattle, USA. We proposed Dynamic Selection and Pricing of OOH (DSPO), an algorithmic pipeline that uses a novel spatial-temporal state encoding and a convolutional neural network (CNN) to estimate the costs of delivery. DSPO subsequently determines optimal prices for a selected subset of OOH locations. We compared DSPO with two state-of-the-art benchmarks: a method adapted from time slot demand management literature, and proximal policy optimization (PPO).

Our insights have significant implications for both theory and practice in the field of last-mile logistics. We show that in a dynamic setting, the savings from offering OOH delivery are mainly {the result of} shorter service times and lower delivery failure rates. Furthermore, we show that using incentives for choosing OOH locations is effective: static delivery charges for home delivery and static discounts for OOH delivery can already save $4.0\%$ in total costs compared to a situation without delivery incentives. However, a balance should be struck, as offering too much discounts can potentially harm overall profitability. Our DSPO pipeline can save $19.9\%$pt in total cost compared to not offering OOH delivery, and $7\%$pt in total costs when compared to a static pricing policy. Compared to the state-of-the-art benchmarks, we can save from $3.8\%$pt up to $19.9\%$pt for the Seattle case study. Understanding the nuances of customer behavior concerning OOH delivery choices, particularly in areas with varying OOH location densities {and OOH capacities}, is vital. Managers can leverage our insights to tailor pricing strategies effectively, optimizing operational efficiency while maintaining high levels of customer satisfaction. {The limitation of our findings lies mainly in the absence of customer choice data. We assume customers to be mainly influenced by the distance they need to travel to an OOH location from their home address, although this might not be accurate. This remains a topic for further research. }

Many more avenues for further research remain unexplored. One area of interest involves problem extensions and variants to the studied problem, including scenarios with {heterogeneous parcel lockers with differing capacities}, time windows for home delivery, more refined customer segmentation, {e.g., different preferences between automated parcel lockers (open 24/7) and staffed shops (limited opening hours)}, shifting failed home deliveries to OOH locations, {a multi-day planning horizon including finite capacity and package dwell time, a combination of dynamic offering and pricing,} and the possibility of autonomously driving parcel lockers. {In general, our problem motivates different pricing schemes based on both spatial and temporal differences between customers, which might be interesting for further research. Our state-encoding could be further refined, e.g., considering an automated way to use knowledge of the spatial and arrival time distributions to define better spatial and temporal aggregations using clustering algorithms.} Additionally, studying the cost-sharing between retailers and third-party logistics providers highlights a complex arrangement: retailers manage discounts and delivery charges to guide customer behavior and aim to reduce logistics costs, but the logistics providers incur these operational expenses.

\section*{Acknowledgment}
Fabian Akkerman conducted his research in the project DynaPlex: Deep Reinforcement Learning for
Data-Driven Logistics, made possible by TKI Dinalog and the Topsector Logistics and funded by
the Ministry of Economic Affairs and Climate Policy (EZK) of the Netherlands.
The authors thank Jana Finkeldei and Alexander Reger for their help in obtaining the Seattle OOH data.


\bibliographystyle{informs2014trsc} 
\bibliography{references} 

\appendix

\section{Problem Settings}\label{appendix:problsettings}

In this section, we provide all information related to the problem settings and problem parameters of the synthetic case and the Amazon case. All parameter settings are summarized in Table~\ref{tab:params}.

\begin{table}[H]
	\caption{Problem settings.}\label{tab:params}
\centering
 \begin{adjustbox}{max width=0.93\textwidth}
		\begin{tabular}{l   c c c  c c c c c c c }
			\hline
			  &    $C^w$ & $C^f$ &  $C^m$ &  $r$ &  $l_i$ & $\mathbb{P}^m$ & V & K &  $|\mathcal{L}|$ &$a^{pricing}\in[a,b]$  \\
			\hline		
			Synthethic case  & 30 & 0.3 & 10 & 50 & Eq.~11 & 0.1 & 9 & 10 & 10 & $[-10,2]$ \\ 

   			Amazon case  & 30 & 0.3 & 10 & 50 & Eq.~11 & 0.1 & 25 & 100 & 299 & $[-10,2]$ \\ 

            \hline
		\end{tabular}
  \end{adjustbox}
\end{table}

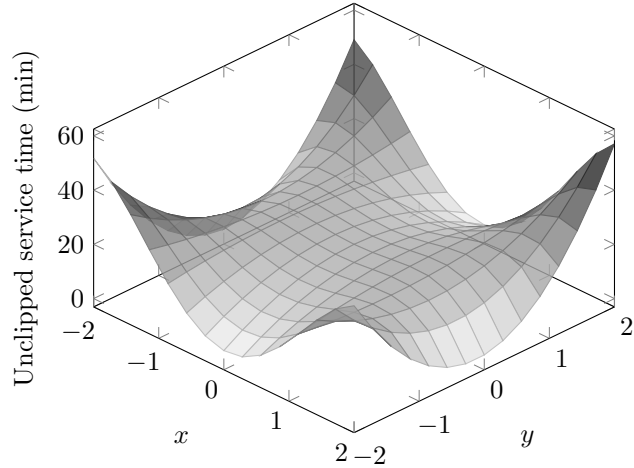
\begin{figure}[hbtp]
    \centering
    \begin{tikzpicture}
\begin{axis}[view={45}{45},samples=15,
    xlabel=$x$,
    ylabel=$y$, 
    zlabel=Unclipped service time (min)
]
    \addplot3 [surf,
        mesh/interior colormap={whiteblack}{
            color=(white) color=(black)
        },
        colormap/blackwhite,
        y domain=-2:2,
         domain=-2:2,
          opacity=0.8,
    ] {
             (4-2.1*x^2+(y)/3) * x^2+
            x*y +
            (-4+4*x^2) * y^2+20
    };
\end{axis}
\end{tikzpicture} 
    \caption{Six-hump camel function for obtaining service times per spatial area \citep{sixhump}. }
    \label{fig:camel}
\end{figure}

Figure~\ref{fig:camel} illustrates the domain of the six-hump camelback function used for both the synthetic and Amazon case to obtain service times $l_i$ per location $i$. The service area is projected onto this function domain (see Equation~11 in the main manuscript) to obtain service durations. We ensure that service times are not too low or high by clipping, $l_i\in[1,10]$.

\begin{figure}[hbtp]
    \centering
     \begin{subfigure}{1.0\textwidth}
     \centering
    \fbox{\includegraphics[scale=0.15]{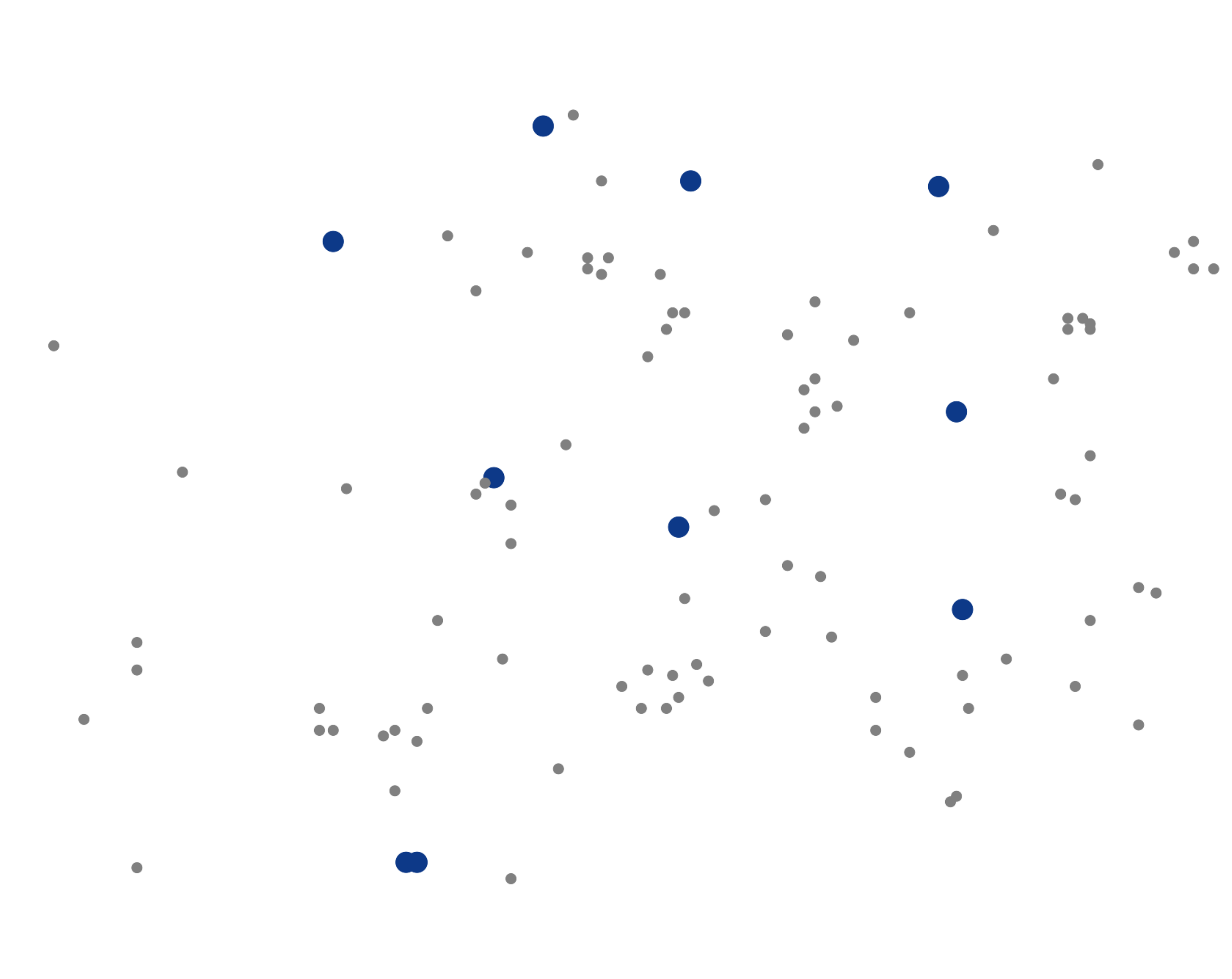}}
    \end{subfigure}\\
    \begin{subfigure}{0.39\textwidth}
    \centering
    \includegraphics[width=\textwidth]{img/legend.pdf}
    \end{subfigure}
   
    \caption{Gehring and Homberger (2002) RC instance map with customer and OOH locations.}
    \label{fig:homberger}
\end{figure}

Figure~\ref{fig:homberger} illustrates an RC instance as used from \citet{Gehring2002}. Here, we indicate the OOH delivery locations in blue and customer locations in grey.

\newpage
\section{Choice Model Tuning}\label{appendix:mnl}

We denote all MNL parameters without customer segments. {Since we do not have detailed customer segment data, we use a tuning procedure to obtain realistic behavior with a single segment. Note that we do conduct a sensitivity analysis where we do consider multiple segments, see Section~5.2.3. For the other experiments, we used this tuning procedure and assume a single customer segment.}

The utility of home delivery, $u_{k}$, with $k=h$, is always the same if we would use Equation~5. Therefore, we replace it with a separate utility term for home delivery, denoted by $u_{0}^+$. We use a simple iterative procedure where we tune three parameters: the utility of home delivery, $u_{0}^+$, and the sensitivity to distance and pricing,  $\beta^k$ and $\beta^d$, respectively. Our data on customer preferences is limited; however, it includes information on the proportion of OOH deliveries in countries where OOH providers are well-established, as well as the OOH market share in countries where OOH services are still gaining adoption, see, for instance, \citet{loq} and \citet{lme_ooh}. For $u_{0}^+$, $\beta^k$, and $\beta^d$ we set a range of possible values $[-5.0,5.0]$, with $0.01$ increments. First, we tune the utility of home delivery using a simulation of $100$ replications where all possible values for $u_{0}^+$ are tested. During this simulation, we use the NoPricing baseline. We store the percentage of customers selecting home delivery, and next, choose the value of $u_{0}^+$ {and $\beta^k$} that yield (close to) $80\%$ of the customers choosing home delivery. Next, we use the StaticPricing benchmark and select the values that yields (close to) $60\%$ home deliveries. The $80\%$ and $60\%$ rates of home deliveries correspond to countries where OOH delivery options are gaining popularity and to those where OOH delivery is already well-established, respectively. All MNL parameter values are summarized in Table~\ref{tab:paramsmnl}.

\begin{table}[htbp]
	\caption{Multinomial choice model parameter settings for a single segment.}\label{tab:paramsmnl}
\centering
 \begin{adjustbox}{max width=0.93\textwidth}
		\begin{tabular}{l c  c c c  c  }
			\hline
			   & $\beta^k$ &  $u_{0}^+$ & $\beta^d$ &  Gumbel $\mu$ &  Gumbel $\beta$   \\
	   \hline
            Synthetic case & 0.02 & 3.2 & -0.25 & 0 & 1.0 \\ 

   			Amazon case & 0.018 & 3.55 & -0.18 & 0 & 1.0 \\ 

            \hline
		\end{tabular}
  \end{adjustbox}
\end{table}

{In Section~5.2.3, we conduct a sensitivity analysis using three segments $g$. In Table~\ref{tab:paramsmnl_sens}, we define the relevant MNL parameters for the three segments. Segment 1 only considers home delivery, so it is not modeled using the MNL parameters. }

\begin{table}[htbp]
	\caption{Multinomial choice model parameter settings for three segments.}\label{tab:paramsmnl_sens}
\centering
 \begin{adjustbox}{max width=0.93\textwidth}
		\begin{tabular}{l c  c c  c c }
			\hline
			   & $\beta^k$ &  $u_{0}^+$ & $\beta^d$ &  Gumbel $\mu$ &  Gumbel $\beta$     \\
	   \hline
            Segment 1 & n/a & n/a & n/a / & n/a & n/a   \\ 

   			Segment 2 & 0.015 & 3.1 & -0.18 & 0 & 1.0  \\ 

            Segment 3 & 0.05 & 1.0 & -0.35 & 0 & 1.0  \\ 

            \hline
		\end{tabular}
  \end{adjustbox}
\end{table}

\vspace{-1em}
\section{{Choice Model Validation}}\label{app:choice_validation}
\begin{figure}[hbtp]
\centering
\begin{subfigure}{.5\textwidth}
     \centering
    \centering
    \begin{tikzpicture}[scale = 0.84]
            \begin{axis}[
                legend style={at={(0.5,-0.2)},anchor=north, draw=none},
                legend columns=1,
                ymin = 80 ,ymax=100.1,
                xmin=-0.1, xmax=15.1,
                 xlabel = OOH density as $\%$ of customer density,
                xlabel near ticks,
                ylabel= Percentage home delivery,
                ylabel near ticks,
                  yticklabel={$\pgfmathprintnumber{\tick}\%$},
                   xticklabel={\pgfmathparse{\tick}\pgfmathprintnumber{\pgfmathresult}\%},
                   xtick={0,3,6,9,12,15},
                  xticklabel shift={0.1cm},
                  legend cell align={left},
                        ]
                
                 \addplot[mark=*,mark options={fill=black, mark size=2.5pt}] %
                coordinates {
                   (0,100)
                    (1,98.4269662921348)
                    (2,97.4531835205992)
                    (3,95.8052434456928)
                    (4,93.5955056179775)
                    (5,92.5468164794007)
                    (6,90.5992509363295)
                    (7,90.0617977528089)
                    (8,88.5393258426966)
                    (9,87.3033707865168)
                    (10,86.1048689138576)
                    (11,85.2059925093632)
                    (12,83.4951310861423)
                    (13,82.8516779026217)
                    (14,82.8464419475655)
                    (15,82.8239700374531)

                };

            \end{axis}
                        \node at (3.5,6.0) {Sensitivity of choice model to OOH density};
                        \node at (3.5,-2) {};
    \end{tikzpicture}

\end{subfigure}%
\begin{subfigure}{.5\textwidth}
    \centering
    \begin{tikzpicture}[scale = 0.84]
            \begin{axis}[
                legend style={at={(0.5,-0.2)},anchor=north, draw=none},
                legend columns=2,
                ymin = 40 ,ymax=100,
                xmin=-0.05, xmax=3.05,
                 xlabel = Fixed OOH discount as $\%$ of revenue,
                xlabel near ticks,
                ylabel= Percentage home delivery,
                ylabel near ticks,
                legend cell align={left},
                  yticklabel={$\pgfmathprintnumber{\tick}\%$},
                   xticklabel={\pgfmathparse{\tick*2}\pgfmathprintnumber{\pgfmathresult}\%},
                  xticklabel shift={0.1cm},
                  legend cell align={left},
                        ]

        \addplot[ mark options={fill=black, mark size=2.5pt},mark=diamond*] %
                coordinates {
                    (0,85.5128205)
                    (0.25,83.4615385)
                    (0.5,82.6923077)
                    (0.75,82.4358974)
                    (1,81.6666667)
                    (1.25,80.3846154)
                    (1.5,78.5897436)
                    (1.75,77.5641026)
                    (2,76.5384615)
                    (2.25,74.7435897)
                    (2.5,73.4615385)
                    (2.75,72.1794872)
                    (3,69.8717949)

                };

 \addplot[mark options={fill=black}, mark=square*] %
                coordinates {
                (0,80.3846154)
                (0.25,78.3333333)
                (0.5,77.5641026)
                (0.75,77.3076923)
                (1,76.5384615)
                (1.25,75.2564103)
                (1.5,73.4615385)
                (1.75,72.4358974)
                (2,71.4102564)
                (2.25,69.6153846)
                (2.5,68.3333333)
                (2.75,67.0512821)
                (3,64.7435897)

                };

                 \addplot[mark options={fill=white}, mark=square*] %
                coordinates {
                    (0,73.5384615)
                    (0.25,71.4871795)
                    (0.5,70.7179487)
                    (0.75,70.4615385)
                    (1,69.6923077)
                    (1.25,68.4102564)
                    (1.5,66.6153846)
                    (1.75,65.5897436)
                    (2,64.5641026)
                    (2.25,62.7692308)
                    (2.5,61.4871795)
                    (2.75,60.2051282)
                    (3,57.8974359)
                
                };

                 \addplot[mark options={mark size=2.5pt,fill=white}, mark=diamond*] %
                coordinates {
                    (0,64.8717949)
                    (0.25,62.8205128)
                    (0.5,62.0512821)
                    (0.75,61.7948718)
                    (1,61.025641)
                    (1.25,59.7435897)
                    (1.5,57.9487179)
                    (1.75,56.9230769)
                    (2,55.8974359)
                    (2.25,54.1025641)
                    (2.5,52.8205128)
                    (2.75,51.5384615)
                    (3,49.2307692)

                };

                \addlegendentry{Home del. charge $0\%$}
                \addlegendentry{Home del. charge $2\%$}
                \addlegendentry{Home del. charge $4\%$}
                \addlegendentry{Home del. charge $6\%$}

            \end{axis}
            \node at (3.5,6.0) {Sensitivity of choice model to discounts};
    \end{tikzpicture}

\end{subfigure}
  \caption{Sensitivity analysis of the MNL choice model, results based on the RC instances.}\label{fig:sensitivityMNL}
\end{figure}
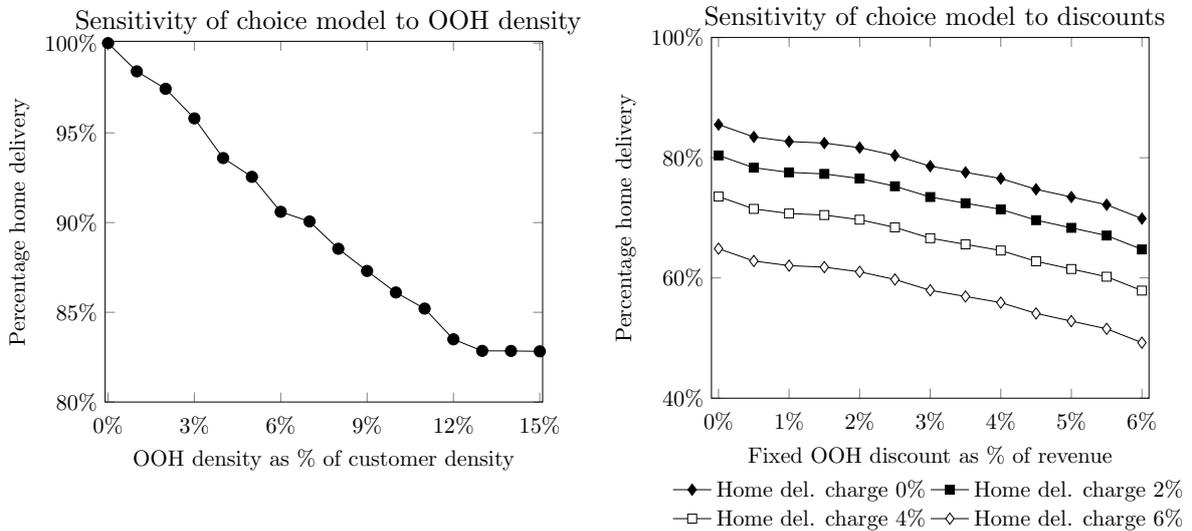

In this section, we validate the choice model tuning procedure by showing the effect of pricing on customer choice. {All graphs in this section are shown for the single segment model, which is used for most experiments.} Figure~\ref{fig:sensitivityMNL} shows a sensitivity analysis of the MNL choice model. The left graph shows the percentage of customers choosing home delivery in relation to the relative OOH density. The relative OOH density is calculated using the number of OOH locations divided by the number of customers in the service region. We do not provide discounts in this setting. The graph clearly shows that more customers choose OOH delivery as the OOH density increases. However, as the OOH density approaches $12\%$, the area is saturated, and the percentage of home deliveries only decreases slightly thereafter. This aligns with the results in \citet{ENTHOVEN2020104919}. Note that our choice model calculates utility based on the distance from customer homes to OOH locations.


The right graph of Figure~\ref{fig:sensitivityMNL} shows the percentage of home delivery for different fixed discounts as a percentage of revenue. Each line in the graph represents different home delivery charges as a percentage of revenue, i.e., the costs that a customer has to pay when choosing for home delivery. The graph validates that our MNL model is sensitive to discounts and charges. 

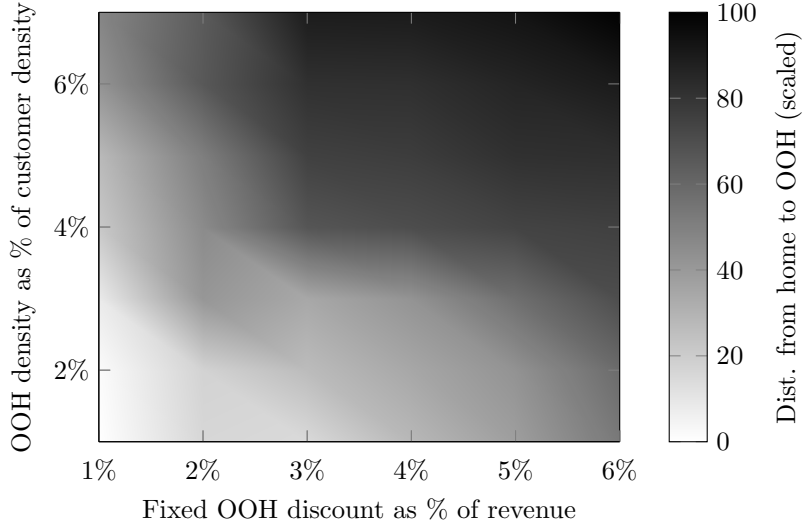
\begin{figure}[hbtp]
    \centering
   \begin{tikzpicture}[scale = 1.0]

\begin{scope}[local bounding box=plots]

    \begin{axis}[ymin = 1 ,ymax=7,
                xmin=1, xmax=6,
                 xlabel = Fixed OOH discount as $\%$ of revenue,
                xlabel near ticks,
                ylabel= OOH density as $\%$ of customer density ,
                ylabel near ticks,
                  view={0}{90},
                 shader=interp,
                mesh/cols=10,
                  colormap = {whiteblack}{color(0cm)  = (white);color(0.001cm) = (black)},
                  colorbar,
                  colorbar style={
                     ylabel= Dist. from home to OOH (scaled)},
                  ylabel near ticks, yticklabel pos=left,
                    yticklabel={$\pgfmathprintnumber{\tick}\%$},
                   xticklabel={\pgfmathparse{\tick}\pgfmathprintnumber{\pgfmathresult}\%},
                  xticklabel shift={0.1cm},
                ]

    \addplot3[surf] coordinates {
            (1,1,0)
            (1,2,1.54289658532625)
            (1,3,7.93752622247577)
            (1,4,21.4529500134928)
            (1,5,27.1744691217668)
            (1,6,42.019570425109)
            (1,7,47.555320963092)
            (1,8,58.5272104700423)
            (1,9,61.4587139821621)
            (1,10,61.4587139821621)
            (2,1,16.0474779054506)
            (2,2,17.6273228037411)
            (2,3,41.2536170097588)
            (2,4,45.7437318831041)
            (2,5,51.0517411723312)
            (2,6,65.9876568273176)
            (2,7,65.0458132012399)
            (2,8,62.6024875823895)
            (2,9,62.6004574553035)
            (2,10,62.6004574553035)
            (3,1,14.9760219434184)
            (3,2,26.5558668417089)
            (3,3,34.3632844749419)
            (3,4,67.0781057560868)
            (3,5,75.2588682718199)
            (3,6,81.6800101383317)
            (3,7,89.6058969655137)
            (3,8,62.9945093229504)
            (3,9,58.0013804864186)
            (3,10,58.0013804864186)
            (4,1,28.067657368685)
            (4,2,34.6475022669756)
            (4,3,42.3685589384567)
            (4,4,70.3831526520232)
            (4,5,76.7918713711483)
            (4,6,82.0155420160676)
            (4,7,89.9414288432496)
            (4,8,63.9940584947286)
            (4,9,65.5804520029621)
            (4,10,65.5804520029621)
            (5,1,38.0907196123815)
            (5,2,42.670564510672)
            (5,3,57.0086754097477)
            (5,4,74.1776631883828)
            (5,5,81.7653985139473)
            (5,6,86.3327089320837)
            (5,7,93.1263262125253)
            (5,8,63.3378829001876)
            (5,9,66.5533756722115)
            (5,10,66.5533756722115)
            (6,1,53.0069159662732)
            (6,2,58.5867608645638)
            (6,3,70.2532245185217)
            (6,4,75.5531097580036)
            (6,5,82.3751788367331)
            (6,6,90.1788775734939)
            (6,7,100)
            (6,8,80.1047644006759)
            (6,9,75.0272771330803)
            (6,10,75.0272771330803)

    };

    \end{axis}

\end{scope}

\end{tikzpicture}

  \caption{Analysis of the distance traveled to OOH locations by customers given OOH density and OOH discount, results based on the RC instances.}\label{fig:analysis_traveldist}

\end{figure}

Finally, we analyze the effect of discounts and OOH density on the willingness of {the modelled} customers to accept an OOH delivery further away from their homes. Figure~\ref{fig:analysis_traveldist} shows how far customers are willing to travel, given OOH density (y-axis) and fixed discounts (x-axis). A darker color indicates a further distance traveled by the customer. We observe in the lower-left corner that when both OOH density and discounts are low, customers are less inclined to travel to OOH locations, as indicated by the lighter color.


Moving to the right from the left-lower corner, we observe the color shifts from light to a darker color. This indicates that discounts can convince {modelled} customers to accept a more remote OOH location.


When moving up from the lower-left corner, we observe that the color also gets darker without providing higher discounts. This indicates that a higher OOH density requires lower discounts to reach the same OOH utilization compared to an area with lower OOH density. This aligns with the observations from \citet{lyu2022}.


\section{DSPO Implementation details}\label{appendix:dspo]}

For the selection decision, $a^{selection}$, we use the heuristic rule to offer the $N=20$ closest OOH locations to the customer's home address. For the synthetic case, this means that we always offer all OOH available locations, as $|\mathcal{L}|<N$.

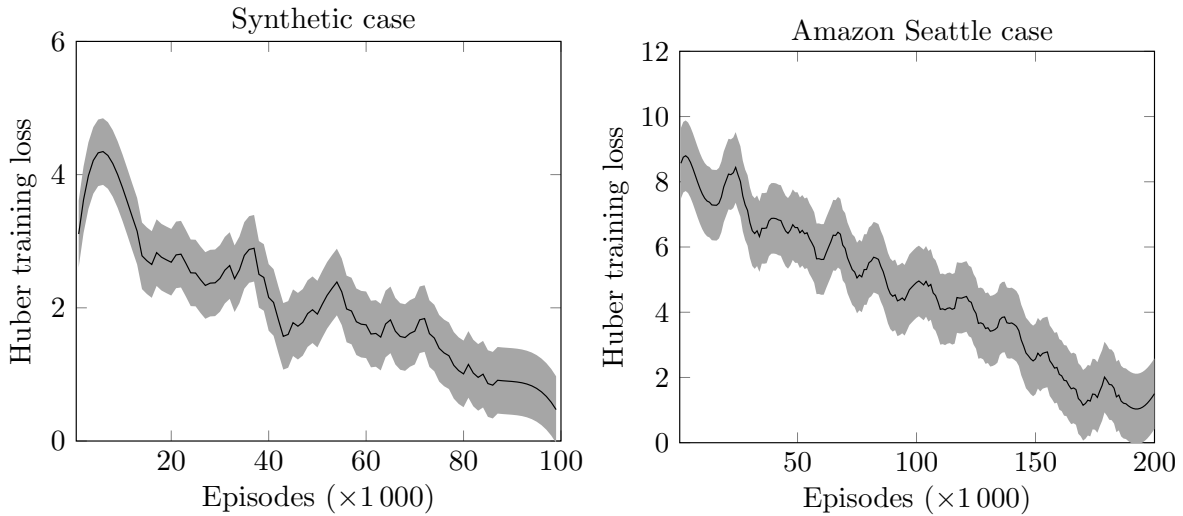
\begin{figure}[H]
    \centering
    \begin{subfigure}[t]{.49\textwidth}
     \centering
    \centering
    \begin{adjustbox}{width=\textwidth}
   \begin{tikzpicture}[scale = 0.84]
            \begin{axis}[
                ymin = 0 ,ymax=6,
                xmin=0.5, xmax=100,
                 xlabel = Episodes ($\times 1\,000$),
                xlabel near ticks,
                 ylabel= Huber training loss,
                ylabel near ticks
                        ]

        \addplot[ name path = A] %
                coordinates {
(1,3.10818)
(2,3.62331)
(3,3.98469)
(4,4.2128)
(5,4.32713)
(6,4.34613)
(7,4.28726)
(8,4.16698)
(9,4.00074)
(10,3.80296)
(11,3.58708)
(12,3.36552)
(13,3.14971)
(14,2.78091)
(15,2.70709)
(16,2.65046)
(17,2.82875)
(18,2.76)
(19,2.72058)
(20,2.68429)
(21,2.79484)
(22,2.80556)
(23,2.66955)
(24,2.52566)
(25,2.52213)
(26,2.42293)
(27,2.33556)
(28,2.37071)
(29,2.37527)
(30,2.44119)
(31,2.56503)
(32,2.63469)
(33,2.43724)
(34,2.57052)
(35,2.78722)
(36,2.87621)
(37,2.89419)
(38,2.50268)
(39,2.45609)
(40,2.156)
(41,2.08098)
(42,1.81907)
(43,1.57168)
(44,1.6004)
(45,1.77673)
(46,1.72388)
(47,1.7816)
(48,1.91352)
(49,1.97113)
(50,1.9057)
(51,2.05241)
(52,2.18797)
(53,2.2915)
(54,2.38757)
(55,2.24069)
(56,1.98288)
(57,1.95204)
(58,1.79089)
(59,1.75418)
(60,1.7449)
(61,1.60644)
(62,1.61521)
(63,1.5601)
(64,1.75731)
(65,1.81889)
(66,1.64925)
(67,1.57001)
(68,1.55373)
(69,1.61514)
(70,1.64968)
(71,1.82039)
(72,1.83883)
(73,1.61219)
(74,1.54632)
(75,1.39197)
(76,1.32311)
(77,1.28067)
(78,1.13622)
(79,1.05711)
(80,1.0079)
(81,1.15006)
(82,1.02075)
(83,0.955492)
(84,1.00352)
(85,0.86284)
(86,0.838437)
(87,0.911998)
(88,0.904981)
(89,0.899577)
(90,0.893963)
(91,0.886191)
(92,0.874192)
(93,0.855769)
(94,0.828604)
(95,0.790253)
(96,0.738148)
(97,0.669598)
(98,0.581788)
(99,0.471777)

                };

 \addplot[name path = B,draw=none,forget plot] %
                coordinates {
(1,2.60818)
(2,3.12331)
(3,3.48469)
(4,3.7128)
(5,3.82713)
(6,3.84613)
(7,3.78726)
(8,3.66698)
(9,3.50074)
(10,3.30296)
(11,3.08708)
(12,2.86552)
(13,2.64971)
(14,2.28091)
(15,2.20709)
(16,2.15046)
(17,2.32875)
(18,2.26)
(19,2.22058)
(20,2.18429)
(21,2.29484)
(22,2.30556)
(23,2.16955)
(24,2.02566)
(25,2.02213)
(26,1.92293)
(27,1.83556)
(28,1.87071)
(29,1.87527)
(30,1.94119)
(31,2.06503)
(32,2.13469)
(33,1.93724)
(34,2.07052)
(35,2.28722)
(36,2.37621)
(37,2.39419)
(38,2.00268)
(39,1.95609)
(40,1.656)
(41,1.58098)
(42,1.31907)
(43,1.07168)
(44,1.1004)
(45,1.27673)
(46,1.22388)
(47,1.2816)
(48,1.41352)
(49,1.47113)
(50,1.4057)
(51,1.55241)
(52,1.68797)
(53,1.7915)
(54,1.88757)
(55,1.74069)
(56,1.48288)
(57,1.45204)
(58,1.29089)
(59,1.25418)
(60,1.2449)
(61,1.10644)
(62,1.11521)
(63,1.0601)
(64,1.25731)
(65,1.31889)
(66,1.14925)
(67,1.07001)
(68,1.05373)
(69,1.11514)
(70,1.14968)
(71,1.32039)
(72,1.33883)
(73,1.11219)
(74,1.04632)
(75,0.89197)
(76,0.82311)
(77,0.78067)
(78,0.63622)
(79,0.55711)
(80,0.5079)
(81,0.65006)
(82,0.52075)
(83,0.455492)
(84,0.50352)
(85,0.36284)
(86,0.338437)
(87,0.411998)
(88,0.404981)
(89,0.399577)
(90,0.393963)
(91,0.386191)
(92,0.374192)
(93,0.355769)
(94,0.328604)
(95,0.290253)
(96,0.238148)
(97,0.169598)
(98,0.081788)
(99,-0.028223)

                };

             \addplot[name path = C,draw=none,forget plot] %
                coordinates {

(1,3.60818)
(2,4.12331)
(3,4.48469)
(4,4.7128)
(5,4.82713)
(6,4.84613)
(7,4.78726)
(8,4.66698)
(9,4.50074)
(10,4.30296)
(11,4.08708)
(12,3.86552)
(13,3.64971)
(14,3.28091)
(15,3.20709)
(16,3.15046)
(17,3.32875)
(18,3.26)
(19,3.22058)
(20,3.18429)
(21,3.29484)
(22,3.30556)
(23,3.16955)
(24,3.02566)
(25,3.02213)
(26,2.92293)
(27,2.83556)
(28,2.87071)
(29,2.87527)
(30,2.94119)
(31,3.06503)
(32,3.13469)
(33,2.93724)
(34,3.07052)
(35,3.28722)
(36,3.37621)
(37,3.39419)
(38,3.00268)
(39,2.95609)
(40,2.656)
(41,2.58098)
(42,2.31907)
(43,2.07168)
(44,2.1004)
(45,2.27673)
(46,2.22388)
(47,2.2816)
(48,2.41352)
(49,2.47113)
(50,2.4057)
(51,2.55241)
(52,2.68797)
(53,2.7915)
(54,2.88757)
(55,2.74069)
(56,2.48288)
(57,2.45204)
(58,2.29089)
(59,2.25418)
(60,2.2449)
(61,2.10644)
(62,2.11521)
(63,2.0601)
(64,2.25731)
(65,2.31889)
(66,2.14925)
(67,2.07001)
(68,2.05373)
(69,2.11514)
(70,2.14968)
(71,2.32039)
(72,2.33883)
(73,2.11219)
(74,2.04632)
(75,1.89197)
(76,1.82311)
(77,1.78067)
(78,1.63622)
(79,1.55711)
(80,1.5079)
(81,1.65006)
(82,1.52075)
(83,1.455492)
(84,1.50352)
(85,1.36284)
(86,1.338437)
(87,1.411998)
(88,1.404981)
(89,1.399577)
(90,1.393963)
(91,1.386191)
(92,1.374192)
(93,1.355769)
(94,1.328604)
(95,1.290253)
(96,1.238148)
(97,1.169598)
(98,1.081788)
(99,0.971777)
      
                };

        \addplot [black!35] fill between [of = B and C];

            \end{axis}
            \node at (3.5,6.0) {Synthetic case};
    \end{tikzpicture}
    \end{adjustbox}

\end{subfigure}%
\begin{subfigure}[t]{.49\textwidth}
     \centering
    \centering
    \begin{adjustbox}{width=\textwidth}
   \begin{tikzpicture}[scale = 0.84]
                 \begin{axis}[
                ymin = 0 ,ymax=12,
                xmin=0.5, xmax=200,
                 xlabel = Episodes ($\times 1\,000$),
                xlabel near ticks,
                 ylabel= Huber training loss,
                ylabel near ticks
                        ]

        \addplot[ name path = A] %
                coordinates {
(1,8.56768)
(2,8.75307)
(3,8.79768)
(4,8.73587)
(5,8.59855)
(6,8.41329)
(7,8.20421)
(8,7.99208)
(9,7.79423)
(10,7.62462)
(11,7.49381)
(12,7.40895)
(13,7.3738)
(14,7.29006)
(15,7.28266)
(16,7.2793)
(17,7.34533)
(18,7.53286)
(19,7.84288)
(20,8.03896)
(21,8.23468)
(22,8.21776)
(23,8.27474)
(24,8.43991)
(25,8.21223)
(26,8.00689)
(27,7.56583)
(28,7.29279)
(29,7.12266)
(30,6.74374)
(31,6.49787)
(32,6.41262)
(33,6.4971)
(34,6.32006)
(35,6.57265)
(36,6.57079)
(37,6.58482)
(38,6.80079)
(39,6.87614)
(40,6.88121)
(41,6.86798)
(42,6.82997)
(43,6.81072)
(44,6.63661)
(45,6.49)
(46,6.41856)
(47,6.51944)
(48,6.68394)
(49,6.59646)
(50,6.59621)
(51,6.43713)
(52,6.51083)
(53,6.40655)
(54,6.43758)
(55,6.25751)
(56,6.1294)
(57,5.98416)
(58,5.63827)
(59,5.6418)
(60,5.61761)
(61,5.62087)
(62,5.81196)
(63,5.98982)
(64,6.16204)
(65,6.36322)
(66,6.35279)
(67,6.45213)
(68,6.39315)
(69,6.05783)
(70,5.95745)
(71,5.66671)
(72,5.48515)
(73,5.30719)
(74,5.23355)
(75,5.05062)
(76,5.14054)
(77,5.08379)
(78,5.30879)
(79,5.30997)
(80,5.51592)
(81,5.57443)
(82,5.68635)
(83,5.66081)
(84,5.62078)
(85,5.40952)
(86,5.24333)
(87,5.00181)
(88,4.73925)
(89,4.56501)
(90,4.48862)
(91,4.53354)
(92,4.34929)
(93,4.38278)
(94,4.43675)
(95,4.3676)
(96,4.52348)
(97,4.6876)
(98,4.76766)
(99,4.84335)
(100,4.91278)
(101,4.95845)
(102,4.89867)
(103,4.83677)
(104,4.94642)
(105,4.79649)
(106,4.85256)
(107,4.59237)
(108,4.51258)
(109,4.32101)
(110,4.08845)
(111,4.10521)
(112,4.07391)
(113,4.11916)
(114,4.1356)
(115,4.08988)
(116,4.11884)
(117,4.45056)
(118,4.43478)
(119,4.43224)
(120,4.46077)
(121,4.48398)
(122,4.34558)
(123,4.30357)
(124,4.08855)
(125,3.96306)
(126,3.65788)
(127,3.67563)
(128,3.63854)
(129,3.50083)
(130,3.52516)
(131,3.41688)
(132,3.43551)
(133,3.47144)
(134,3.53585)
(135,3.77218)
(136,3.84213)
(137,3.85658)
(138,3.67733)
(139,3.6584)
(140,3.67124)
(141,3.62767)
(142,3.56593)
(143,3.43547)
(144,3.15184)
(145,2.96434)
(146,2.75164)
(147,2.66514)
(148,2.56911)
(149,2.50785)
(150,2.56791)
(151,2.74341)
(152,2.65559)
(153,2.73742)
(154,2.76391)
(155,2.78317)
(156,2.51489)
(157,2.41042)
(158,2.28635)
(159,2.30832)
(160,2.09577)
(161,2.05063)
(162,1.91196)
(163,1.89713)
(164,1.75915)
(165,1.65348)
(166,1.67901)
(167,1.63122)
(168,1.35429)
(169,1.27907)
(170,1.14764)
(171,1.20374)
(172,1.30326)
(173,1.27555)
(174,1.49176)
(175,1.48)
(176,1.38794)
(177,1.58333)
(178,1.75578)
(179,2.01158)
(180,1.91218)
(181,1.78646)
(182,1.77255)
(183,1.67168)
(184,1.40568)
(185,1.4599)
(186,1.27491)
(187,1.21028)
(188,1.19412)
(189,1.13164)
(190,1.08311)
(191,1.05008)
(192,1.03372)
(193,1.03474)
(194,1.05347)
(195,1.08982)
(196,1.1433)
(197,1.21299)
(198,1.29756)
(199,1.39528)
(200,1.50401)

                };

 \addplot[name path = C,draw=none,forget plot] %
                coordinates {
(1,7.48568)
(2,7.67107)
(3,7.71568)
(4,7.65387)
(5,7.51655)
(6,7.33129)
(7,7.12221)
(8,6.91008)
(9,6.71223)
(10,6.54262)
(11,6.41181)
(12,6.32695)
(13,6.2918)
(14,6.20806)
(15,6.20066)
(16,6.1973)
(17,6.26333)
(18,6.45086)
(19,6.76088)
(20,6.95696)
(21,7.15268)
(22,7.13576)
(23,7.19274)
(24,7.35791)
(25,7.13023)
(26,6.92489)
(27,6.48383)
(28,6.21079)
(29,6.04066)
(30,5.66174)
(31,5.41587)
(32,5.33062)
(33,5.4151)
(34,5.23806)
(35,5.49065)
(36,5.48879)
(37,5.50282)
(38,5.71879)
(39,5.79414)
(40,5.79921)
(41,5.78598)
(42,5.74797)
(43,5.72872)
(44,5.55461)
(45,5.408)
(46,5.33656)
(47,5.43744)
(48,5.60194)
(49,5.51446)
(50,5.51421)
(51,5.35513)
(52,5.42883)
(53,5.32455)
(54,5.35558)
(55,5.17551)
(56,5.0474)
(57,4.90216)
(58,4.55627)
(59,4.5598)
(60,4.53561)
(61,4.53887)
(62,4.72996)
(63,4.90782)
(64,5.08004)
(65,5.28122)
(66,5.27079)
(67,5.37013)
(68,5.31115)
(69,4.97583)
(70,4.87545)
(71,4.58471)
(72,4.40315)
(73,4.22519)
(74,4.15155)
(75,3.96862)
(76,4.05854)
(77,4.00179)
(78,4.22679)
(79,4.22797)
(80,4.43392)
(81,4.49243)
(82,4.60435)
(83,4.57881)
(84,4.53878)
(85,4.32752)
(86,4.16133)
(87,3.91981)
(88,3.65725)
(89,3.48301)
(90,3.40662)
(91,3.45154)
(92,3.26729)
(93,3.30078)
(94,3.35475)
(95,3.2856)
(96,3.44148)
(97,3.6056)
(98,3.68566)
(99,3.76135)
(100,3.83078)
(101,3.87645)
(102,3.81667)
(103,3.75477)
(104,3.86442)
(105,3.71449)
(106,3.77056)
(107,3.51037)
(108,3.43058)
(109,3.23901)
(110,3.00645)
(111,3.02321)
(112,2.99191)
(113,3.03716)
(114,3.0536)
(115,3.00788)
(116,3.03684)
(117,3.36856)
(118,3.35278)
(119,3.35024)
(120,3.37877)
(121,3.40198)
(122,3.26358)
(123,3.22157)
(124,3.00655)
(125,2.88106)
(126,2.57588)
(127,2.59363)
(128,2.55654)
(129,2.41883)
(130,2.44316)
(131,2.33488)
(132,2.35351)
(133,2.38944)
(134,2.45385)
(135,2.69018)
(136,2.76013)
(137,2.77458)
(138,2.59533)
(139,2.5764)
(140,2.58924)
(141,2.54567)
(142,2.48393)
(143,2.35347)
(144,2.06984)
(145,1.88234)
(146,1.66964)
(147,1.58314)
(148,1.48711)
(149,1.42585)
(150,1.48591)
(151,1.66141)
(152,1.57359)
(153,1.65542)
(154,1.68191)
(155,1.70117)
(156,1.43289)
(157,1.32842)
(158,1.20435)
(159,1.22632)
(160,1.01377)
(161,0.96863)
(162,0.82996)
(163,0.81513)
(164,0.67715)
(165,0.57148)
(166,0.59701)
(167,0.54922)
(168,0.27229)
(169,0.19707)
(170,0.0656399999999999)
(171,0.12174)
(172,0.22126)
(173,0.19355)
(174,0.40976)
(175,0.398)
(176,0.30594)
(177,0.50133)
(178,0.67378)
(179,0.92958)
(180,0.83018)
(181,0.70446)
(182,0.69055)
(183,0.58968)
(184,0.32368)
(185,0.3779)
(186,0.19291)
(187,0.12828)
(188,0.11212)
(189,0.0496399999999999)
(190,0.00110999999999994)
(191,-0.0319200000000002)
(192,-0.0482800000000001)
(193,-0.0472600000000001)
(194,-0.0285300000000002)
(195,0.00781999999999994)
(196,0.0612999999999999)
(197,0.13099)
(198,0.21556)
(199,0.31328)
(200,0.42201)

                };

                      \addplot[name path = B,draw=none,forget plot] %
                coordinates {
(1,9.64968)
(2,9.83507)
(3,9.87968)
(4,9.81787)
(5,9.68055)
(6,9.49529)
(7,9.28621)
(8,9.07408)
(9,8.87623)
(10,8.70662)
(11,8.57581)
(12,8.49095)
(13,8.4558)
(14,8.37206)
(15,8.36466)
(16,8.3613)
(17,8.42733)
(18,8.61486)
(19,8.92488)
(20,9.12096)
(21,9.31668)
(22,9.29976)
(23,9.35674)
(24,9.52191)
(25,9.29423)
(26,9.08889)
(27,8.64783)
(28,8.37479)
(29,8.20466)
(30,7.82574)
(31,7.57987)
(32,7.49462)
(33,7.5791)
(34,7.40206)
(35,7.65465)
(36,7.65279)
(37,7.66682)
(38,7.88279)
(39,7.95814)
(40,7.96321)
(41,7.94998)
(42,7.91197)
(43,7.89272)
(44,7.71861)
(45,7.572)
(46,7.50056)
(47,7.60144)
(48,7.76594)
(49,7.67846)
(50,7.67821)
(51,7.51913)
(52,7.59283)
(53,7.48855)
(54,7.51958)
(55,7.33951)
(56,7.2114)
(57,7.06616)
(58,6.72027)
(59,6.7238)
(60,6.69961)
(61,6.70287)
(62,6.89396)
(63,7.07182)
(64,7.24404)
(65,7.44522)
(66,7.43479)
(67,7.53413)
(68,7.47515)
(69,7.13983)
(70,7.03945)
(71,6.74871)
(72,6.56715)
(73,6.38919)
(74,6.31555)
(75,6.13262)
(76,6.22254)
(77,6.16579)
(78,6.39079)
(79,6.39197)
(80,6.59792)
(81,6.65643)
(82,6.76835)
(83,6.74281)
(84,6.70278)
(85,6.49152)
(86,6.32533)
(87,6.08381)
(88,5.82125)
(89,5.64701)
(90,5.57062)
(91,5.61554)
(92,5.43129)
(93,5.46478)
(94,5.51875)
(95,5.4496)
(96,5.60548)
(97,5.7696)
(98,5.84966)
(99,5.92535)
(100,5.99478)
(101,6.04045)
(102,5.98067)
(103,5.91877)
(104,6.02842)
(105,5.87849)
(106,5.93456)
(107,5.67437)
(108,5.59458)
(109,5.40301)
(110,5.17045)
(111,5.18721)
(112,5.15591)
(113,5.20116)
(114,5.2176)
(115,5.17188)
(116,5.20084)
(117,5.53256)
(118,5.51678)
(119,5.51424)
(120,5.54277)
(121,5.56598)
(122,5.42758)
(123,5.38557)
(124,5.17055)
(125,5.04506)
(126,4.73988)
(127,4.75763)
(128,4.72054)
(129,4.58283)
(130,4.60716)
(131,4.49888)
(132,4.51751)
(133,4.55344)
(134,4.61785)
(135,4.85418)
(136,4.92413)
(137,4.93858)
(138,4.75933)
(139,4.7404)
(140,4.75324)
(141,4.70967)
(142,4.64793)
(143,4.51747)
(144,4.23384)
(145,4.04634)
(146,3.83364)
(147,3.74714)
(148,3.65111)
(149,3.58985)
(150,3.64991)
(151,3.82541)
(152,3.73759)
(153,3.81942)
(154,3.84591)
(155,3.86517)
(156,3.59689)
(157,3.49242)
(158,3.36835)
(159,3.39032)
(160,3.17777)
(161,3.13263)
(162,2.99396)
(163,2.97913)
(164,2.84115)
(165,2.73548)
(166,2.76101)
(167,2.71322)
(168,2.43629)
(169,2.36107)
(170,2.22964)
(171,2.28574)
(172,2.38526)
(173,2.35755)
(174,2.57376)
(175,2.562)
(176,2.46994)
(177,2.66533)
(178,2.83778)
(179,3.09358)
(180,2.99418)
(181,2.86846)
(182,2.85455)
(183,2.75368)
(184,2.48768)
(185,2.5419)
(186,2.35691)
(187,2.29228)
(188,2.27612)
(189,2.21364)
(190,2.16511)
(191,2.13208)
(192,2.11572)
(193,2.11674)
(194,2.13547)
(195,2.17182)
(196,2.2253)
(197,2.29499)
(198,2.37956)
(199,2.47728)
(200,2.58601)

                };

                \addplot [black!35] fill between [of = B and C];

            \end{axis}
             \node at (3.5,6.0) {Amazon Seattle case};
    \end{tikzpicture}
    \end{adjustbox}

\end{subfigure}%
    \caption{Convergence curves of the Huber training loss of DSPO for the synthetic and Amazon cases, reported over 5 training seeds with a training seed shaded area of 2 standard deviations.}
    \label{fig:dspo_convergence}
\end{figure}

For the pricing decision, $a^{pricing}$, we collect a large data set of $1$ million data points for the initial training phase. This data set is collected while using the NoPricing benchmark during simulation. DSPO uses a ReLU activation function for the fully connected layers and no activation function for the output node, as DSPO is a regression model. {For the experiments with finite capacity OOH locations, we add a small penalty to the cost function (See Equation~4) during training. This penalty ensures that the model prefers to nudge customers to OOH locations that still have a lot of remaining capacity. } 
\begin{equation}
    {P(k_{t,l}) = \lambda_t \frac{1}{1 + \exp\left(-w \left( \frac{k_{t,l}}{C_l} - \alpha \right) \right)},}
\end{equation}
{where $P(k_{t,l})$ is a penalty given the remaining capacity of location $l$, $C_l$ is the maximum capacity of location $l$, $w$ is the steepness of the penalty increase, and $\alpha$ is a threshold parameter representing the fraction of the capacity at which the penalty starts to increase significantly. The penalty is weighted using $\lambda$. After tuning, we found the following values: $w=\lceil0.1C_l\rceil$, and $\alpha=0.8$, i.e., the penalty function starts to increase noticeably when the locker is $80\%$ full, and it increases sharply as it approaches full capacity. The weight $\lambda_t$ starts at $0.1$ and increases linearly by $0.001$ at each time step. This way, the penalty has more weight at the end of the horizon.} The model is trained using adaptive moment estimation (adam) \citep{kingma} and the goal is to minimize the Huber loss, which is less sensitive to outliers compared to the $L_1$ loss. The Huber loss function is defined by:
\begin{equation}
    \mathcal{L}_{\delta}(x) = 
        \begin{cases}
        \frac{1}{2} x^2 & \text{for } |x| \leq \delta,\\
        \delta (|x| - \frac{1}{2} \delta) & \text{otherwise},
        \end{cases}
\end{equation}  
where $\delta$ is a tunable parameter. The convergence of the Huber loss is depicted in Figure~\ref{fig:dspo_convergence}, on the left for the synthetic case and on the right for the Amazon case. The figures show the result over $5$ training seeds.

\section{Benchmarks}\label{appendix::benchmarks}
We describe four benchmarks: the Hindsight and Foresight benchmarks in Section~\ref{appendix:hindforsight}, the linear benchmark in Section~\ref{appendix:linearBenchmark}, and PPO in Section~\ref{appendix:benchmarkppo}. We provide a detailed description and show training and loss curves.

\subsection{Hindsight and Foresight Benchmark}\label{appendix:hindforsight}

The Hindsight and Foresight benchmarks are based on the policies proposed in \citet{yang2016}. In their paper, the estimated costs of inserting a customer in a time slot are calculated. Instead of calculating the costs over time slots, we calculate the costs per insertion of a new delivery location $k$. The estimated costs of adding a delivery location $k$ are denoted by $\hat{C}^{Hindsight}_{k,t}$ and $\hat{C}^{Foresight}_{k,t}$, respectively. Obtaining $\hat{C}^{Hindsight}_{k,t}$ is relatively straightforward: the additional travel time of inserting a customer in route ${R}_{t-1}$ are first calculated, after which the salary, fuel, and service time costs related to delivery location $k$ can be determined. {The preliminary route $R_{t-1}$ contains a routing plan serving all known customers, potentially with multiple vehicles.} The Foresight benchmark is similar, but includes a weighted cost term:
\begin{equation}
    \hat{C}^{Foresight}_{k,t} = (1-\theta^H_t) \hat{C}^{Hindsight}_{k,t} + \theta^H_t\frac{1}{|\mathcal{P}|}\sum_{p\in\mathcal{P}}f(k,p),
\end{equation}
where $f(k,p)$ represents the insertion costs of feasibly inserting delivery location $k$ into historic route $p$. We use a pool of $10$ historic routes $R_T \in \mathcal{R}_T$, obtained using the StaticPricing benchmark on the training data set. {Note that these are final routing plans, as denoted by the capital $T$.} The weight $\theta^H_t$ is initialized on the value $\theta^H_0$ and decreased after every new insertion by $\Delta\theta^H$. Both parameters are considered tunable hyperparameters, see Appendix~\ref{appendix:hyperparams} for the tuning results. {We pre-select such that only delivery options that still have remaining capacity are considered. } For further details on these policies in a time-slotting setting, we refer to \citet{yang2016}.

\subsection{Linear benchmark}\label{appendix:linearBenchmark}

\begin{figure}[hbtp]
    \centering
    \begin{subfigure}[t]{.49\textwidth}
     \centering
    \centering
    \begin{adjustbox}{width=\textwidth}
   \begin{tikzpicture}[scale = 0.84]
            \begin{axis}[
                ymin = 0 ,ymax=10,
                xmin=0.5, xmax=50,
                 xlabel = Episodes ($\times 1\,000$),
                xlabel near ticks,
                 ylabel= Huber training loss,
                ylabel near ticks
                        ]

        \addplot[ name path = A] %
                coordinates {
(1,5.28043)
(2,5.10496)
(3,4.93367)
(4,4.76655)
(5,4.6036)
(6,4.44483)
(7,4.29023)
(8,4.1398)
(9,3.99355)
(10,3.85148)
(11,3.71358)
(12,3.57985)
(13,3.4503)
(14,3.32933)
(15,3.2295)
(16,3.25859)
(17,3.16983)
(18,2.97987)
(19,2.73268)
(20,2.73044)
(21,2.66067)
(22,2.59511)
(23,2.44906)
(24,2.48993)
(25,2.46359)
(26,2.26294)
(27,1.98715)
(28,1.95125)
(29,1.98107)
(30,1.90357)
(31,1.9041)
(32,2.00746)
(33,2.05746)
(34,2.07039)
(35,2.12937)
(36,2.20677)
(37,2.33748)
(38,2.19945)
(39,2.19806)
(40,2.1956)
(41,2.19208)
(42,2.18749)
(43,2.18183)
(44,2.17511)
(45,2.16733)
(46,2.15847)
(47,2.14856)
(48,2.13757)
(49,2.12552)
(50,2.11241)

                };

 \addplot[name path = B,draw=none,forget plot] %
                coordinates {
(1,3.78043)
(2,3.60496)
(3,3.43367)
(4,3.26655)
(5,3.1036)
(6,2.94483)
(7,2.79023)
(8,2.6398)
(9,2.49355)
(10,2.35148)
(11,2.21358)
(12,2.07985)
(13,1.9503)
(14,1.82933)
(15,1.7295)
(16,1.75859)
(17,1.66983)
(18,1.47987)
(19,1.23268)
(20,1.23044)
(21,1.16067)
(22,1.09511)
(23,0.94906)
(24,0.98993)
(25,0.96359)
(26,0.76294)
(27,0.48715)
(28,0.45125)
(29,0.48107)
(30,0.40357)
(31,0.4041)
(32,0.50746)
(33,0.55746)
(34,0.57039)
(35,0.62937)
(36,0.70677)
(37,0.83748)
(38,0.69945)
(39,0.69806)
(40,0.6956)
(41,0.69208)
(42,0.68749)
(43,0.68183)
(44,0.67511)
(45,0.66733)
(46,0.65847)
(47,0.64856)
(48,0.63757)
(49,0.62552)
(50,0.61241)

                };

             \addplot[name path = C,draw=none,forget plot] %
                coordinates {

(1,6.78043)
(2,6.60496)
(3,6.43367)
(4,6.26655)
(5,6.1036)
(6,5.94483)
(7,5.79023)
(8,5.6398)
(9,5.49355)
(10,5.35148)
(11,5.21358)
(12,5.07985)
(13,4.9503)
(14,4.82933)
(15,4.7295)
(16,4.75859)
(17,4.66983)
(18,4.47987)
(19,4.23268)
(20,4.23044)
(21,4.16067)
(22,4.09511)
(23,3.94906)
(24,3.98993)
(25,3.96359)
(26,3.76294)
(27,3.48715)
(28,3.45125)
(29,3.48107)
(30,3.40357)
(31,3.4041)
(32,3.50746)
(33,3.55746)
(34,3.57039)
(35,3.62937)
(36,3.70677)
(37,3.83748)
(38,3.69945)
(39,3.69806)
(40,3.6956)
(41,3.69208)
(42,3.68749)
(43,3.68183)
(44,3.67511)
(45,3.66733)
(46,3.65847)
(47,3.64856)
(48,3.63757)
(49,3.62552)
(50,3.61241)

                };

        \addplot [black!35] fill between [of = B and C];

            \end{axis}
            \node at (3.5,6.0) {Synthetic case};
    \end{tikzpicture}
    \end{adjustbox}

\end{subfigure}%
\begin{subfigure}[t]{.49\textwidth}
     \centering
    \centering
    \begin{adjustbox}{width=\textwidth}
   \begin{tikzpicture}[scale = 0.84]
                 \begin{axis}[
                ymin = 0 ,ymax=15,
                xmin=0.5, xmax=100,
                 xlabel = Episodes ($\times 1\,000$),
                xlabel near ticks,
                 ylabel= Huber training loss,
                ylabel near ticks
                        ]

        \addplot[ name path = A] %
                coordinates {
(1,8.72495)
(2,8.52989)
(3,8.36163)
(4,8.21789)
(5,8.09638)
(6,7.99482)
(7,7.91092)
(8,7.84241)
(9,7.78701)
(10,7.74242)
(11,7.70637)
(12,7.67657)
(13,7.65073)
(14,7.69621)
(15,7.58336)
(16,7.52212)
(17,7.56037)
(18,7.49801)
(19,7.37798)
(20,7.24984)
(21,7.1125)
(22,7.05789)
(23,6.8117)
(24,6.82408)
(25,6.67126)
(26,6.59316)
(27,6.41823)
(28,6.55855)
(29,6.47187)
(30,6.37005)
(31,6.29832)
(32,6.336)
(33,6.32142)
(34,6.35646)
(35,6.32354)
(36,6.11566)
(37,6.09127)
(38,6.06428)
(39,6.00678)
(40,5.99155)
(41,5.91445)
(42,5.60619)
(43,5.55462)
(44,5.47511)
(45,5.45292)
(46,5.36654)
(47,5.29995)
(48,5.29919)
(49,5.08131)
(50,4.95684)
(51,4.89648)
(52,4.96995)
(53,4.92296)
(54,4.85527)
(55,4.69039)
(56,4.70407)
(57,4.58712)
(58,4.5753)
(59,4.39337)
(60,4.27398)
(61,4.26479)
(62,4.27437)
(63,4.23696)
(64,4.27286)
(65,4.14038)
(66,4.11309)
(67,4.16795)
(68,4.1379)
(69,3.9337)
(70,3.81332)
(71,3.61085)
(72,3.37236)
(73,3.26331)
(74,3.04288)
(75,3.02372)
(76,3.06406)
(77,3.04327)
(78,2.97937)
(79,2.86682)
(80,2.83382)
(81,2.75633)
(82,2.68199)
(83,2.58904)
(84,2.6739)
(85,2.72705)
(86,2.85685)
(87,2.7797)
(88,2.82086)
(89,2.80739)
(90,2.7916)
(91,2.77409)
(92,2.7555)
(93,2.73643)
(94,2.71751)
(95,2.69934)
(96,2.68256)
(97,2.66777)
(98,2.65559)
(99,2.64664)
(100,2.64153)

                };

 \addplot[name path = C,draw=none,forget plot] %
                coordinates {
(1,6.22495)
(2,6.02989)
(3,5.86163)
(4,5.71789)
(5,5.59638)
(6,5.49482)
(7,5.41092)
(8,5.34241)
(9,5.28701)
(10,5.24242)
(11,5.20637)
(12,5.17657)
(13,5.15073)
(14,5.19621)
(15,5.08336)
(16,5.02212)
(17,5.06037)
(18,4.99801)
(19,4.87798)
(20,4.74984)
(21,4.6125)
(22,4.55789)
(23,4.3117)
(24,4.32408)
(25,4.17126)
(26,4.09316)
(27,3.91823)
(28,4.05855)
(29,3.97187)
(30,3.87005)
(31,3.79832)
(32,3.836)
(33,3.82142)
(34,3.85646)
(35,3.82354)
(36,3.61566)
(37,3.59127)
(38,3.56428)
(39,3.50678)
(40,3.49155)
(41,3.41445)
(42,3.10619)
(43,3.05462)
(44,2.97511)
(45,2.95292)
(46,2.86654)
(47,2.79995)
(48,2.79919)
(49,2.58131)
(50,2.45684)
(51,2.39648)
(52,2.46995)
(53,2.42296)
(54,2.35527)
(55,2.19039)
(56,2.20407)
(57,2.08712)
(58,2.0753)
(59,1.89337)
(60,1.77398)
(61,1.76479)
(62,1.77437)
(63,1.73696)
(64,1.77286)
(65,1.64038)
(66,1.61309)
(67,1.66795)
(68,1.6379)
(69,1.4337)
(70,1.31332)
(71,1.11085)
(72,0.87236)
(73,0.76331)
(74,0.54288)
(75,0.52372)
(76,0.56406)
(77,0.54327)
(78,0.47937)
(79,0.36682)
(80,0.33382)
(81,0.25633)
(82,0.18199)
(83,0.0890399999999998)
(84,0.1739)
(85,0.22705)
(86,0.35685)
(87,0.2797)
(88,0.32086)
(89,0.30739)
(90,0.2916)
(91,0.27409)
(92,0.2555)
(93,0.23643)
(94,0.21751)
(95,0.19934)
(96,0.18256)
(97,0.16777)
(98,0.15559)
(99,0.14664)
(100,0.14153)

                };

                      \addplot[name path = B,draw=none,forget plot] %
                coordinates {
(1,11.22495)
(2,11.02989)
(3,10.86163)
(4,10.71789)
(5,10.59638)
(6,10.49482)
(7,10.41092)
(8,10.34241)
(9,10.28701)
(10,10.24242)
(11,10.20637)
(12,10.17657)
(13,10.15073)
(14,10.19621)
(15,10.08336)
(16,10.02212)
(17,10.06037)
(18,9.99801)
(19,9.87798)
(20,9.74984)
(21,9.6125)
(22,9.55789)
(23,9.3117)
(24,9.32408)
(25,9.17126)
(26,9.09316)
(27,8.91823)
(28,9.05855)
(29,8.97187)
(30,8.87005)
(31,8.79832)
(32,8.836)
(33,8.82142)
(34,8.85646)
(35,8.82354)
(36,8.61566)
(37,8.59127)
(38,8.56428)
(39,8.50678)
(40,8.49155)
(41,8.41445)
(42,8.10619)
(43,8.05462)
(44,7.97511)
(45,7.95292)
(46,7.86654)
(47,7.79995)
(48,7.79919)
(49,7.58131)
(50,7.45684)
(51,7.39648)
(52,7.46995)
(53,7.42296)
(54,7.35527)
(55,7.19039)
(56,7.20407)
(57,7.08712)
(58,7.0753)
(59,6.89337)
(60,6.77398)
(61,6.76479)
(62,6.77437)
(63,6.73696)
(64,6.77286)
(65,6.64038)
(66,6.61309)
(67,6.66795)
(68,6.6379)
(69,6.4337)
(70,6.31332)
(71,6.11085)
(72,5.87236)
(73,5.76331)
(74,5.54288)
(75,5.52372)
(76,5.56406)
(77,5.54327)
(78,5.47937)
(79,5.36682)
(80,5.33382)
(81,5.25633)
(82,5.18199)
(83,5.08904)
(84,5.1739)
(85,5.22705)
(86,5.35685)
(87,5.2797)
(88,5.32086)
(89,5.30739)
(90,5.2916)
(91,5.27409)
(92,5.2555)
(93,5.23643)
(94,5.21751)
(95,5.19934)
(96,5.18256)
(97,5.16777)
(98,5.15559)
(99,5.14664)
(100,5.14153)

                };

                \addplot [black!35] fill between [of = B and C];

            \end{axis}
             \node at (3.5,6.0) {Amazon Seattle case};
    \end{tikzpicture}
    \end{adjustbox}

\end{subfigure}%
    \caption{Convergence curves of the Huber training loss of the linear benchmark for the synthetic and Amazon cases, reported over 5 training seeds with a training seed shaded area of 2 standard deviations.}
    \label{fig:linear_convergence}
\end{figure}
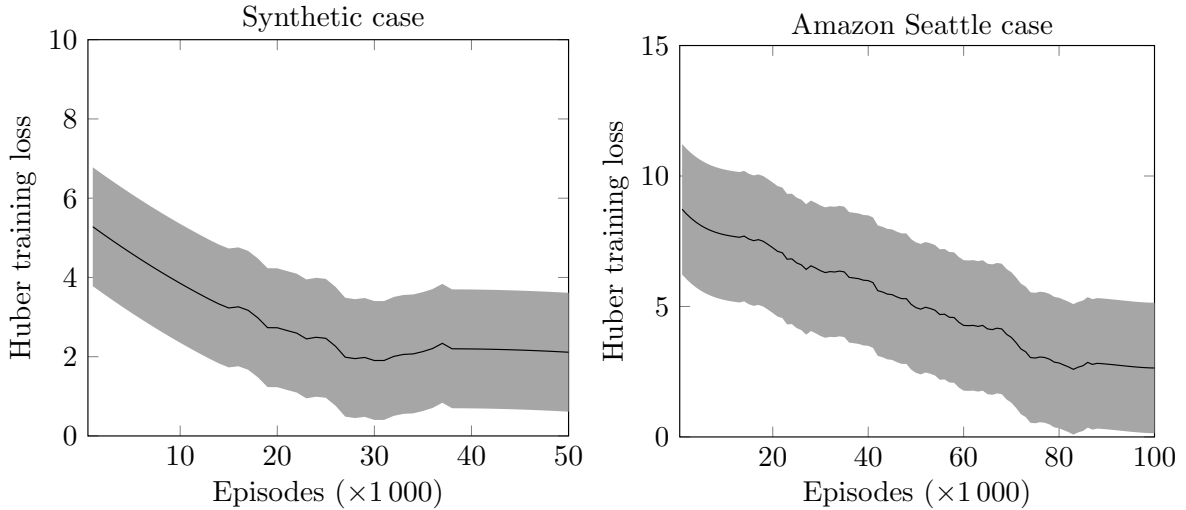

For the linear benchmark, we flatten the encoded state $\phi(s_t)$ as depicted in Figure~3 in the main manuscript. This flattened state is directly fed to the linear regression model, which yields the cost approximation $\hat{C}^{Linear}_{k,t}$ per delivery option $k$. We use stochastic gradient descent to train the model and minimize the Huber loss function. Figure~\ref{fig:linear_convergence} shows the convergence curves of the loss for the synthetic case on the left and the Amazon case on the right.

\subsection{PPO benchmark}\label{appendix:benchmarkppo}

For PPO, we provide the following state information: (i) the coordinates of the home address of the new customer, (ii) the coordinates of the $20$ delivery stops in $\mathcal{B}_{t-1}$ closest to the customer's home address, and {(iii) the remaining capacity of the $N$ closest OOH locations.} The continuous feature values are represented using the $f^{\textrm{th}}$ order coupled Fourier basis, which is a linear approximation using the terms of the Fourier series as features, see \citet{Konidaris_Osentoski_Thomas_2011}. We consider $f$, the order of the Fourier basis, to be a tunable hyperparameter, see Appendix~\ref{appendix:hyperparams} for the tuning results. Both the actor and critic use a neural network with three hidden layers and ReLU activation. The actor output layer uses a sigmoid activation function, which ensures that the Gaussian mean decision is in the range $[0,1]$. As a final step, we multiply the pricing vector with a scaling factor, corresponding to the pricing bounds $[-10,2]$. The price for home delivery is multiplied by $2$, and the prices for OOH delivery locations are multiplied by $-10$. {We provide the same penalty function for capacitated OOH locations to PPO as we do for DSPO, see Appendix~\ref{appendix:dspo]}.} We train the critic and actor using adam \citep{kingma} and calculate critic loss with the Huber loss function.

Algorithm~\ref{alg:algo3} outlines the PPO algorithm as detailed in \citet{SchulmanEtAl2017}. We begin by initializing the network weights $\boldsymbol{w}$ for the critic and $\boldsymbol{\theta}$ for the actor (1), followed by setting hyperparameters such as the Gaussian standard deviation $\boldsymbol{\sigma}$ and the learning rates $\alpha_{\mathrm{cr}}$ and $\alpha_{\mathrm{ac}}$ for the critic and actor, respectively (2). After initializing a state ${s}_0$ (3), we enter a loop for each timestep in the booking horizon (4). A continuous decision ${{a}}$ is generated by sampling from the policy $\pi_{\boldsymbol{\theta}}$, guided by the learned mean $\boldsymbol{\mu}$ and standard deviation ${\boldsymbol{\sigma}}$, along with the weights $\boldsymbol{\theta}$ (5). Upon applying decision ${a}$ to the environment, we observe the transition to the subsequent state ${s}_{t+1}$ (6), which we then store in the trajectory buffer $\mathcal{T}$ (7). Every $T$ steps, updates are made to the actor and critic networks (8). Advantages are computed using the truncated Generalized Advantage Estimation (GAE) as follows (9):
\begin{equation}
\hat{A}_t(r,{s}_t,{{a}},{s}_{t+1}) = \sum_{t'=t}^{T}(\lambda\gamma)^{t'-t}\delta_{t'},
\end{equation}
where $\lambda$ is the temporal difference discount parameter and $\delta$ is defined by:
\begin{equation}
    \delta = r + \gamma\, Q({s}_{t+1},{{a}},\boldsymbol{w}) - Q({s}_t,{{{a}}},\boldsymbol{w}).
\end{equation}

We then proceed to optimize the policy loss (10) as described in \citet{SchulmanEtAl2017} and the value loss (11) based on the $n$-step return:
\begin{equation}
    G_t^n = r_t + \gamma\,r_{t+1} + \ldots + \gamma^{n}\,Q_{\boldsymbol{w}}({s}_{t+n}, {{a}}).
\end{equation}

The trajectory buffer $\mathcal{T}$ is emptied thereafter (12).
\begin{algorithm}[hbtp]
\linespread{1}\selectfont
	\caption{{PPO} algorithm (Gaussian policy)}\label{alg:algo3}
	\begin{algorithmic}[1]
		\State  Initialize critic and actor network weights $\boldsymbol{w}$, $\boldsymbol{\theta}$
  		\State Set hyperparameters: $\boldsymbol{\sigma}$, $\alpha_{\mathrm{cr}}$, $\alpha_{\mathrm{ac}}$
		\For{each episode}
		\For{t=1,2,\ldots,T}
		\State ${{a}} \gets \pi_{\boldsymbol{\theta}}({s}_t)$ (based on $\boldsymbol{\sigma}$)
		\State Apply ${a}$ to environment, observe successor state ${s}_{t+1}$
        \State Store $({s}_t,{{a}},{s}_{t+1})_t$ in $\mathcal{T}$
    \EndFor
    \State Compute negative rewards $r$ based on Equation~4 and add them to $\mathcal{T}_T$
        \State Compute advantages $\hat{A}_t(r,{s}_t,{{a}},{s}_{t+1})$ and $\log\left(\pi_{\boldsymbol{\theta}}({{a}})\right)$
        \State Optimize clipped policy loss based on $\hat{A}(r,{s}_t,{{a}},{s}_{t+1})$ and $\log\left(\pi_{\boldsymbol{\theta}}({{a}})\right)$
        \State Optimize critic loss based on $n$-step return
    \State Empty $\mathcal{T}$     
		
		\EndFor
	\end{algorithmic}
\end{algorithm}

\begin{figure}[H]
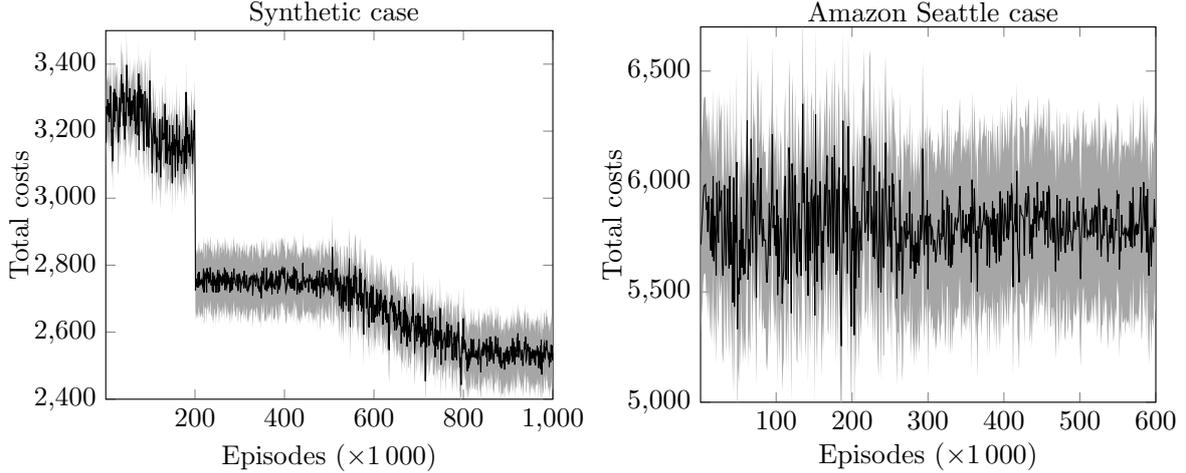

    \centering
    \begin{subfigure}[t]{.49\textwidth}
     \centering
    \centering
    \begin{adjustbox}{width=\textwidth}

    \end{adjustbox}

\end{subfigure}%
    \caption{Convergence curves of PPO for the synthetic and Amazon cases, reported over 5 training seeds with a training seed shaded area of 2 standard deviations.}
    \label{fig:ppo convergence}
\end{figure}

The convergence of the costs over training episodes is depicted in Figure~\ref{fig:ppo convergence}, on the left for the synthetic case, and on the right for the Amazon case. PPO requires many episodes to train, probably caused by the sparse cost structure of the problem. For the synthetic case, PPO seems to require \num[group-separator={,}]{200000} episodes to explore, after which it suddenly jumps to a well-performing policy. For the Amazon case, PPO is unable to converge to a performant policy within \num[group-separator={,}]{600000} episodes. 

\section{Hyperparameters}\label{appendix:hyperparams}
In this section, we provide an overview of the hyperparameter tuning results for the Foresight benchmark, linear benchmark, PPO, and DSPO, see Table~\ref{tab:all_hyperparameters}. 

The Foresight benchmark has two parameters: the initial weight given to the insertion costs in the current route ${R}_{t-1}$, $\theta_0^H$, and by how much this weight is decreased after every new customer insertion, $\Delta\theta^H$. We found that setting $\theta_0^H=1.0$ and $\Delta\theta^H=\frac{\theta_0^H}{\mathbb{E}[{D}]}$ yields the best results, which is the same setting as in \citet{yang2016}.

For PPO, we study both (i) learning the second moment of the Gaussian distribution, \(\sigma\), and (ii) setting \(\sigma\) to a constant value. In Table~\ref{tab:all_hyperparameters}, a \(\dag\) indicates that \(\sigma\) was learned by the actor. We maintain a learning rate for the actor that is ten times smaller than that of the critic, to guarantee that the critic's provided values are up-to-date. Both the actor and critic use a fully connected neural network with three hidden layers and ReLU activation. 

Apart from the DSPO hyperparameters in Table~\ref{tab:all_hyperparameters}, we use the following default settings for the CNN. The first convolutional layer has 32 filters for the synthetic case and 64 filters for the Amazon case. The second convolutional layer has two times more filters compared to the first layer. The kernel has dimension $(3,3)$ for the convolution layers and $(2,2)$ for the pooling layer. The stride of the kernel is $(1,1)$, padding is $(1,1)$, and we do not use dilation.

\begin{table}[H]
	\centering
	 \def\arraystretch{1.1}
	\caption{Hyperparameter tuning, FB = Foresight benchmark, LB = Linear benchmark.}
	\label{tab:all_hyperparameters}
	\begin{tabular}{llccc}
		\hline
        & &   & \multicolumn{2}{c}{Selected values} \\
        \cmidrule(r){4-5}
		& Hyperparameters & Set of values & Synthetic case & Amazon case \\
		\hline
   \multirow{2}{*}{\begin{sideways} FB \end{sideways}} & $\theta^H_0$ (initial insertion weight) & $\{1.0,0.75,0.25\}$ & $1.0$ & $1.0$  \\
   & $\Delta\theta^H$ (insertion weight update)  & $\{\theta^H_0 \slash \mathbb{E}[{D}],0.05,0.1\}$ & $\theta^H_0\slash\mathbb{E}[{D}]$ & $\theta^H_0\slash\mathbb{E}[{D}]$ \\
  \hline
  
    \multirow{2}{*}{\begin{sideways} LB \end{sideways}}  & Huber loss $\delta$ & $\{0.5,0.75,1.0,1.35,1.5\}$ & 1.0 & 1.0  \\
        & $\alpha^{LB}$ (learning rate) & $\{10^{-2},10^{-3},10^{-4}\}$ & $10^{-2}$ & $10^{-3}$  \\
  \hline 
       \multirow{11}{*}{\begin{sideways} PPO \end{sideways}}  & $\gamma$ (discount factor) & $\{0.9,0.99\}$ & 0.99 & 0.99  \\
        & $\alpha^{PPO}_{cr}$ (learning rate critic) & $\{10^{-2},10^{-3},10^{-4},10^{-5}\}$ & $10^{-4}$ & $10^{-4}$  \\
        & $\alpha^{PPO}_{ac}$ (learning rate actor) & $\{10^{-2},10^{-3},10^{-4},10^{-5}\}$ & $10^{-5}$ & $10^{-5}$  \\
        & $\sigma$ & $\{\dag,0.25,0.5,1\}$ & 0.25 & 0.5  \\
       & Huber loss $\delta$ & $\{0.5,0.75,1.0,1.35,1.5\}$ & 1.0 & 1.0  \\
        & \# actor NN nodes/layer & $\{8,16,32\}$ & 8 & 16 \\
         & \# critic NN nodes/layer & $\{8,16,32,64\}$ & 16 & 32 \\
         & Batch size & $\{32,64,128\}$ & 128 & 128 \\
         & $f$ (Fourier order) & $\{2,3,4\}$ & 3 & 3 \\
         & Clipping factor & $\{0.1,0.2,0.3\}$ & 0.2 & 0.2 \\
         & GAE $\lambda$ & $\{0.9,0.95,0.99,1.0\}$ & 0.95 & 0.95 \\
  \hline 
  \multirow{7}{*}{\begin{sideways} {DSPO} \end{sideways}} 
  & M (\# grids state spatial dimension) & \{25,100,900,1600,3600\} & 100  & 1600 \\
  & $D^T$ (\# layers state temporal dimension) & \{1,2,3,4,5,6,7,9,10\} & 3 & 8 \\
  & Batch size & $\{32,64,128\}$ & 64 & 128 \\
   & $\alpha^{DSPO}$ (learning rate)  & $\{10^{-2},10^{-3},10^{-4},10^{-5}\}$ & $10^{-3}$ & $10^{-3}$  \\
   &Huber loss $\delta$ & $\{0.5,0.75,1.0,1.35,1.5\}$ & 1.0 & 1.0  \\
   & \# NN nodes/FC layer & $\{32,64,128,256\}$ & 128 & 256 \\
   & Dropout rate FC layer & $\{0,0.01,0.05,0.1\}$ & 0.0 & 0.05 \\
		\hline
	\end{tabular}
\end{table} 

\section{Complementary Figures}\

\begin{figure}[hbtp]
    \centering
     \begin{subfigure}{1.0\textwidth}
     \centering
    \includegraphics[scale=0.25,angle =-90 ]{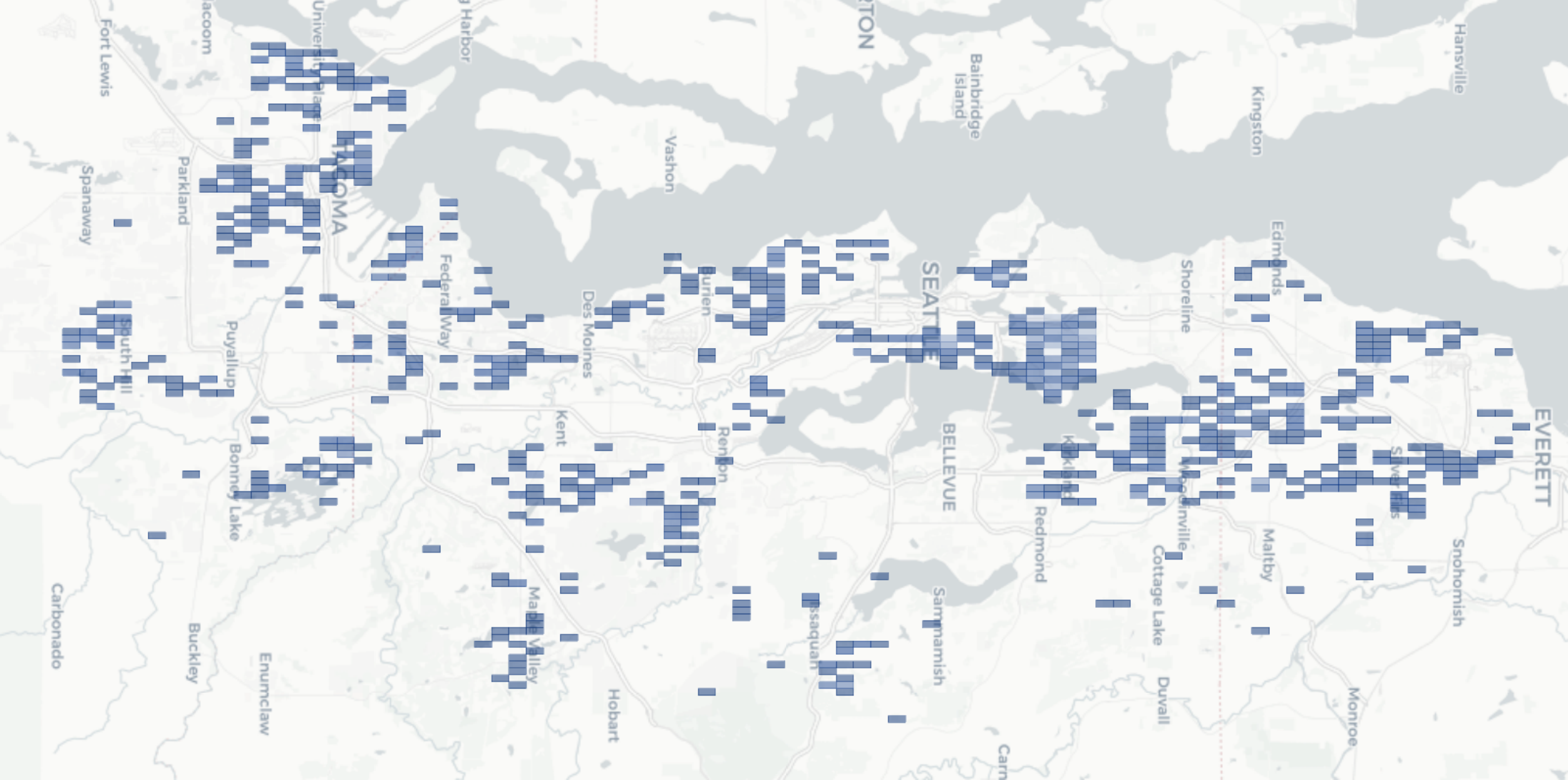}
    \end{subfigure}   
    \caption{Seattle heatmap of delivery stops on an average day.}
    \label{fig:seattle_heat_delivery}
\end{figure}

In this section, we provide complementary figures. Figure~\ref{fig:seattle_heat_delivery} provides a heatmap of the number of delivery stops per cell over the complete service area of Seattle on an average day. Note that the figure is tilted by $90$ degrees such that the north is oriented to the right.

\begin{figure}[hbtp]
    \centering
     \begin{subfigure}{0.85\textwidth}
     \centering
    \includegraphics[scale=0.19]{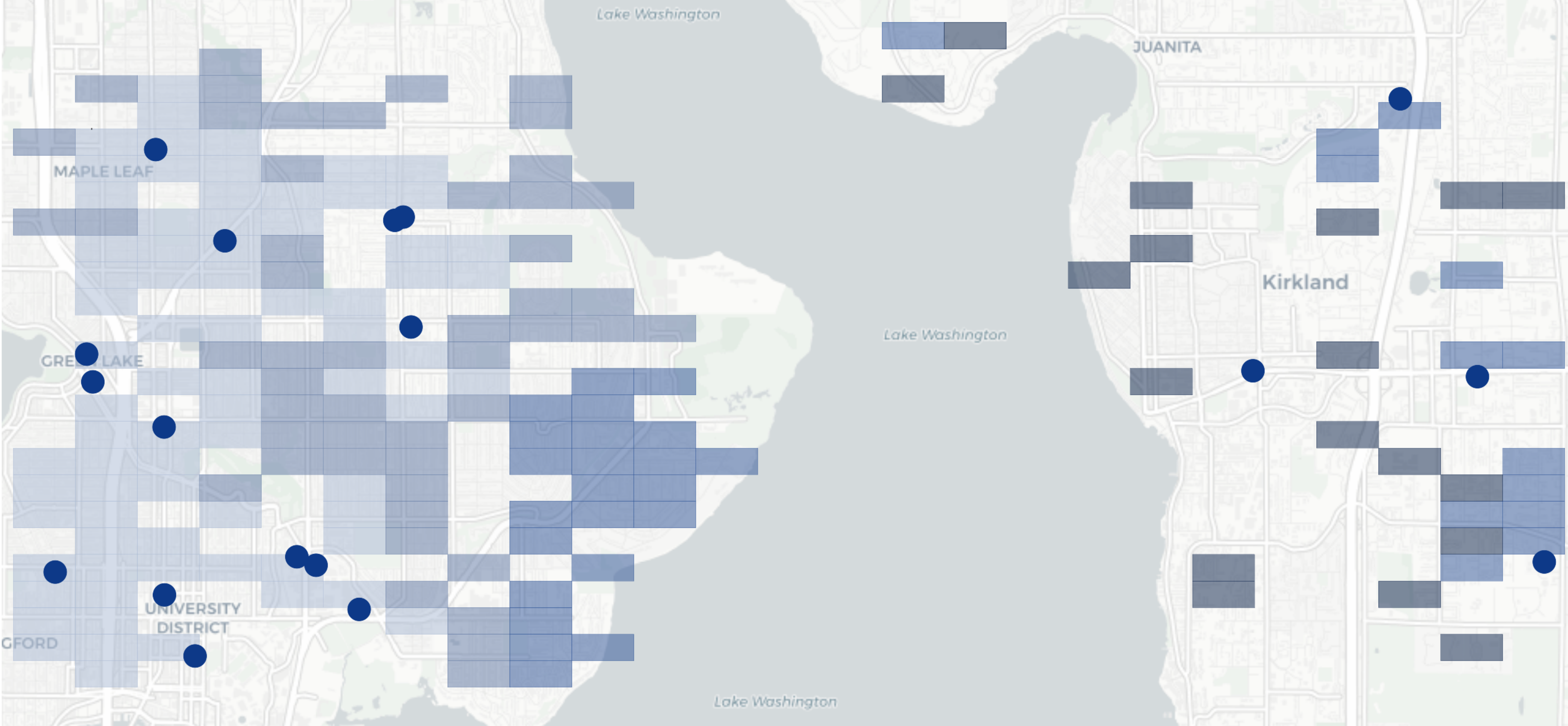}
    \end{subfigure}  
     \begin{subfigure}{0.1\textwidth}
    \centering
        \begin{tikzpicture}
            \pgfplotscolorbardrawstandalone[
            view={0}{90},
            colorbar,
             point meta min=2,
            point meta max=70,
            colormap = {blueDark}{color(0cm)  = (colorTSLight);color(1cm) = (colorTSDark)},
            colorbar style={
            at={(0,0)},anchor=south west,ylabel=Predicted costs as $\%$ of revenue,yticklabel = \pgfmathprintnumber\tick\%}
            ]
\end{tikzpicture}
    \end{subfigure} 
     \begin{subfigure}{0.85\textwidth}
     \centering
    \includegraphics[clip, trim=3.5cm 0.0cm 0cm 0.0cm]{img/legend.pdf}
    \end{subfigure}
   
    \caption{Northern Seattle heatmap of the estimated delivery costs according to DSPO at the cutoff time.}
    \label{fig:seattleheat_pred}
\end{figure}

\begin{figure}[hbtp]
    \centering
     \begin{subfigure}{0.85\textwidth}
     \centering
    \includegraphics[scale=0.19]{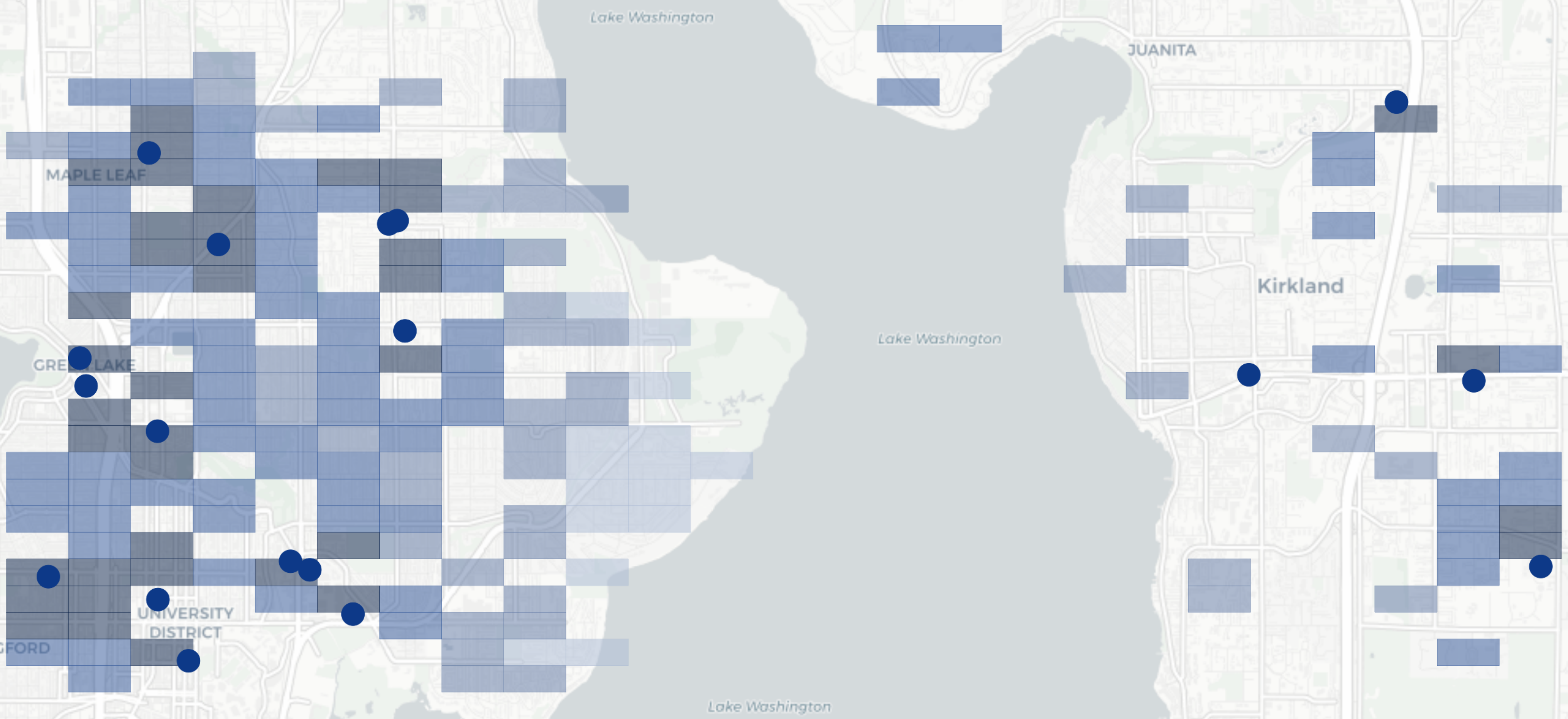}
    \end{subfigure}  
     \begin{subfigure}{0.1\textwidth}
    \centering
        \begin{tikzpicture}
    \pgfplotscolorbardrawstandalone[
            view={0}{90},
            colorbar,
            point meta min=10,
            point meta max=80,
            colormap = {blueDark}{color(0cm)  = (colorTSLight);color(1cm) = (colorTSDark)},
            colorbar style={
            at={(0,0)},anchor=south west,ylabel=OOH utilization,yticklabel = \pgfmathprintnumber\tick\%}
            ]
\end{tikzpicture}
    \end{subfigure} 
     \begin{subfigure}{0.85\textwidth}
     \centering
    \includegraphics[clip, trim=3.5cm 0.0cm 0cm 0.0cm]{img/legend.pdf}
    \end{subfigure}
   
    \caption{Northern Seattle heatmap of the utilization of OOH delivery when using DSPO.}
    \label{fig:seattleheat_util}
\end{figure}

Figures~\ref{fig:seattleheat_pred} and~\ref{fig:seattleheat_util} depict heatmaps of the costs predicted by DSPO at cutoff time and the OOH utilization when using DSPO, respectively. {These exemplary figures help to understand the decisions resulting from DSPO.} For visualization purposes, we zoom in on an exemplary part of Seattle: the densely populated neighborhoods in Northern Seattle, and the suburb of east Seattle, Kirkland. The suburb has fewer customers and is harder to reach since it is separated from Seattle by Lake Washington.


Figure~\ref{fig:seattleheat_pred} shows that DSPO predicts higher delivery costs for more remote areas and areas that have lower OOH density. The Kirkland areas and the parts with lower OOH density have higher predicted costs. DSPO provides higher discounts for choosing OOH delivery to customers in these areas.

Figure~\ref{fig:seattleheat_util} show that in areas with high OOH density, the utilization of such delivery options is highest, ranging from $60\%$ to $80\%$.  Conversely, in areas with sparser OOH locations, the usage rate is lower. Interestingly, the utilization in remote Kirkland areas is slightly higher compared to the low OOH density areas in Northern Seattle. Probably, customers in Kirkland areas are marked as more expensive to serve since they are more remote, and consequently, higher incentives are given to those customers.

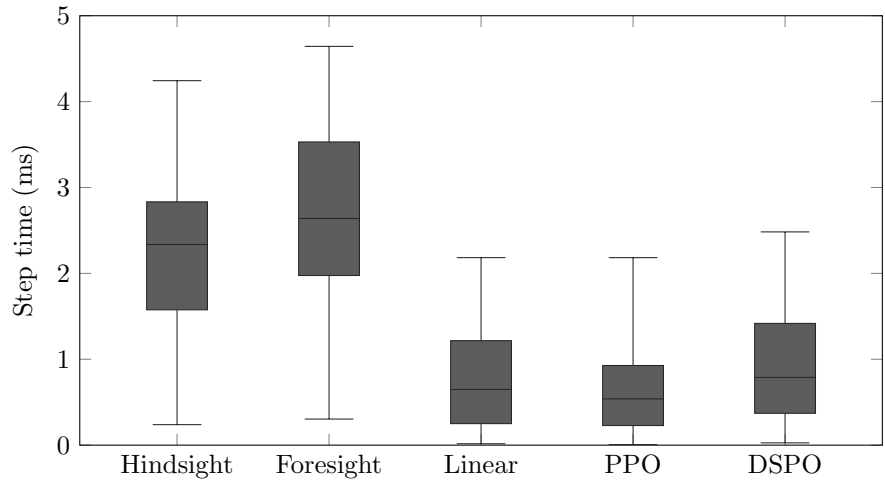
\begin{figure}[hbtp]
    \centering
   \begin{tikzpicture}
  \begin{axis} [
         boxplot/draw direction=y,
    /pgfplots/boxplot/box extend=0.4,
     xtick={1,2,3,4,5},
    xticklabels={Hindsight, Foresight, Linear, PPO, DSPO},
     ylabel = Step time (ms),
                ylabel near ticks,
    cycle list={{black!85}},
    every axis plot/.append style={fill,fill opacity=0.75},
     x tick label style={
        text width=2.5cm,
        align=center
    },
    ymax=5,ymin=0,
       x=2cm,
    ]
    \addplot+ [
    area legend,boxplot prepared={
   upper quartile=2.8332,
      lower quartile=1.57577,
      upper whisker=4.24387,
       median=2.3362,
      lower whisker=0.2386,
      draw position=1
    },
    ] coordinates {};

      \addplot+ [
    area legend,boxplot prepared={
   upper quartile=3.5304,
      lower quartile=1.97537,
      upper whisker=4.64327,
       median=2.6392,
      lower whisker=0.30386,
      draw position=2
    },
    ] coordinates {};

      \addplot+ [
     area legend,boxplot prepared={
   upper quartile=1.216435,
      lower quartile=0.250655,
      upper whisker=2.18336,
       median=0.648455,
      lower whisker=0.016432,
      draw position=3
    },
    ] coordinates {};

      \addplot+ [
      area legend,boxplot prepared={
   upper quartile=0.927435,
      lower quartile=0.2283,
      upper whisker=2.18321,
       median=0.538435,
      lower whisker=0.00642,
      draw position=4
    },
    ] coordinates {};

      \addplot+ [
    area legend,boxplot prepared={
   upper quartile=1.417635,
      lower quartile=0.3709185,
      upper whisker=2.48316,
       median=0.788455,
      lower whisker=0.02642,
      draw position=5
    },
    ] coordinates {};

  \end{axis}
\end{tikzpicture}
    \caption{Step time in milliseconds of the studied dynamic pricing methods (Amazon case).}
    \label{fig:boxplot_steptime}
\end{figure}

Figure~\ref{fig:boxplot_steptime} shows a boxplot of the online step time in milliseconds for all dynamic pricing methods. The online step time is the time required to obtain a single decision. This is relevant in light of website load times required for providing the delivery options to the online shoppers. Although the route insertion calculation is in essence relatively cheap in terms of computational effort, it does require the calculation of new travel times, i.e., in a practical situation it requires an API request. The linear benchmark, PPO, and DSPO do not require this calculation and, hence, provide faster results. We note that in absolute terms the differences between the policies are relatively small.

\end{document}